\documentclass{article}

\PassOptionsToPackage{numbers, compress}{natbib}

\usepackage[final]{neurips_2022}

\usepackage[utf8]{inputenc} 
\usepackage[T1]{fontenc}    
\usepackage{hyperref}       
\usepackage{url}            
\usepackage{booktabs}       
\usepackage{amsfonts}       
\usepackage{amsmath}
\usepackage{bm}
\usepackage{mathtools}
\usepackage{amsthm}
\usepackage{nicefrac}       
\usepackage{microtype}      
\usepackage{xcolor}         
\usepackage{graphicx}
\usepackage{caption}
\usepackage{subcaption}
\usepackage{wrapfig}
\usepackage{multirow}
\usepackage{longtable}
\usepackage{rotating}
\usepackage{threeparttable}
\usepackage{algorithm}
\usepackage{algorithmic}
\usepackage[capitalise]{cleveref}
\usepackage[misc]{ifsym}

\makeatletter
\newtheorem*{rep@theorem}{\rep@title}
\newcommand{\newreptheorem}[2]{%
\newenvironment{rep#1}[1]{%
 \def\rep@title{#2 \ref{##1}}%
 \begin{rep@theorem}}%
 {\end{rep@theorem}}}
\makeatother

\newtheorem{prop}{Proposition}

\newtheorem{lemma}{Lemma}
\newreptheorem{lemma}{Lemma}

\newcommand{\ourmodel}[1][]{{D$^{3}$VAE#1}}

\Crefname{equation}{Eq.}{Eqs.}
\crefname{theorem}{Theorem}{Theorems}
\crefname{prop}{Proposition}{Propositions}
\crefname{lemma}{Lemma}{Lemmas}

\title{
Generative Time Series Forecasting with Diffusion, Denoise, and Disentanglement
}

\author{%
  Yan~Li%
  \textsuperscript{$\S \, \dagger \,$}%
  \thanks{
  This work was done when the first author was an intern at 
  Baidu Research under the supervision of 
  the second author.
  }~~,~
  Xinjiang~Lu%
  \textsuperscript{$\dagger \, {\textrm{\Letter}}$}~,~%
  Yaqing~Wang%
  \textsuperscript{$\dagger$},~%
  Dejing~Dou%
  \textsuperscript{$\dagger$}
  \\
  \textsuperscript{$\dagger$}Business~Intelligence~Lab,~Baidu~Research
  \\
  \textsuperscript{$\S$}Zhejiang~University,~China
  \\
  \texttt{ly21121@zju.edu.cn,~$\{$luxinjiang,wangyaqing01,doudejing$\}$@baidu.com}
  \\
}

\begin{document}

\maketitle

\begin{abstract}

Time series forecasting has been a widely explored task of great importance in many applications.
However, it is common that real-world time series data are recorded in a short time period, which results in a big gap between the deep model and the limited and noisy time series. 
In this work, we propose to address the time series forecasting problem with generative modeling and propose a bidirectional variational auto-encoder (BVAE) equipped with diffusion, denoise, and disentanglement, namely \ourmodel.
Specifically, a coupled diffusion probabilistic model is proposed to augment the time series data without increasing the aleatoric uncertainty and implement a more tractable inference process with BVAE. 
To ensure the generated series move toward the true target, we further propose to adapt and integrate the multiscale denoising score matching into the diffusion process for time series forecasting. 
In addition, to enhance the interpretability and stability of the prediction, we treat the latent variable in a multivariate manner and disentangle them on top of minimizing total correlation. 
Extensive experiments on synthetic and real-world data show that \ourmodel~outperforms competitive algorithms with remarkable margins. 
Our implementation is available at \url{https://github.com/PaddlePaddle/PaddleSpatial/tree/main/research/D3VAE}. 

\end{abstract}

\setcounter{footnote}{0}

\section{Introduction}

Time series forecasting is of great importance for risk-averse and decision-making.~%
Traditional RNN-based methods capture temporal dependencies of the time series to predict the future. 
Long short-term memories (LSTMs) and gated recurrent units (GRUs)~\cite{yu2019review,greff2016lstm,gers2002learning,sherstinsky2020fundamentals} introduce the gate functions into the cell structure to handle long-term dependencies effectively.~%
The models based on convolutional neural networks (CNNs)
capture complex inner patterns of the time series through convolutional operations~\cite{lea2016temporal,borovykh2017conditional,binkowski2018autoregressive}.~%
Recently, the Transformer-based models have shown great performance in time series forecasting~\cite{xu2021autoformer,zhou2021informer,kitaev2020reformer,li2019enhancing} with the help of multi-head self-attention. 
However, one big issue of neural networks in time series forecasting is the   uncertainty~\cite{gawlikowski2021survey,abdar2021review} resulting from the properties of the deep structure.~%
The models based on vector autoregression (VAR)~\cite{cao2003support,fokianos2009poisson,kim2003financial} try to model the distribution of time series from hidden states, which could provide more reliability to the prediction, while the performance is not satisfactory~\cite{lai2018modeling}. 

Interpretable representation learning is another merit of time series forecasting.~%
Variational auto-encoders (VAEs) have shown not only the superiority in modeling latent distributions of the data and reducing the gradient noise~\cite{roeder2017sticking,kingma2016improved, li2016renyi, vahdat2020nvae} but also the interpretability of time series forecasting~\cite{fortuin2020gp,fortuin2019som}. 
However, the interpretability of VAEs might be inferior due to the entangled latent variables. 
There have been  efforts to learn representation disentangling~\cite{kim2018disentangling,bengio2013representation,higgins2016beta}, which show that the well-disentangled representation can  improve the performance and robustness of the algorithm.

Moreover, real-world time series are often noisy and recorded in a short time period,  which may result in overfitting and generalization issues~\cite{gamboa2017deep,wang2018gaussian,zou2019complex,arima_2020}{\footnote{The detailed literature review can be found in \cref{sec:related}.}}.~%
To this end, we address the time series forecasting problem with generative modeling.
Specifically, we propose a bidirectional variational auto-encoder (BVAE) equipped with diffusion, denoise, and disentanglement, namely \ourmodel. 
More specifically, 
we first propose a coupled diffusion probabilistic model to remedy the limitation of time series data by augmenting the input time series, as well as the output time series, inspired by the forward process of the diffusion model~\cite{sohl2015deep,ho2020denoising,nichol2021improved,rasul2021autoregressive}.~%
Besides, 
we adapt the Nouveau VAE~\cite{vahdat2020nvae} to the time series forecasting task and develop a BVAE as a substitute for the reverse process of the diffusion model.~%
In this way, the expressiveness of the diffusion model plus the tractability of the VAE can be leveraged together for generative time series forecasting.~%
Though the merit of generalizability is helpful, the diffused samples might be corrupted, which results in a generative model moving toward the noisy target.
Therefore, we further develop a scaled denoising score-matching network for cleaning diffused target time series.~%
In addition, we disentangle the latent variables of the time series by assuming that different disentangled dimensions of the latent variables correspond to different temporal patterns (such as trend, seasonality, etc.). 
Our contributions can be summarized as follows: 

\begin{itemize}
\item
We propose a coupled diffusion probabilistic model aiming to reduce the aleatoric uncertainty of the time series and improve the generalization capability of the generative model. 

\item
We integrate the multiscale denoising score matching into the coupled diffusion process to improve the accuracy of generated results. 

\item 
We disentangle the latent variables of the generative model to improve the interpretability for time series forecasting. 

\item 
Extensive experiments on synthetic and real-world datasets demonstrate that \ourmodel ~outperforms  competitive  baselines with satisfactory  margins.

\end{itemize}

\section{Methodology} 
\label{sec:method}

\subsection{Generative Time Series Forecasting}

{\bf Problem Formulation.~}
Given an input multivariate time series $X = \{x_1, x_2, \cdots, x_n \,|\, x_i \in \mathbb{R}^{d} \} $ and the corresponding target time series $ Y = \{y_{n+1}, y_{n+2}, \cdots , y_{n+m} \,|\, y_j \in \mathbb{R}^{d'} \}$ ($ d^{'} \leq d $).~%
We assume that $Y$ can be generated from latent variables $Z \in \Omega_Z$ that can be drawn from the Gaussian distribution $Z \sim p(Z|X)$. 
The latent distribution can be further formulated as $ p_\phi(Z|X) = g_\phi(X) $ where  $g_\phi $ denotes a nonlinear function.
Then, the data density of the target series is given by:
\begin{equation}
p_\theta(Y) = \int_{\Omega_Z} p_\phi(Z|X)( Y - f_\theta(Z) ) dZ \, ,
\end{equation}
where $f_\theta$ denotes a parameterized function.~%
The target time series can be obtained directly  by sampling from  $p_\theta(Y)$.

\begin{figure}[t]
    \centering
    \includegraphics[width=.9\textwidth]{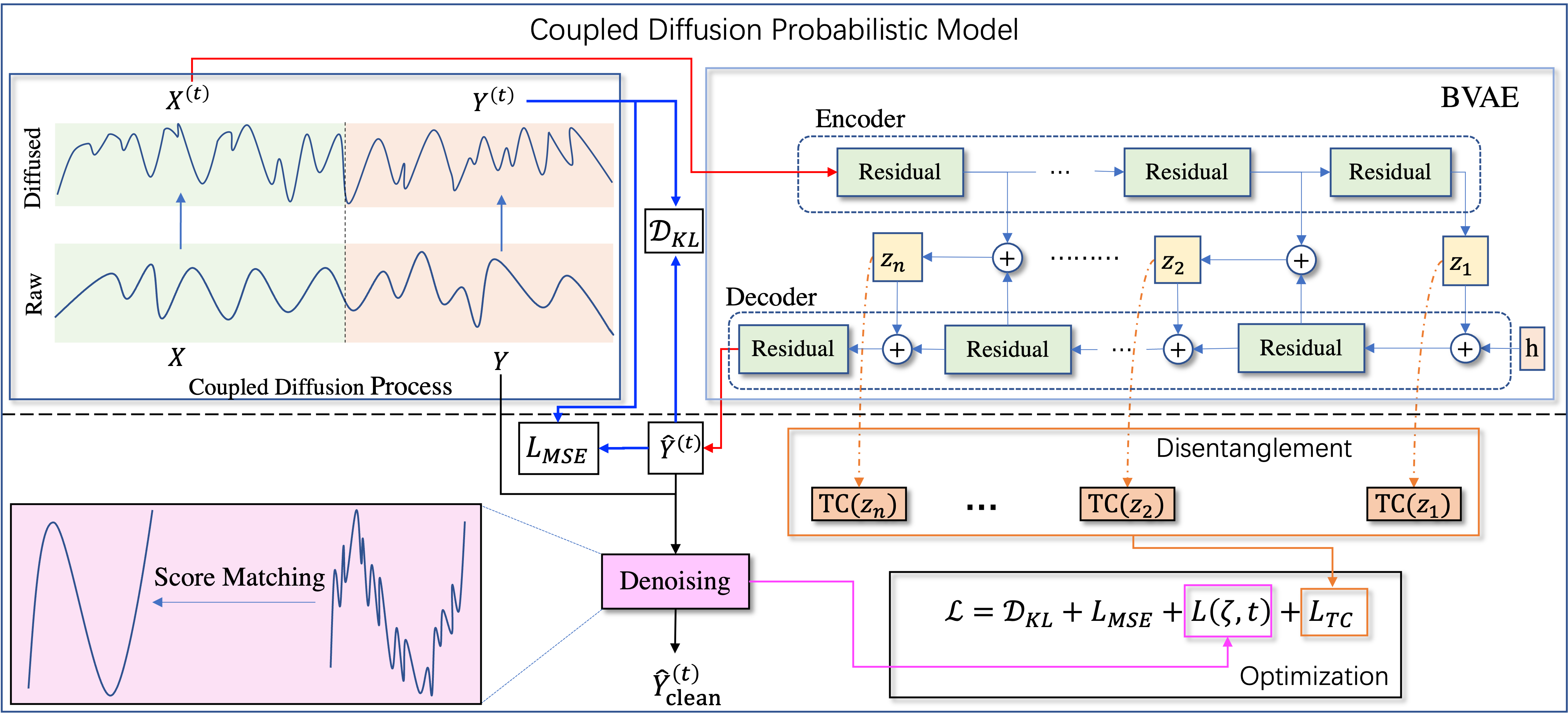}
    \caption{
    The framework overview of \ourmodel. 
    First, the input and output series are augmented simultaneously with the {\em coupled diffusion process}.~%
    Then the diffused input series are fed into a proposed BVAE model for inference, which can be deemed a {\em reverse process}.~%
    A denoising score-matching mechanism is applied to make the estimated target move toward the true target series. 
    Meanwhile, the latent states in BVAE are leveraged for disentangling such that the model interpretability and reliability can be improved.
    }	
    \label{fig:method:framework}
\end{figure}

In our problem setting, 
time series forecasting is to learn the representation $Z$ that captures useful signals of $X$, 
and map the low dimensional $X$ to the latent space with high expressiveness. 
The framework overview of \ourmodel ~is demonstrated in \cref{fig:method:framework}. 
Before diving into the detailed techniques, we first introduce a preliminary proposition.

\begin{prop} \label{prop1}
Given a time series $X$ and its inherent noise  $\epsilon_X$,  we have the  decomposition: $X = \langle X_{r}, \, \epsilon_X \rangle $, where $X_r$ is the ideal time series data without noise. 
$ X_{r}$ and $\epsilon_X $ are independent of each other. 
Let $p_{\phi} (Z|X) = p_{\phi} (Z| X_{r} , \epsilon_X) $,
the estimated target series $\widehat{Y}$  can be generated with the distribution $p_{\theta} (\widehat{Y}|Z) = p_{\theta} (\widehat{Y}_r|Z) \cdot p_{\theta} (\epsilon_{\widehat{Y}}|Z)$ where $ \widehat{Y}_r $ is the ideal part of $ \widehat{Y} $ and $ \epsilon_{\widehat{Y}} $ is the estimation noise. 
Without loss of generality, $ \widehat{Y}_r $ can be fully captured by the model.
That is,  $ \| Y_r - \widehat{Y}_r \| \longrightarrow 0 $ where $Y_r$ is the ideal part of ground truth target series $Y$.
In addition, $Y$ can be decomposed as $Y = \langle \widehat{Y}_r , \epsilon_Y \rangle$ ($\epsilon_Y$ denotes the noise of $Y$). 
Therefore, the error between ground truth and prediction, i.e., $ \| Y - \widehat{Y} \| = \| \epsilon_Y - \epsilon_{\widehat{Y}} \| > 0$, 
can be deemed as the combination of aleatoric uncertainty and epistemic uncertainty.
\end{prop}

\subsection{Coupled Diffusion Probabilistic Model} \label{sec2.2}

The diffusion probabilistic model (diffusion model for brevity) is a family of latent variable models aiming to generate high-quality samples. 
To equip the generative time series forecasting model with high expressiveness, a coupled {\em forward process} is developed to augment the input series and target series synchronously. 
Besides, in the forecasting task, more tractable and accurate prediction is expected. 
To achieve this, we propose a bidirectional variational auto-encoder (BVAE) to take the place of the {\em reverse process} in the diffusion model. 
We present the technical details in the following two parts, respectively.

\subsubsection{Coupled Diffusion Process}

\begin{figure}[t]
    \centering
    \begin{minipage}[t]{1.0\textwidth}
    \centering
    \includegraphics[width=0.8\textwidth]{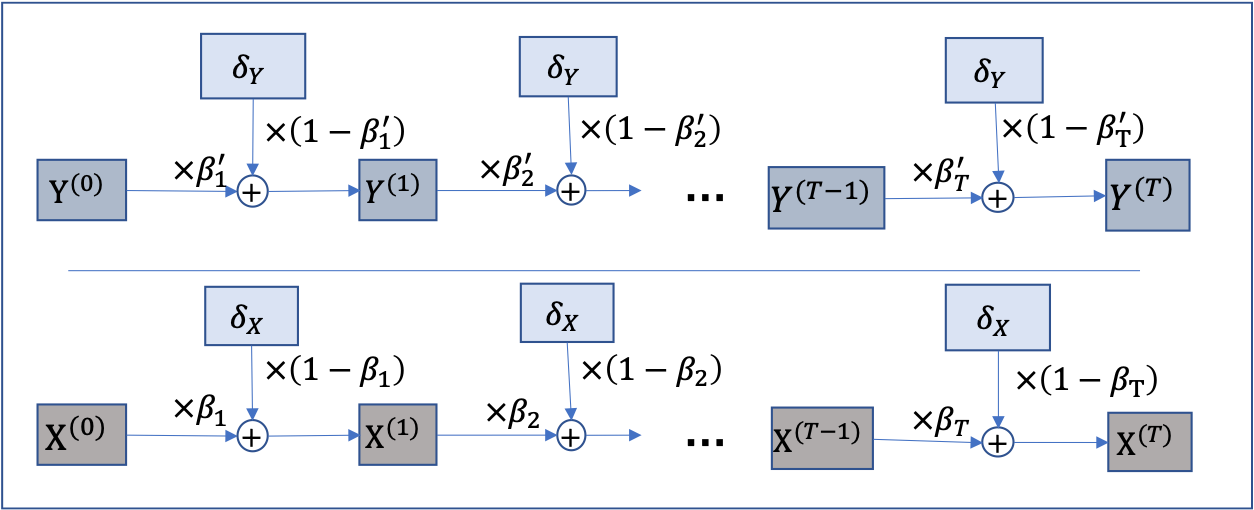}
    \caption{
    An illustration of the coupled diffusion process.
    The input $X^{(0)}$ and the corresponding target $Y^{(0)}$ are diffused simultaneously with different variance schedules.
    $\bm{\beta} = \{ \beta_1, \cdots, \beta_T \}$ is the variance schedule for the input and $\bm{\beta}^{\prime} = \{ \beta_1^{\prime},  \cdots, \beta_T^{\prime} \}$ is for the target.
    }   \label{fig:diffusion_process}
    \end{minipage}
\end{figure}

The forward diffusion process is fixed to a Markov chain that gradually adds Gaussian noise to the data \cite{sohl2015deep,ho2020denoising}.~%
To diffuse the input and output series, we propose a coupled diffusion process, which is demonstrated in \cref{fig:diffusion_process}.~%
Specifically, given the input  $X = X^{(0)} \sim q(X^{(0)})$, the approximate posterior $q(X^{(1:T)}|X^{(0)})$  can be obtained as 
\begin{equation}
    q(X^{(1:T)}|X^{(0)}) = \prod_{t=1}^{T} q(X^{(t)}|X^{(t-1)})\, , \quad 
    q(X^{(t)}|X^{(t-1)}) = \mathcal{N} (X^{(t)}; \sqrt{1 - \beta_t} X^{(t)}, \beta_t I) \, ,
\end{equation}
where a uniformly increasing 
variance schedule $ \bm{\beta} = \{\beta_1, \cdots, \beta_{T} \, | \, \beta_t \in [0, 1) \} $  is employed to control the level of noise to be added. 
Then, let $ \alpha_t = 1 - \beta_t$ and $\bar{\alpha}_t = \prod_{s=1}^{t}\alpha_s$, we have
\begin{equation}    \label{diff}
    q(X^{(t)}|X^{(0)}) = \mathcal{N} ( X^{(t)}; \sqrt{\bar{\alpha}_t} X^{(0)}, (1 - \bar{\alpha}_t) I) \, .
\end{equation}

Furthermore, according to \cref{prop1} we decompose $X^{(0)}$ as $X^{(0)} = \langle X_r, \epsilon_X \rangle$.~%
Then, with \cref{diff}, the diffused $X^{(t)}$ can be decomposed as follows: 
\begin{equation}    \label{x_t}
    X^{(t)} = \sqrt{\bar{\alpha}_t}X^{(0)} + (1-\bar{\alpha}_t)\delta_X 
    \vcentcolon = 
    \langle 
    \underbrace{\sqrt{\bar{\alpha}_t}X_r}_{\text{ideal}\,\, \text{part}}, \,
    \underbrace{\sqrt{\bar{\alpha}_t}\epsilon_X + (1-\bar{\alpha}_t)\delta_{X}}_{\text{noisy} \,\, \text{part}}
    \rangle 
    \, ,
\end{equation}
where $\delta_X$ denotes the standard Gaussian noise of $X$. 
As $\bm{\alpha}$ can be determined when the variance schedule $\bm{\beta}$ is known, the ideal part is also determined in the diffusion process. 
Let $\widetilde{X}_r^{(t)} = \sqrt{\bar{\alpha}_t}X_r$ and ${\delta}_{\widetilde{X}}^{(t)} = \sqrt{\bar{\alpha}_t} \epsilon_X + (1-\bar{\alpha}_t) \delta_X $,  
then, according to \cref{prop1} and \cref{x_t}, we have   
\begin{equation}    \label{eq:target}
    p_\phi (Z^{(t)} | X^{(t)}) = p_\phi (Z^{(t)}|\widetilde{X}_r^{(t)}, {\delta}_{\widetilde{X}}^{(t)}) 
\, , \quad 
    p_\theta (\widehat{Y}^{(t)}|Z^{(t)}) = p_\theta (\widehat{Y}_r^{(t)} | Z^{(t)})  p_\theta ({\delta}_{\widehat{Y}}^{(t)} | Z^{(t)})
    \, ,
\end{equation}
where $ \delta_{\widehat{Y}}^{(t)} $ denotes the generated noise of $\widehat{Y}^{(t)}$.~%
To relieve the effect of aleatoric uncertainty resulting from time series data, we further apply the  diffusion process to the target series $Y = Y^{(0)} \sim q(Y^{(0)})$. 
In particular, 
a scale parameter $\omega \in (0, 1)$ is adopted, such that 
$ {\beta}_t^{\prime} = \omega \beta_t, \alpha_t^{\prime} = 1 - {\beta}_t^{\prime}$ 
and 
$ \bar{\alpha}^{\prime}_{t} =  \prod_{s=1}^{t} \alpha_s^{\prime}$. 
Then, according to \cref{prop1}, we can obtain the following decomposition (similar to \cref{x_t}):
\begin{equation}    \label{eq:true}
    Y^{(t)} = \sqrt{\bar{\alpha}^{\prime}_t} Y^{(0)} + (1-\bar{\alpha}^{\prime}_t) \delta_Y 
    \vcentcolon = 
    \langle 
    \underbrace{\sqrt{\bar{\alpha}^{\prime}_t}Y_r}_{\text{ideal} \, \text{part}},
    \underbrace{\sqrt{\bar{\alpha}^{\prime}_t}\epsilon_{Y} + (1-\bar{\alpha}^{\prime}_t)\delta_Y}_{\text{noisy} \, \text{part}} 
    \rangle 
    = \langle \widetilde{Y}_r^{(t)}, {\delta}_{\widetilde{Y}}^{(t)} \rangle \, . 
\end{equation}
Consequently, we have 
$  q(Y^{(t)})  =  q(\widetilde{Y}_r^{(t)})  q({\delta}_{\widetilde{Y}}^{(t)}) $.
Afterward, we can draw the following conclusions with \cref{prop1} and \cref{eq:target,eq:true}. 
The proofs can be found in \cref{appendix:lemma:derivation}.

\begin{lemma}  \label{theo1}
$ \forall \varepsilon > 0 $, 
there exists a probabilistic model $ \, f_{\phi, \theta} \vcentcolon = (p_{\phi}, p_{\theta})$ 
to guarantee that $ \mathcal{D}_{\mathrm{KL}} (q(\widetilde{Y}_r^{(t)}) || p_{\theta}(\widehat{Y}_r^{(t)})) < \varepsilon $, 
where $ \widehat{Y}_r^{(t)} = f_{\phi,\theta} (X^{(t)})  $. 
\end{lemma}

\begin{lemma} \label{theo2}
With the coupled diffusion process, 
the difference between  diffusion noise and  generation noise will be reduced, 
i.e., 
$ \lim_{t \rightarrow \infty} \mathcal{D}_{\mathrm{KL}} (q ({\delta_{\widetilde{Y}}^{(t)}}) || p_{\theta} (\delta_{\widehat{Y}}^{(t)} | Z^{(t)}) ) < 
\mathcal{D}_{\mathrm{KL}} (q(\epsilon_Y) || p_{\theta}(\epsilon_{\widehat{Y}}))$ . 
\end{lemma}

Therefore, the uncertainty raised by the generative model and the inherent data noise can be reduced through the coupled diffusion process. 
In addition, the diffusion process simultaneously augments the input series and the target series, which can improve the generalization capability for (esp. short) time series forecasting.

\subsubsection{Bidirectional Variational Auto-Encoder}

Traditionally, in the diffusion model, a reverse process is adopted to generate high-quality samples~\cite{sohl2015deep,ho2020denoising}. 
However, for the generative time series forecasting problem, 
not only the expressiveness but also the supervision of the ground truths should be considered. 
In this work, 
we employ a more efficient generative model, i.e.,  bidirectional variational auto-encoder (BVAE)~\cite{vahdat2020nvae}, to take the place of the reverse process of the diffusion model.~%
The architecture of BVAE is described in~\cref{fig:method:framework} where $Z$ is treated in a multivariate fashion $Z = \{z_1, \cdots, z_n\}$ ($ z_i \in \mathbb{R}^{m}, z_i = [z_{i,1}, \cdots, z_{i,m}] $) and $z_{i+1} \sim p(z_{i+1}|z_{i}, X)$.
Then, $n$ is determined in accordance with the number of residual blocks in the encoder, as well as the decoder. 
Another merit of BVAE is that it opens an interface to integrate the disentanglement for improving model interpretability (refer to \cref{sec:disentangle}).

\subsection{Scaled Denoising Score Matching for Diffused Time Series Cleaning} \label{sec_denoise}

Although the time series data can be augmented with the aforementioned coupled diffusion probabilistic model, the generative distribution $ p_{\theta}(\widehat{Y}^{(t)}) $ tends to move toward the diffused target series $ Y^{(t)} $ which has been corrupted~\cite{li2019learning, song2019generative}. 
To further ``clean'' the generated target series, we employ the Denoising Score Matching (DSM) to accelerate the de-uncertainty process without sacrificing the model flexibility. 
DSM~\cite{vincent2011connection,li2019learning} was proposed to link Denoising Auto-Encoder (DAE)~\cite{vincent2010stacked} to Score Matching  (SM)~\cite{hyvarinen2009estimation}.
Let $ \widehat{Y} $ denote the generated target series, then we have the objective 
\begin{equation}
    L_{\text{DSM}}(\zeta) = \mathbb{E}_{ p_{\sigma_0} (\widehat{Y}, Y)} 
    \| \nabla_{\widehat{Y}} \log(q_{\sigma_0} (\widehat{Y}|Y)) + \nabla_{\widehat{Y}} E(\widehat{Y}; \zeta) \|^{2} \, ,
\end{equation}
where $p_{\sigma_0} (\widehat{Y}, Y)$ is the joint density of pairs of corrupted and clean samples $(\widehat{Y}, Y)$, 
$ \nabla_{\widehat{Y}} \log(q_{\sigma_0} (\widehat{Y}|Y)) $ is derivative of the log density of a single noise kernel, which is dedicated to replacing the Parzen density estimator: $ p_{\sigma_0} (\widehat{Y}) = \int q_{\sigma_0} (\widehat{Y}|Y) p(Y) dY$ in score matching, 
and $E(\widehat{Y}; \zeta)$ is the energy function. 
In the particular case of Gaussian noise, 
$\log(q_{\sigma_0} (\widehat{Y}|Y)) = - (\widehat{Y} - Y)^2/2 {\sigma_0}^2 + C$. 
Thus, we have 
\begin{equation}
    L_{\text{DSM}}(\zeta) = \mathbb{E}_{p_{\sigma_0}(\widehat{Y}, Y)} \| Y - \widehat{Y} + \sigma_0^2 \nabla_{\widehat{Y}} E(\widehat{Y}; \zeta) \|^{2} \, .
\end{equation}
Then, for the diffused target series at step $t$, we can obtain 
\begin{equation}
    L_{\text{DSM}}(\zeta, t) = \mathbb{E}_{p_{\sigma_0} (\widehat{Y}^{(t)}, Y)} \| Y - \widehat{Y}^{(t)} + \sigma_0^2 \nabla_{\widehat{Y}^{(t)}} E(\widehat{Y}^{(t)}; \zeta) \|^2 \, .
\end{equation}
To scale the noise of different levels~\cite{li2019learning}, a monotonically decreasing series of fixed $\sigma$ values $ \{ \sigma_1, \cdots, \sigma_T \,|\, \sigma_t = 1-\bar{\alpha}_t \}$ (refer to the aforementioned variance schedule $\bm{\beta}$ in \cref{sec2.2}) is adopted.
Therefore, the objective of the multi-scaled DSM is 
\begin{equation}    \label{eq:dsm}
    L(\zeta,t) = \mathbb{E}_{q_{\sigma}(\widehat{Y}^{(t)}|Y)p(Y)} 
    l(\sigma_t) 
    \| Y - \widehat{Y}^{(t)} + \sigma_0^2 \nabla_{\widehat{Y}^{(t)}} 
    E(\widehat{Y}^{(t)}; \zeta) \|^2
    \, ,
\end{equation}
where $ \sigma \in \{ \sigma_1, \cdots, \sigma_T \} $ and $l(\sigma_t) = \sigma_t$. 
With~\cref{eq:dsm}, we can ensure that the gradient has the right magnitude by setting $\sigma_0$.

In the generative time series forecasting setting, the generated samples will be tested without applying the diffusion process. 
To further denoise the generated target series $\widehat{Y}$, we apply a single-step gradient denoising jump~\cite{saremi2019neural}:
\begin{equation}	    \label{eq:clean}
    \widehat{Y}_{\text{clean}} = \widehat{Y} - \sigma_0^2 \nabla_{\widehat{Y}} E (\widehat{Y}; \zeta) \, .
\end{equation}
The generated results tend to possess a larger distribution space than the true target,  and the noisy term in~\cref{eq:clean} approximates the noise between the generated target series and the ``cleaned'' target series.
Therefore,  $\sigma_0^2 \nabla_{\widehat{Y}} E (\widehat{Y}; \zeta)$  can be treated as the estimated uncertainty of the prediction.

\subsection{Disentangling Latent Variables for Interpretation}     \label{sec:disentangle}

The interpretability of the time series forecasting model is of great importance for many downstream tasks~\cite{tonekaboni2020went, hardt2020explaining, ismail2020benchmarking}.
Through disentangling the latent variables of the generative model, not only the interpretability but also the reliability of the prediction can be further enhanced~\cite{li2021learning}.

To disentangle the latent variables $ Z = \{ z_1, \cdots, z_n \} $, we attempt to minimize the Total Correlation (TC)~\cite{watanabe1960information, kim2018disentangling}, which is a popular metric to measure  dependencies among multiple random variables, %
\begin{equation}
    \text{TC}(z_i) = \mathcal{D}_{\mathrm{KL}} (p_\phi (z_i) || \bar{p}_{\phi} (z_i)), 
    \qquad 
    \bar{p}_\phi (z_i) = \prod_{j=1}^{m} p_\phi (z_{i,j})
\end{equation}
%
where $m$ denotes the number of factors of $z_i$ that need to be disentangled. 
Lower TC generally means better disentanglement if the latent variables preserve useful information.
However, a very low TC can still be obtained when the latent variables carry no meaningful signals. 
Through the bidirectional structure of BVAE, such issues can be tackled without too much effort.
As shown in~\cref{fig:method:framework}, the signals are disseminated in both the encoder and decoder, such that rich semantics are aggregated into the latent variables. 
Furthermore, to alleviate the effect of potential irregular values, we average the total correlations of $z_{1:n}$, 
then the loss w.r.t. the TC  score of BVAE can be obtained:  
\begin{equation}	    \label{tc}
    L_{\text{TC}} = \frac{1}{n}\sum_{i=1}^{n}\text{TC}(z_i) \, .
\end{equation}


\par\smallskip\noindent
\centerline{
\begin{minipage}[t]{.8\textwidth}
\begin{algorithm}[H]
\caption{Training Procedure.} \label{alg_train}
\begin{algorithmic}[1]
  \small{
  \REPEAT 
  \STATE
  $ X^{(0)} \sim q(X^{(0)}), \quad Y^{(0)} \sim q(Y^{(0)}), \quad \delta_X \sim N(0, I_d), \quad \delta_Y \sim N(0, I_d)$ 
  \STATE 
  Randomly choose $t \in \{ 1, \cdots, T\}$ and with \cref{x_t,eq:true}, 
  \STATE 
  $\quad X^{(t)} = \sqrt{\bar{\alpha}_t}X^{(0)} + (1-\bar{\alpha}_t)\delta_X, 
   \quad  Y^{(t)} = \sqrt{\bar{\alpha}^{\prime}_{t} }Y^{(0)} + (1-\bar{\alpha}^{\prime}_{t})\delta_Y 
  $
  \STATE
  Generate the latent variable $Z$ with BVAE, $Z \sim p_\phi(Z|X^{(t)})$
  \STATE 
  Sample $\widehat{Y}^{(t)} \sim p_{\theta}(\widehat{Y}^{(t)}|Z)$ and calculate $\mathcal{D}_{\mathrm{KL}} (q(Y^{(t)})||p_\theta(\widehat{Y}^{(t)}))$ 
  \STATE 
  Calculate DSM loss with \cref{eq:dsm} 
  \STATE 
  Calculate total correlation of $Z$ with \cref{tc}
  \STATE 
  Construct the total loss $ \mathcal{L} $ with \cref{loss}
  \STATE 
  $\theta, \phi \leftarrow \mathrm{argmin} (\mathcal{L})$
  \UNTIL 
  Convergence
  }
\end{algorithmic} 
\end{algorithm}
\end{minipage}
}

\vspace{-3ex}

\par\smallskip\noindent
\centerline{
\begin{minipage}[t]{.65\textwidth}
\begin{algorithm}[H]
\caption{Forecasting Procedure.} \label{alg_pred}
\begin{algorithmic}[1]
\small{
  \STATE 
  \textbf{Input:} $X \sim q(X)$
  \STATE 
  Sample $Z \sim p_{\phi}(Z|X)$
  \STATE 
  Generate $\widehat{Y} \sim p_{\theta}(\widehat{Y}|Z)$
  \STATE 
  \textbf{Output:} $\widehat{Y}_{\text{clean}}$ and the estimated uncertainty with \cref{eq:clean}
}
\end{algorithmic} 
\end{algorithm}
\end{minipage}
}

\subsection{Training and Forecasting}

\textbf{Training Objective.} 
To reduce the effect of uncertainty, the coupled diffusion equipped with the denoising network is proposed without sacrificing generalizability. 
Then we disentangle the latent variables of the generative model by minimizing the TC of the latent variables. 
Finally, we reconstruct the loss with several trade-off parameters, 
and with \cref{eq:dsm,eq:clean,tc} we have
\begin{equation}    \label{loss}
\mathcal{L} 
= 
\psi \cdot  \mathcal{D}_{\mathrm{KL}} (q(Y^{(t)})||p_\theta(\widehat{Y}^{(t)})) 
+ 
\lambda \cdot {L} (\zeta, t) 
+ 
\gamma \cdot L_{\text{TC}} 
+ 
L_{\text{mse}} (\widehat{Y}^{(t)}, Y^{(t)}) 
\, ,
\end{equation}
where $L_{\text{mse}}$ calculates the mean square error (MSE) between $\widehat{Y}^{(t)}$ and $Y^{(t)}$.
We minimize the above objective to learn the generative model accordingly.

{\bf Algorithms.}
\cref{alg_train} displays the complete training procedure of \ourmodel~with the loss function in~\cref{loss}. 
For inference, as described in \cref{alg_pred}, given the input series $X$, the target series can be generated directly from the distribution $p_{\theta}$ which is conditioned on the latent states drawn from the distribution $p_{\phi}$.

\section{Experiments} \label{experiment}

\subsection{Experiment Settings}  \label{setting}

\textbf{Datasets.~}
We generate two synthetic datasets suggested by~\cite{farnoosh2020deep}, 
\begin{equation*}
\begin{gathered}
     w_t = a \cdot  w_{t-1} + \text{tanh}(b \cdot  w_{t-2}) + \text{sin}(w_{t-3}) + \mathcal{N}(0, 0.5I)\\
     X = [w_1, w_2, ..., w_N] \cdot  F + \mathcal{N}(0, 0.5I) \, ,
\end{gathered}
\end{equation*}
where $w_t \in \mathbb{R}^2$ and $ 0 \leq w_{t, 1}, w_{t, 2} \leq 1 $ ($t = 1,2,3$), 
$F \in \mathbb{R}^{2 \times k} \sim \mathcal{U}[-1, 1]$, $k$ denotes the dimensionality and $N$ is the number of time points, $a, b$ are two constants.~%
We set $a=0.9, b=0.2, k=20$ to generate D$_1$, and $a=0.5, b=0.5, k=40$ for D$_2$, and $N = 800$ for both D$_1$ and D$_2$.

\begin{table}[t]
    \caption{
    Performance comparisons on synthetic data in terms of MSE and CRPS. 
    The best results are boldfaced.
    }
    \centering
    \small
    \setlength\tabcolsep{2.5pt}
    \renewcommand{\arraystretch}{1.2}
    \begin{tabular}{c|c|ccccccccc}
    \toprule 
    \multicolumn{2}{c}{Model}& \ourmodel &NVAE &$\beta$-TCVAE &f-VAE &DeepAR &TimeGrad &GP-copula & VAE \\
   \midrule
    \multirow{4}{*}{D$_1$} & \multirow{2}{*}{8}&$\textbf{0.512}_{\pm.033}$ &$1.201_{\pm.027}$&$0.631_{\pm.003}$&$0.854_{\pm.099}$&$1.153_{\pm.125}$&$0.966_{\pm.102}$&$1.202_{\pm.108}$&$0.912_{\pm.132}$\\
    ~&~&$\textbf{0.585}_{\pm.021}$&$0.905_{\pm.011}$&$0.658_{\pm.002}$&$0.745_{\pm.036}$&$0.758_{\pm.038}$&$0.698_{\pm.024}$&$0.773_{\pm.033}$&$0.786_{\pm.053}$\\
    \cline{2-10}
   
    ~& \multirow{2}{*}{16}& $\textbf{0.571}_{\pm.025}$&$1.184_{\pm.025}$&$0.758_{\pm.047}$&$1.046_{\pm.270}$&$0.911_{\pm.046}$&$0.945_{\pm.315}$&$0.915_{\pm.059}$&$0.908_{\pm.177}$\\
    ~&~&$\textbf{0.625}_{\pm.013}$&$0.897_{\pm.012}$&$0.747_{\pm.027}$&$0.835_{\pm.108}$&$0.699_{\pm.014}$&$0.709_{\pm.100}$&$0.704_{\pm.020}$&$0.765_{\pm.067}$\\
    
    \midrule
    \multirow{4}{*}{D$_2$} &  \multirow{2}{*}{8} &$\textbf{0.599}_{\pm.049}$&$1.966_{\pm.047}$&$3.096_{\pm.197}$&$3.353_{\pm.430}$&$0.977_{\pm.137}$&$0.963_{\pm.385}$&$1.037_{\pm.082}$&$3.079_{\pm.345}$\\
    ~&~&$\textbf{0.628}_{\pm.027}$& $1.255_{\pm.021}$&$1.680_{\pm.062}$&$1.640_{\pm.154}$&$0.727_{\pm.058}$&$0.706_{\pm.123}$&$0.753_{\pm.026}$&$1.504_{\pm.098}$\\
    \cline{2-10}
    ~& \multirow{2}{*}{16}& $\textbf{0.786}_{\pm.041}$&$1.955_{\pm.051}$&$3.067_{\pm.443}$&$3.109_{\pm.428}$&$0.972_{\pm.144}$&$0.850_{\pm.061}$&$1.082_{\pm.071}$&$3.132_{\pm.160}$\\
    
    ~&~&$0.728_{\pm.026}$ &$1.251_{\pm.020}$&$1.643_{\pm.183}$&$1.558_{\pm.157}$&$0.720_{\pm.050}$&$\textbf{0.649}_{\pm.017}$&$0.762_{\pm.008}$&$1.560_{\pm.060}$\\
    \bottomrule
    \end{tabular}
    \label{toy1}
\vspace{-2ex}
\end{table}

\begin{table}[t]
  \caption{
  The performance comparisons on real-world datasets in terms of MSE and CRPS, and the best results are in boldface.
  }
  \label{real}
    \centering
    \small
    \setlength\tabcolsep{2.0pt}
    \renewcommand{\arraystretch}{1.2}
    \begin{threeparttable}
    \begin{tabular}{c|c|ccccccccc}
    \toprule  
    \multicolumn{2}{c}{Model}& \ourmodel &NVAE &$\beta$-TCVAE &f-VAE & DeepAR & TimeGrad & GP-copula & VAE \\
   \midrule
    \multirow{4}{*}{\rotatebox{90}{Traffic}} & \multirow{2}{*}{8}&$\textbf{0.081}_{\pm.003}$&$1.300_{\pm.024}$&$1.003_{\pm.006}$&$0.982_{\pm.059}$&$3.895_{\pm.306}$&$3.695_{\pm.246}$&$4.299_{\pm.372}$&$0.794_{\pm.130}$\\
    ~&~&$\textbf{0.207}_{\pm.003}$&$0.593_{\pm.004}$&$0.894_{\pm.003}$&$0.666_{\pm.032}$&$1.391_{\pm.071}$&$1.410_{\pm.027}$&$1.408_{\pm.046}$&$0.759_{\pm.07}$\\
    \cline{2-10}
    ~& \multirow{2}{*}{16}&$\textbf{0.081}_{\pm.009}$&$1.271_{\pm.019}$&$0.997_{\pm.004}$&$0.998_{\pm.042}$&$4.141_{\pm.320}$&$3.495_{\pm.362}$&$4.575_{\pm.141}$&$0.632_{\pm.057}$\\
    ~&~&$\textbf{0.200}_{\pm.014}$&$0.589_{\pm.001}$&$0.893_{\pm.002}$&$0.692_{\pm.026}$&$1.338_{\pm.043}$&$1.329_{\pm.057}$&$1.506_{\pm.025}$&$0.671_{\pm.038}$\\
     \midrule
    \multirow{4}{*}{\rotatebox{90}{Electricity}} &  \multirow{2}{*}{8}&$\textbf{0.251}_{\pm.015}$&$1.134_{\pm.029}$&$0.901_{\pm.052}$&$0.893_{\pm.069}$&$2.934_{\pm.173}$&$2.703_{\pm.087}$&$2.924_{\pm.218}$&$0.853_{\pm.040}$\\
    ~&~&$\textbf{0.398}_{\pm.011}$&$0.542_{\pm.003}$&$0.831_{\pm.004}$&$0.809_{\pm.024}$&$1.244_{\pm.037}$&$1.208_{\pm.024}$&$1.249_{\pm.048}$&$0.795_{\pm.016}$\\
    \cline{2-10}
    ~& \multirow{2}{*}{16}&$\textbf{0.308}_{\pm.030}$&$1.150_{\pm.032}$&$0.850_{\pm.003}$&$0.807_{\pm.034}$&$2.803_{\pm.199}$&$2.770_{\pm.237}$&$3.065_{\pm.186}$&$0.846_{\pm.062}$\\
    ~&~&$\textbf{0.437}_{\pm.020}$&$0.531_{\pm.003}$&$0.814_{\pm.002}$&$0.782_{\pm.024}$&$1.220_{\pm.048}$&$1.240_{\pm.048}$&$1.307_{\pm.042}$&$0.793_{\pm.029}$\\
     \midrule
    \multirow{4}{*}{\rotatebox{90}{Weather}} & \multirow{2}{*}{8} & $\textbf{0.169}_{\pm.022}$&$0.801_{\pm.024}$&$0.234_{\pm.042}$&$0.591_{\pm.198}$&$2.317_{\pm.357}$&$2.715_{\pm.189}$&$2.412_{\pm.761}$&$0.560_{\pm.192}$\\
    ~&~&$\textbf{0.357}_{\pm.024}$&$0.757_{\pm.013}$&$0.404_{\pm.040}$&$0.565_{\pm.080}$&$0.858_{\pm.078}$&$0.920_{\pm.013}$&$0.897_{\pm.115}$&$0.572_{\pm.077}$\\
     \cline{2-10}
    ~& \multirow{2}{*}{16}&$\textbf{0.187}_{\pm.047}$&$0.811_{\pm.016}$&$0.212_{\pm.012}$&$0.530_{\pm.167}$&$1.269_{\pm.187}$&$1.110_{\pm.083}$&$1.357_{\pm.145}$&$0.424_{\pm.141}$\\
    ~&~&$\textbf{0.361}_{\pm.046}$&$0.759_{\pm.009}$&$0.388_{\pm.014}$&$0.547_{\pm.067}$&$0.783_{\pm.059}$&$0.733_{\pm.016}$&$0.811_{\pm.032}$&$0.503_{\pm.068}$\\
     \midrule
    \multirow{4}{*}{\rotatebox{90}{ETTm1}} &  \multirow{2}{*}{8} & $\textbf{0.527}_{\pm.073}$&$0.921_{\pm.026}$&$1.538_{\pm.254}$&$2.326_{\pm.445}$&$2.204_{\pm.420}$&$1.877_{\pm.245}$&$2.024_{\pm.143}$&$2.375_{\pm.405}$\\
    ~&~&$\textbf{0.557}_{0.048}$&$0.760_{\pm.026}$&$1.015_{\pm.112}$&$1.260_{\pm.167}$&$0.984_{\pm.074}$&$0.908_{\pm.038}$&$0.961_{\pm.027}$&$1.258_{\pm.104}$\\
    \cline{2-10}
    ~& \multirow{2}{*}{16}&$\textbf{0.968}_{\pm.104}$&$1.100_{\pm.032}$&$1.744_{\pm.100}$&$2.339_{\pm.270}$&$2.350_{\pm.170}$&$2.032_{\pm.234}$&$2.486_{\pm.207}$&$2.321_{\pm.469}$\\
    ~&~&$\textbf{0.821}_{\pm.072}$&$0.822_{\pm.026}$&$1.104_{\pm.041}$&$1.249_{\pm.088}$&$0.974_{\pm.016}$&$0.919_{\pm.031}$&$0.984_{\pm.016}$&$1.259_{\pm.132}$\\
     \midrule
    \multirow{4}{*}{\rotatebox{90}{ETTh1}} & \multirow{2}{*}{8} &$\textbf{0.292}_{\pm.036}$&$0.483_{\pm.017}$&$0.703_{\pm.054}$&$0.870_{\pm.134}$&$3.451_{\pm.335}$&$4.259_{\pm1.13}$&$4.278_{\pm1.12}$&$1.006_{\pm.281}$\\
    ~&~&$\textbf{0.424}_{\pm.033}$&$0.461_{\pm.011}$&$0.644_{\pm.038}$&$0.730_{\pm.060}$&$1.194_{\pm.034}$&$1.092_{\pm.028}$&$1.169_{\pm.055}$&$0.762_{\pm.115}$\\
    \cline{2-10}
    ~& \multirow{2}{*}{16}&$\textbf{0.374}_{\pm.061}$&$0.488_{\pm.010}$&$0.681_{\pm.018}$&$0.983_{\pm.139}$&$1.929_{\pm.105}$&$1.332_{\pm.125}$&$1.701_{\pm.088}$&$0.681_{\pm.104}$\\
    ~&~&$0.488_{\pm.039}$&$ \textbf{0.463}_{\pm.018}$&$0.640_{\pm.008}$&$0.760_{\pm.062}$&$1.029_{\pm.030}$&$0.879_{\pm.037}$&$0.999_{\pm.023}$&$0.641_{\pm.055}$\\
    \midrule
    \multirow{4}{*}{\rotatebox{90}{Wind}} & \multirow{2}{*}{8} & $\textbf{0.681}_{\pm.075}$&$1.854_{\pm.032}$&$1.321_{\pm.379}$&$1.942_{\pm.101}$&$12.53_{\pm2.25}$&$12.67_{\pm1.75}$&$11.35_{\pm6.61}$&$2.006_{\pm.145}$\\
    ~&~&$\textbf{0.596}_{\pm.052}$&$1.223_{\pm.014}$&$0.863_{\pm.143}$&$1.067_{\pm.086}$&$1.370_{\pm.107}$&$1.440_{\pm.059}$&$1.305_{\pm.369}$&$1.103_{\pm.100}$\\
     \cline{2-10}
    ~& \multirow{2}{*}{16}&$1.033_{\pm.062}$&$1.955_{\pm.015}$&$\textbf{0.894}_{\pm.038}$&$1.262_{\pm.178}$&$13.96_{\pm.1.53}$&$12.86_{\pm2.60}$&$13.79_{\pm5.37}$&$1.138_{\pm.205}$\\
    ~&~&$\textbf{0.757}_{\pm.053}$&$1.247_{\pm.011}$&$0.785_{\pm.037}$&$0.843_{\pm.066}$&$1.347_{\pm.060}$&$1.240_{\pm.070}$&$1.261_{\pm.171}$&$0.862_{\pm.092}$\\
    \bottomrule
    \end{tabular}
    \end{threeparttable}
\vspace{-2ex}
\end{table}

Six real-world datasets with diverse spatiotemporal dynamics are selected, 
including Traffic~\cite{lai2018modeling}, 
Electricity\footnote{\url{https://archive.ics.uci.edu/ml/datasets/ElectricityLoadDiagrams20112014}}, 
Weather\footnote{\url{https://www.bgc-jena.mpg.de/wetter/}}, 
Wind (Wind Power) \footnote{This dataset is published at 
\url{https://github.com/PaddlePaddle/PaddleSpatial/tree/main/paddlespatial/datasets/WindPower}.
},
and ETTs~\cite{zhou2021informer} (ETTm1 and ETTh1). 
To highlight the uncertainty in short time series scenarios, for each dataset, we slice a subset from the starting point to make sure that each sliced dataset contains at most 1000 time points.~%
Subsequently, we obtained 
$5\%$-Traffic, $3\%$-Electricity, $2\%$-Weather, $2\%$-Wind, $1\%$-ETTm1, and $5\%$-ETTh1.~%
The statistical descriptions of the real-world datasets can be found in \cref{exp_repro}.~%
All datasets are split chronologically and adopt the same train/validation/test ratios, i.e., 7:1:2.

\textbf{Baselines.~}
We compare \ourmodel ~with one GP (Gaussian Process) based method 
(GP-copula~\cite{salinas2019high}), two auto-regressive methods (DeepAR~\cite{salinas2020deepar} and TimeGrad~\cite{rasul2021autoregressive}), 
and four VAE-based methods, i.e., vanilla VAE, NVAE~\cite{vahdat2020nvae}, factor-VAE (f-VAE for short)~\cite{kim2018disentangling} and $\beta$-TCVAE~\cite{chen2018isolating}.

\textbf{Implementation Details.~}
An input-$l_x$-predict-$l_y$ window is applied to roll the train, validation, and test sets with stride one time-step, respectively, and this setting is adopted for all datasets. 
Hereinafter, the last dimension of the multivariate time series is selected as the target variable by default.

We use the Adam optimizer with an initial learning rate of $5e-4$.
The batch size is $16$, and the training is set to $20$ epochs at most equipped with early stopping. 
The number of disentanglement factors is chosen from $\{4, 8\}$, 
and $\beta_t \in \bm{\beta} $ is set to range from $0.01$ to $0.1$ with different diffusion steps $T \in [100, 1000]$, 
then $\omega$ is set to $0.1$. 
The trade-off hyperparameters are set as $\psi = 0.05, \lambda=0.1, \gamma=0.001$ for ETTs, and $\psi = 0.5, \lambda=1.0, \gamma=0.01$ for others. 
All the experiments were carried out on a Linux machine with a single NVIDIA P40 GPU.
The experiments are repeated five times, and the average and variance of the predictions are reported. 
We use the Continuous Ranked Probability Score~(CRPS)~\cite{matheson1976scoring} and Mean Squared Error~(MSE) as the evaluation metrics.
For both metrics, the lower, the better. 
In particular, CRPS is used to evaluate the similarity of two distributions and is equivalent to Mean Absolute Error (MAE) when two distributions are discrete.

\subsection{Main Results} \label{e1}

Two different prediction lengths, i.e., $l_y \in \{8, 16\}$ ($l_x = l_y$), are evaluated.
The results of longer prediction lengths are available in~\cref{extra}.

\textbf{Toy Datasets.~~}
In~\cref{toy1}, we can observe that 
\ourmodel ~achieves SOTA performance most of the time, and achieves competitive CRPS in D$_2$ for prediction length 16. 
Besides, VAEs outperform VARs and GP on D$_1$, 
but VARs achieve better performance on D$_2$, which demonstrates the advantage of VARs in learning complex temporal dependencies.

\textbf{Real-World Datasets.~~} 
As for the experiments on real-world data, \ourmodel ~achieves consistent  SOTA performance except for the prediction length 16 on the Wind dataset (\cref{real}). 
Particularly, under the input-8-predict-8 setting, \ourmodel~can provide remarkable improvements in Traffic, Electricity, Wind, ETTm1, ETTh1 and Weather w.r.t. MSE reduction (90\%, 71\%, 48\%, 43\%, 40\% and 28\%). 
Regarding the CRPS reduction, \ourmodel~achieves a 73\% reduction in Traffic, 31\% in Wind, and 27\% in Electricity under the input-8-predict-8 setting, 
and a 70\% reduction in Traffic, 18\% in Electricity, and 7\% in Weather under the input-16-predict-16 setting.~%
Overall, \ourmodel~gains the averaged 43\% MSE reduction and 23\% CRPS reduction among the above settings.~%
More results under longer prediction-length settings and on full datasets can be found in  \cref{appendix:supp-main-results}.

\textbf{Uncertainty Estimation.~~} 
The  uncertainty can be assessed by estimating the noise of the outcome series when doing the prediction~(see~\cref{sec_denoise}).  
Through scale parameter $\omega$, the generated distribution space can be adjusted accordingly (results on the effect of $\omega$ can be found  in Appendix D.3). 
The showcases in \cref{noise} demonstrate the uncertainty estimation of the yielded series in the Traffic dataset, where the last six dimensions are treated as target variables. 
We can find that noise estimation can quantify the uncertainty effectively. 
For example, the estimated uncertainty grows rapidly when extreme values are encountered.

\begin{figure}[htbp]
    \centering
    \begin{subfigure}[t]{0.31\textwidth}
      \includegraphics[width=\textwidth]{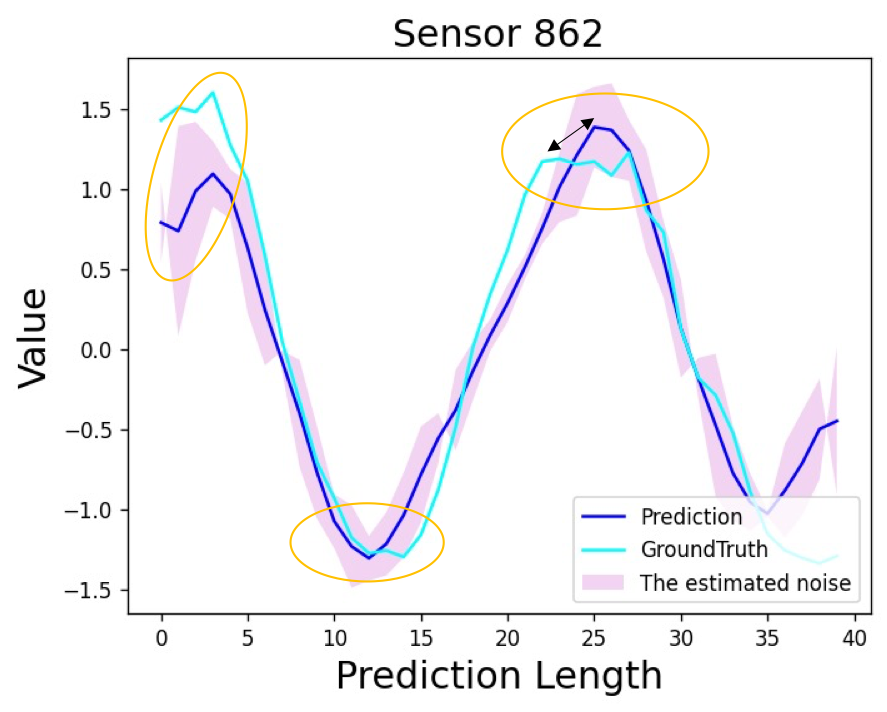}
    \end{subfigure}
    \begin{subfigure}[t]{0.31\textwidth}
      \includegraphics[width=\textwidth]{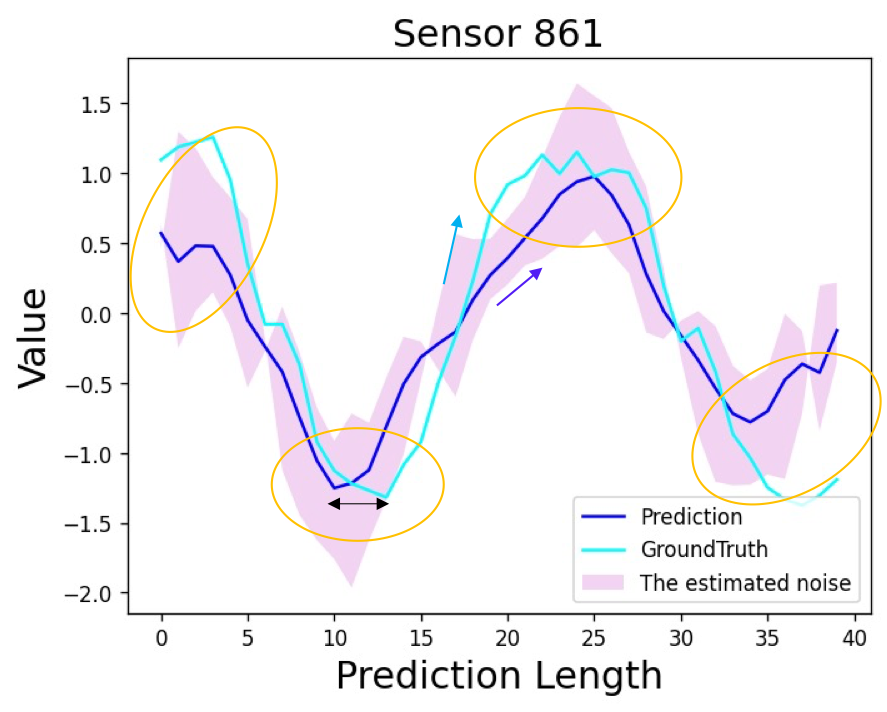}
    \end{subfigure}
    \begin{subfigure}[t]{0.31\textwidth}
      \includegraphics[width=\textwidth]{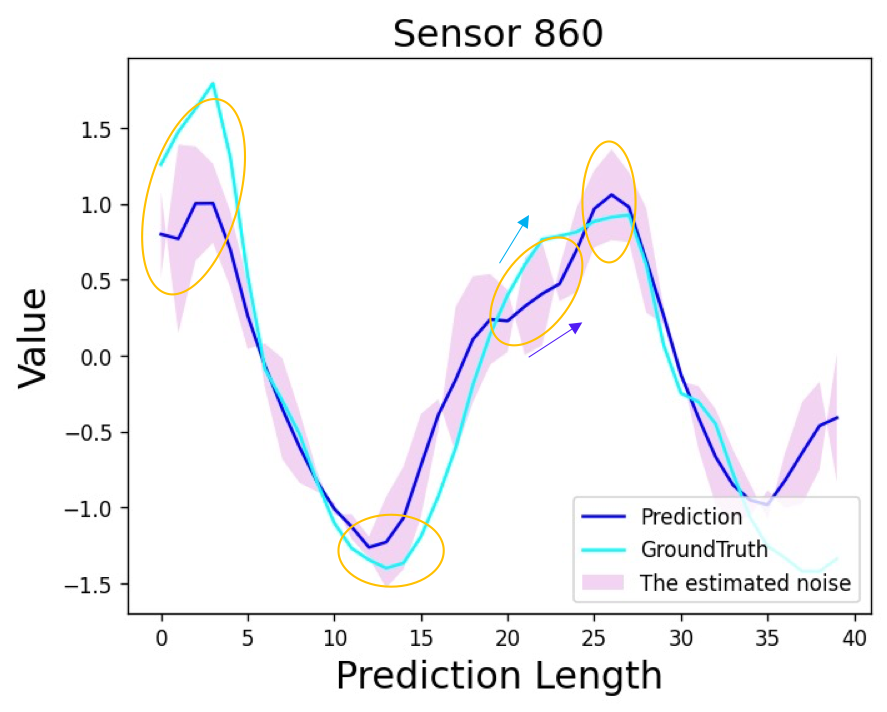}
    \end{subfigure}
    
    \begin{subfigure}[t]{0.31\textwidth}
      \includegraphics[width=\textwidth]{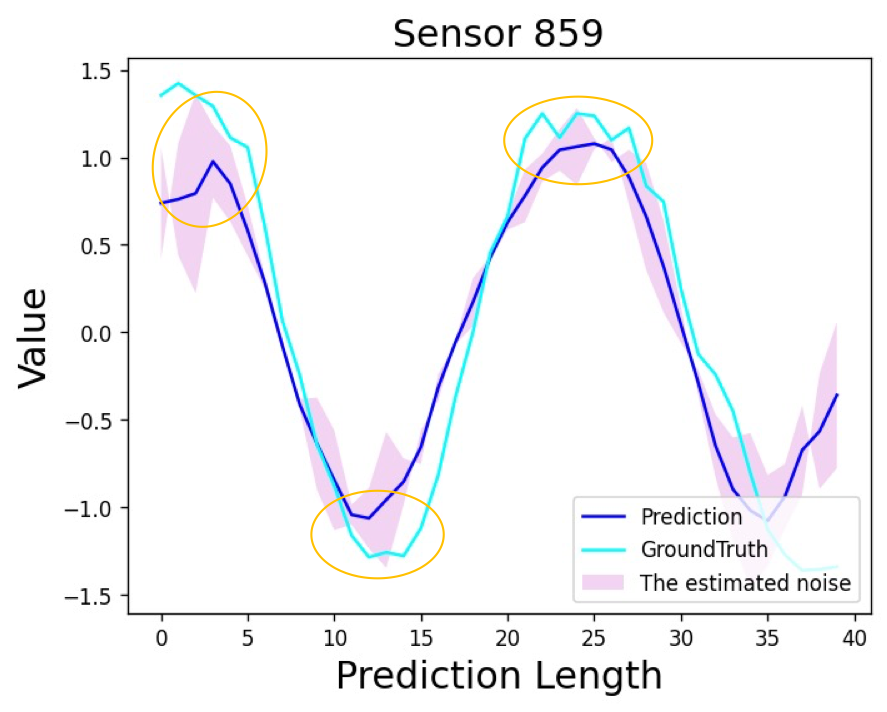}
    \end{subfigure}
    \begin{subfigure}[t]{0.31\textwidth}
      \includegraphics[width=\textwidth]{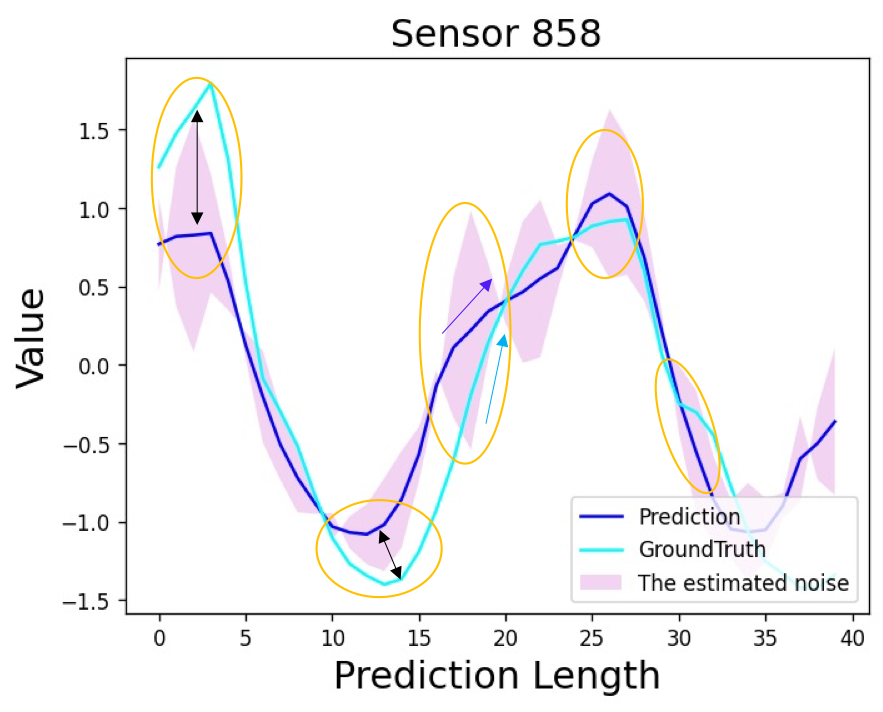}
    \end{subfigure}
    \begin{subfigure}[t]{0.31\textwidth}
      \includegraphics[width=\textwidth]{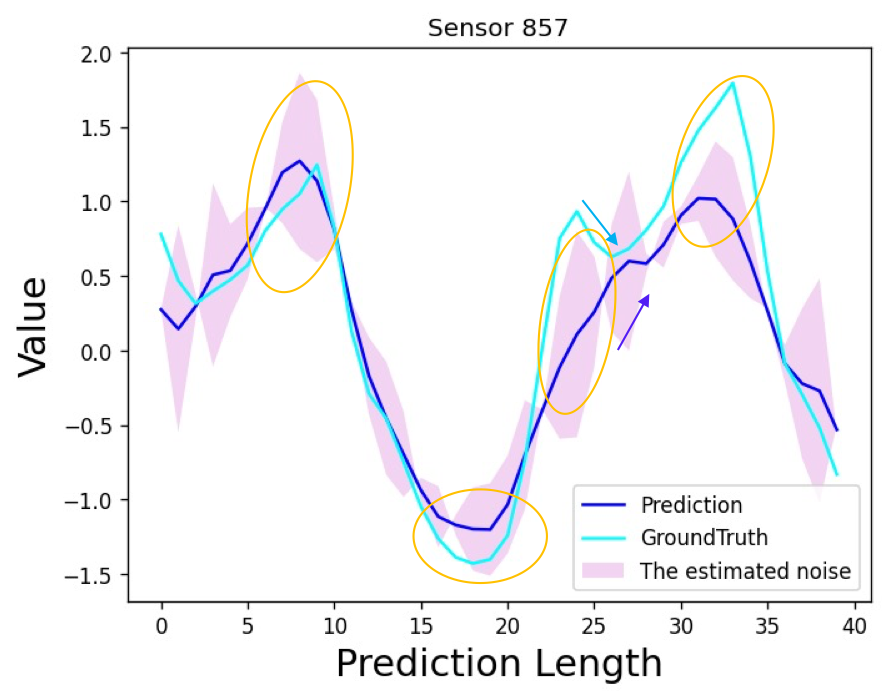}
    \end{subfigure}
    \caption{
    Uncertainty estimation of the prediction of the last six dimensions in the Traffic dataset and the colored envelope denotes the estimated uncertainty.
    }
    \label{noise}
\end{figure}

\textbf{Disentanglement Evaluation.~} 
For time series forecasting,
it is difficult to label disentangled factors by hand, 
thus we take different dimensions of $Z$ as the factors to be disentangled: $ z_i = [ z_{i, 1}, \cdots, z_{i, m}] $ ($z_i \in Z $). 
We build a classifier to discriminate whether an instance $z_{i,j}$ belongs to class $j$ such that the disentanglement quality can be assessed by evaluating the classification performance.  
Besides, 
we adopt the Mutual Information Gap (MIG) \cite{chen2018isolating} as a metric to evaluate the disentanglement more straightforwardly. 
Due to the space limit, the evaluation of disentanglement with different factors can be found in~\cref{disentangle}.

\begin{table}[t]
   \begin{minipage}[b]{0.63\textwidth}
    \centering
    \small
    \caption{
    Ablation study of the coupled diffusion probabilistic model w.r.t. MSE and CSPR. 
    } \label{tab:abla:diffusion}
    \renewcommand{\arraystretch}{1.0}
    \setlength\tabcolsep{0.8pt}
    \begin{threeparttable}
    \begin{tabular}{c|cc|cc}
    \toprule
     \multirow{2}{*}{Dataset}& \multicolumn{2}{c|}{Traffic} & \multicolumn{2}{c}{Electricity}\\
     \cline{2-5}
     ~&16&32&16&32\\
    \midrule
    \multirow{2}{*}{\ourmodel $_{-\widetilde{Y}}$} & 
    $0.122_{\pm.006}$&$0.126_{\pm.013}$&
    $0.350_{\pm.043}$&$0.422_{\pm.012}$\\
    ~& 
    $0.250_{\pm.008}$&$0.261_{\pm.017}$&
    $0.480_{\pm.032}$&$0.551_{\pm.012}$\\
    \midrule
    \multirow{2}{*}{\ourmodel $_{-\widetilde{Y}-\mathrm{DSM}}$} & 
    $0.096_{\pm.006}$& $0.092_{\pm.008}$ & $ 0.331_{\pm.023}$&$0.502_{\pm.079} $\\
    ~& $0.217_{\pm.010}$ & $0.220_{\pm.013}$& $ 0.450_{\pm.021}$&$0.584_{\pm.053}$\\
    \midrule
    \multirow{2}{*}{\ourmodel $_{-\widetilde{X}}$}&
    $0.123_{\pm.003}$&$0.117_{\pm.007}$&
    $0.351_{\pm.047}$&$0.420_{\pm.056}$\\
    ~&
    $0.256_{\pm.006}$&$0.253_{\pm.013}$&
    $0.481_{\pm.036}$&$0.540_{\pm.046}$\\
    \midrule
    \multirow{2}{*}{\ourmodel $_{-\mathrm{CDM}}$}&
    $0.123_{\pm.004}$&$0.118_{\pm.008}$&
    $0.365_{\pm.025}$&$0.439_{\pm.014}$\\
    ~&
    $0.255_{\pm.007}$&$0.252_{\pm.015}$&
    $0.498_{\pm.018}$&$0.561_{\pm.016}$\\
    \midrule
    \multirow{2}{*}{\ourmodel $_{-\mathrm{CDM}-\mathrm{DSM}}$}&
    $0.123_{\pm.003}$&$0.119_{\pm.003}$&
    $0.338_{\pm.041}$&$0.448_{\pm.062}$\\
    ~&
    $0.255_{\pm.003}$&$0.253_{\pm.005}$&
    $0.467_{\pm.029}$&$0.555_{\pm.041}$\\
    \midrule
    \multirow{2}{*}{\ourmodel}&$\textbf{0.081}_{\pm.009}$&$\textbf{0.091}_{\pm.007}$&$\textbf{0.308}_{\pm.030}$&$\textbf{0.410}_{\pm.075}$\\
    ~&$\textbf{0.200}_{\pm.014}$&$\textbf{0.216}_{\pm.012}$&$\textbf{0.437}_{\pm.020}$&$\textbf{0.534}_{\pm.058}$\\
    \bottomrule
    \end{tabular}
    \end{threeparttable}
    \label{tab:my_label}
    \end{minipage}
\hfill
\begin{minipage}[p]{0.35\textwidth}
    \centering
    \begin{subfigure}[t]{\textwidth}
     \includegraphics[width=0.9\textwidth,height=0.54\textwidth]{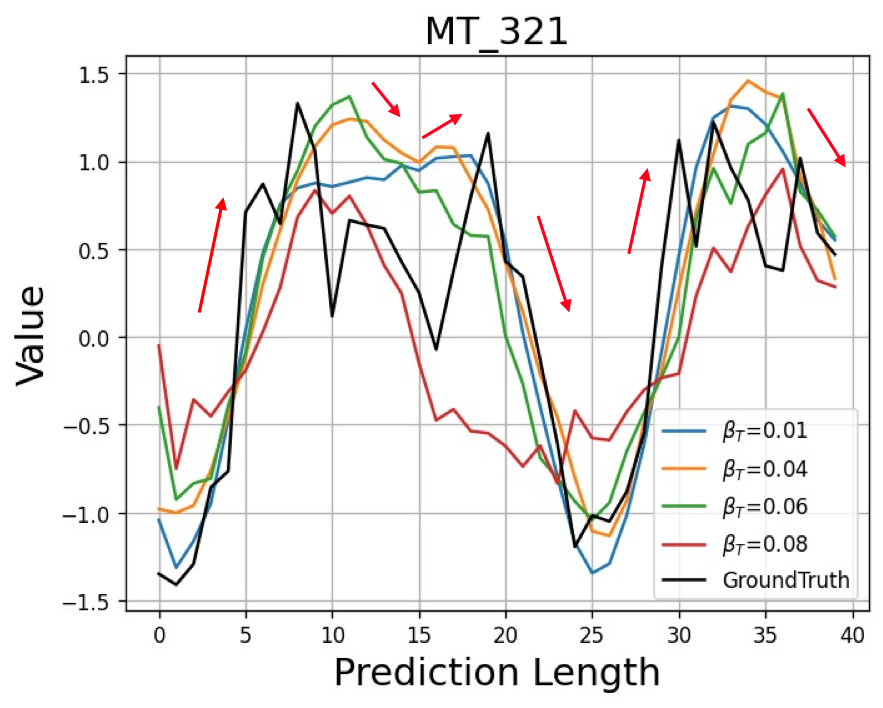}
    \end{subfigure}
    \vspace{1ex}
     \begin{subfigure}[t]{\textwidth}
     \includegraphics[width=0.9\textwidth,height=0.54\textwidth]{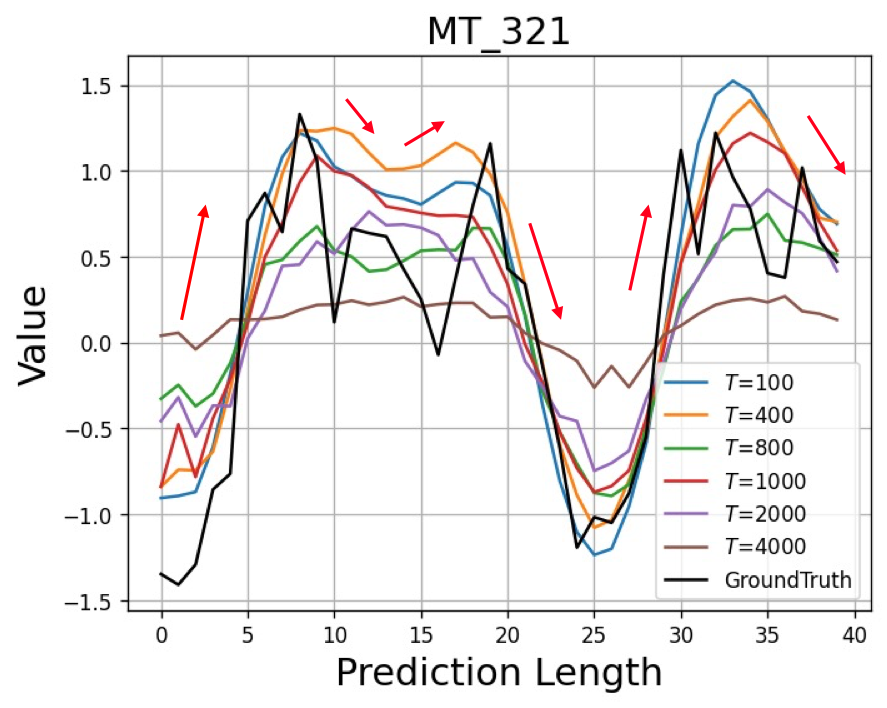}
    \end{subfigure}
    \captionof{figure}{
      Comparisons of predictions with different $\beta_{T}$ and varying $T$ on the Electricity dataset.
    }
    \label{fig:beta}
\end{minipage}
\vspace{-3ex}
\end{table}

\subsection{Model Analysis} \label{ablation}

\textbf{Ablation Study of the Coupled Diffusion and Denoising Network.} 
To evaluate the effectiveness of the coupled diffusion model (CDM), we compare the full versioned \ourmodel ~with its three variants: 
i) \ourmodel $_{-\widetilde{Y}}$, i.e. \ourmodel ~without diffused $Y$, 
ii) \ourmodel $_{-\widetilde{X}}$, i.e. \ourmodel ~without diffused $X$, 
and 
iii) \ourmodel $_{-\mathrm{CDM}}$, i.e. \ourmodel ~without any diffusion. 
Besides, the performance of \ourmodel~without denoising score matching (DSM) is also reported when the target series is not diffused, which are denoted as \ourmodel $_{-\widetilde{Y}-\mathrm{DSM}}$ and \ourmodel $_{-\mathrm{CDM}-\mathrm{DSM}}$. 
The ablation study is carried out on Traffic and Electricity datasets under input-16-predict-16 and input-32-predict-32. 
In \cref{tab:abla:diffusion},  we can find that the diffusion process can effectively augment the input or the target. 
Moreover, when the target is not diffused, the denoising network would be deficient since the noise level of the target cannot be estimated by then.

\textbf{Variance Schedule $\bm{\beta}$ and The Number of Diffusion Steps $T$.} 
To reduce the effect of the uncertainty while preserving the informative temporal patterns, the extent of the diffusion should be configured properly.~%
Too small a variance schedule or inadequate diffusion steps will lead to a meaningless diffusion process. Otherwise, the diffusion could be out of control {\footnote{
An illustrative showcase can be found in Appendix F. 
}}.~%
Here we analyze the effect of the variance schedule $\bm{\beta}$ and the number of diffusion steps $T$.  
We set $\beta_1 = 0$ and change the value of $\beta_t$ in the range of $[0.01, 0.1]$, and $T$ ranges from $100$ to $4000$.
As shown in~\cref{fig:beta}, we can  see  that the prediction performance can be improved if proper $\bm{\beta}$ and $T$ are employed.

\section{Discussion} \label{discuss}

\textbf{Sampling for Generative Time Series Forecasting.~}\\
The Langevin dynamics has been widely applied to the sampling of energy-based models~(EBMs)~\cite{xie2016theory,du2019implicit,xie2019learning},
\begin{equation}
    Y_k = Y_{k-1} - \frac{\rho}{2}\nabla_{Y}E_{\phi}(Y_{k-1})  + \rho^{\frac{1}{2}} \mathcal{N} (0, I_d) \, ,
\end{equation}
where $k \in \{ 0, \cdots, K \} $, 
$K$ denotes the number of sampling steps, 
and $\rho$ is a constant. 
With $K$ and $\rho$ being properly configured, high-quality samples can be generated.  
The Langevin dynamics has been successfully applied to applications in computer vision~\cite{kumar2019maximum,xie2021generative}, and natural language processing~\cite{deng2020residual}.

\begin{figure}[ht]
    \centering
      \begin{subfigure}[t]{0.24\textwidth}
      \centering
       \includegraphics[width=0.96\textwidth]{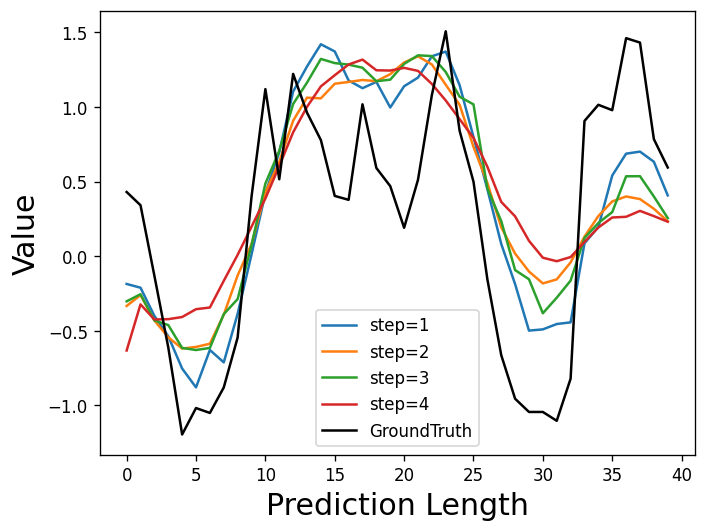}
       \caption{
       Multi-step denoising.
       }
       \label{steps}
      \end{subfigure}
      \begin{subfigure}[t]{0.24\textwidth}
      \centering
       \includegraphics[width=0.96\textwidth]{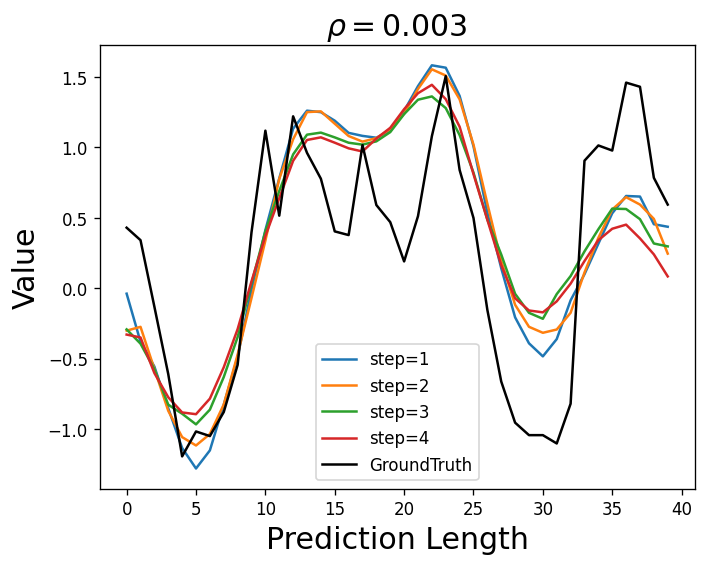}
       \caption{$\rho = 0.003$.}
       \label{rho=0.003}
      \end{subfigure}
     \begin{subfigure}[t]{0.24\textwidth}
        \centering
       \includegraphics[width=0.96\textwidth]{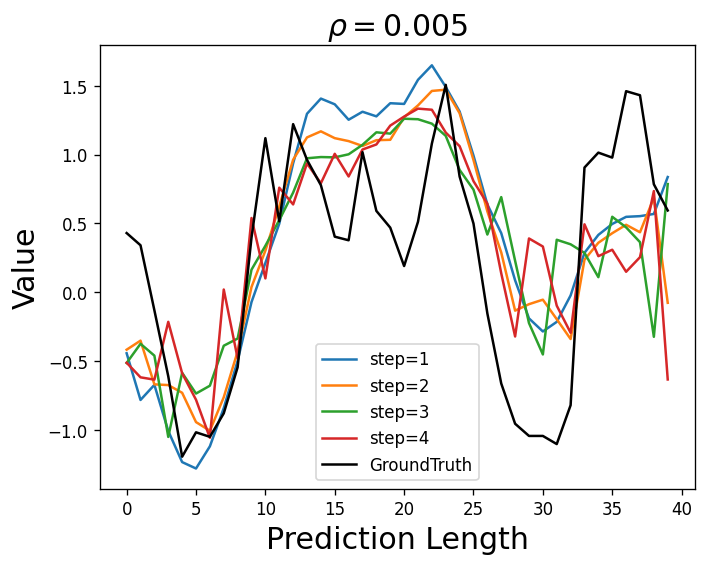}
       \caption{$\rho = 0.005$.}
       \label{rho=0.005}
      \end{subfigure}
      \begin{subfigure}[t]{0.24\textwidth}
      \centering
       \includegraphics[width=0.96\textwidth]{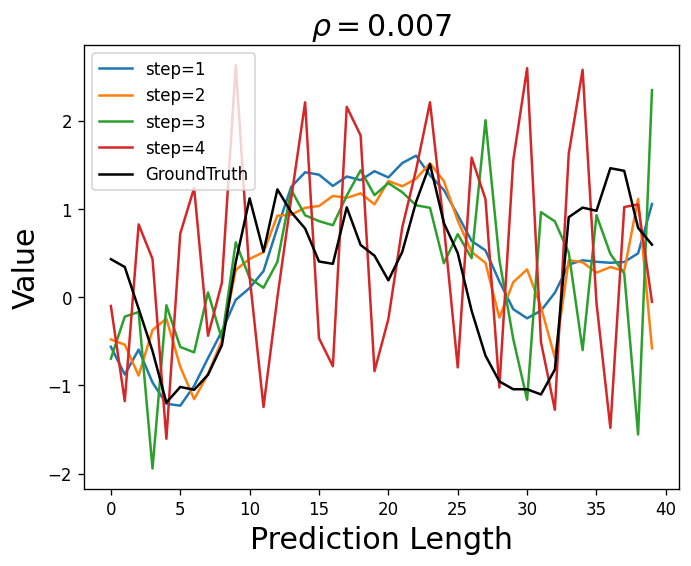}
       \caption{$\rho = 0.007$.}
       \label{rho=0.007}
      \end{subfigure}
      \caption{
      The prediction showcases in the Electricity dataset with different sampling strategies.
      }
     \label{fig:sample}
\end{figure}

We employ a single-step gradient denoising jump in this work to generate the target series. 
The experiments that were carried out demonstrate the effectiveness of such single-step sampling. 
We conduct an extra empirical study to investigate whether it is worth taking more sampling steps for further performance improvement of time series forecasting. 
We showcase the prediction results under different sampling strategies in \cref{fig:sample}. 
By omitting the additive noise in Langevin dynamics, we employ the multi-step denoising for \ourmodel~to generate the target series and plot the generated results in \cref{steps}. 
Then, with the standard Langevin dynamics, we can implement a generative procedure instead of denoising and compare the generated target series with different $\rho$ (see~\cref{rho=0.003,rho=0.005,rho=0.007}). 
%
We can observe that more sampling steps might not be helpful in improving prediction performance for generative time series forecasting~(\cref{steps}). 
Besides, larger sampling steps would lead to high computational complexity. 
On the other hand, different configurations of Langevin dynamics (with varying $\rho$) cannot bring indispensable benefits for time series forecasting (\cref{rho=0.003,rho=0.005,rho=0.007}).

\textbf{Limitations.~} 
\\
With the coupled diffusion probabilistic model, although the aleatoric uncertainty of the time series can be reduced, a new bias is brought into the series to mimic the distribution of the input and target.
However, as a common issue in VAEs that any introduced bias in the input will result in bias in the generated output~\cite{von2021self}, the diffusion steps and variance schedule need to be chosen cautiously, such that this model can be applied to different time series tasks smoothly. 
The proposed model is devised for general time series forecasting, it should be used properly to avoid the potential negative societal impacts, such as illegal applications.

In time series predictive analysis, disentanglement of the latent variables has been very important for interpreting the prediction to provide more reliance.~%
Due to the lack of prior knowledge of the entangled factors in generative time series forecasting, only unsupervised disentanglement learning can be done, which has been proven theoretically feasible for time series~\cite{li2021learning}. 
Despite this, for boarder applications of disentanglement and better performance, it is still worth exploring how to label the factors of time series in the future.  
Moreover, because of the uniqueness of time series data, it is also a promising direction to explore more generative and sampling methods for the time series generation task.


\section{Conclusion} \label{conclusion}

In this work, we propose a generative model with the bidirectional VAE as the backbone. 
To further improve the generalizability, we devise a coupled diffusion probabilistic model for time series forecasting. 
Then a scaled denoising network is developed to guarantee the prediction accuracy. 
Afterward, the latent variables are further disentangled for better model interpretability. 
Extensive experiments on synthetic data and real-world data validate that our proposed generative model achieves SOTA performance compared to existing competitive generative models.

\section*{Acknowledgement}

We thank Longyuan Power Group Corp. Ltd. for supporting this work.

\newpage
\bibliographystyle{plain}
\bibliography{ref}

\begin{thebibliography}{100}

\bibitem{abdar2021review}
Moloud Abdar, Farhad Pourpanah, Sadiq Hussain, Dana Rezazadegan, Li~Liu,
  Mohammad Ghavamzadeh, Paul Fieguth, Xiaochun Cao, Abbas Khosravi, U~Rajendra
  Acharya, et~al.
\newblock A review of uncertainty quantification in deep learning: Techniques,
  applications and challenges.
\newblock {\em Information Fusion}, 76:243--297, 2021.

\bibitem{ancona2018towards}
Marco Ancona, Enea Ceolini, Cengiz {\"O}ztireli, and Markus Gross.
\newblock Towards better understanding of gradient-based attribution methods
  for deep neural networks.
\newblock In {\em International Conference on Learning Representations}, 2018.

\bibitem{ariyo2014stock}
Adebiyi~A Ariyo, Adewumi~O Adewumi, and Charles~K Ayo.
\newblock Stock price prediction using the {ARIMA} model.
\newblock In {\em UKSim-AMSS 16th International Conference on Computer
  Modelling and Simulation}, pages 106--112. IEEE, 2014.

\bibitem{bai2018empirical}
Shaojie Bai, J~Zico Kolter, and Vladlen Koltun.
\newblock An empirical evaluation of generic convolutional and recurrent
  networks for sequence modeling.
\newblock {\em arXiv preprint arXiv:1803.01271}, 2018.

\bibitem{barahona1996detection}
Mauricio Barahona and Chi-Sang Poon.
\newblock Detection of nonlinear dynamics in short, noisy time series.
\newblock {\em Nature}, 381(6579):215--217, 1996.

\bibitem{bengio2013representation}
Yoshua Bengio, Aaron Courville, and Pascal Vincent.
\newblock Representation learning: A review and new perspectives.
\newblock {\em IEEE Transactions on Pattern Analysis and Machine Intelligence},
  35(8):1798--1828, 2013.

\bibitem{binkowski2018autoregressive}
Mikolaj Binkowski, Gautier Marti, and Philippe Donnat.
\newblock Autoregressive convolutional neural networks for asynchronous time
  series.
\newblock In {\em International Conference on Machine Learning}, pages
  580--589. PMLR, 2018.

\bibitem{borovykh2017conditional}
Anastasia Borovykh, Sander Bohte, and Cornelis~W Oosterlee.
\newblock Conditional time series forecasting with convolutional neural
  networks.
\newblock {\em STAT}, 1050:16, 2017.

\bibitem{box1968some}
George~EP Box and Gwilym~M Jenkins.
\newblock Some recent advances in forecasting and control.
\newblock {\em Journal of the Royal Statistical Society. Series C (Applied
  Statistics)}, 17(2):91--109, 1968.

\bibitem{brahim2004gaussian}
Sofiane Brahim-Belhouari and Amine Bermak.
\newblock Gaussian process for nonstationary time series prediction.
\newblock {\em Computational Statistics \& Data Analysis}, 47(4):705--712,
  2004.

\bibitem{cao2020spectral}
Defu Cao, Yujing Wang, Juanyong Duan, Ce~Zhang, Xia Zhu, Congrui Huang, Yunhai
  Tong, Bixiong Xu, Jing Bai, Jie Tong, et~al.
\newblock Spectral temporal graph neural network for multivariate time-series
  forecasting.
\newblock {\em Advances in Neural Information Processing Systems},
  33:17766--17778, 2020.

\bibitem{cao2003support}
Li-Juan Cao and Francis Eng~Hock Tay.
\newblock Support vector machine with adaptive parameters in financial time
  series forecasting.
\newblock {\em IEEE Transactions on Neural Networks}, 14(6):1506--1518, 2003.

\bibitem{chang2017dilated}
Shiyu Chang, Yang Zhang, Wei Han, Mo~Yu, Xiaoxiao Guo, Wei Tan, Xiaodong Cui,
  Michael Witbrock, Mark~A Hasegawa-Johnson, and Thomas~S Huang.
\newblock Dilated recurrent neural networks.
\newblock {\em Advances in Neural Information Processing Systems}, 30, 2017.

\bibitem{chatzis2017recurrent}
Sotirios~P Chatzis.
\newblock Recurrent latent variable conditional heteroscedasticity.
\newblock In {\em IEEE International Conference on Acoustics, Speech and Signal
  Processing}, pages 2711--2715. IEEE, 2017.

\bibitem{chen2018isolating}
Ricky~TQ Chen, Xuechen Li, Roger~B Grosse, and David~K Duvenaud.
\newblock Isolating sources of disentanglement in variational autoencoders.
\newblock {\em Advances in Neural Information Processing Systems}, 31, 2018.

\bibitem{chung2015recurrent}
Junyoung Chung, Kyle Kastner, Laurent Dinh, Kratarth Goel, Aaron~C Courville,
  and Yoshua Bengio.
\newblock A recurrent latent variable model for sequential data.
\newblock {\em Advances in Neural Information Processing Systems}, 28, 2015.

\bibitem{dau2019ucr}
Hoang~Anh Dau, Anthony Bagnall, Kaveh Kamgar, Chin-Chia~Michael Yeh, Yan Zhu,
  Shaghayegh Gharghabi, Chotirat~Ann Ratanamahatana, and Eamonn Keogh.
\newblock The {UCR} time series archive.
\newblock {\em IEEE/CAA Journal of Automatica Sinica}, 6(6):1293--1305, 2019.

\bibitem{de2020normalizing}
Emmanuel de~B{\'e}zenac, Syama~Sundar Rangapuram, Konstantinos Benidis, Michael
  Bohlke-Schneider, Richard Kurle, Lorenzo Stella, Hilaf Hasson, Patrick
  Gallinari, and Tim Januschowski.
\newblock Normalizing kalman filters for multivariate time series analysis.
\newblock {\em Advances in Neural Information Processing Systems},
  33:2995--3007, 2020.

\bibitem{debnath2021exploring}
Ankur Debnath, Govind Waghmare, Hardik Wadhwa, Siddhartha Asthana, and Ankur
  Arora.
\newblock Exploring generative data augmentation in multivariate time series
  forecasting: Opportunities and challenges.
\newblock {\em Solar-Energy}, 137:52--560, 2021.

\bibitem{deng2020residual}
Yuntian Deng, Anton Bakhtin, Myle Ott, Arthur Szlam, and Marc'Aurelio Ranzato.
\newblock Residual energy-based models for text generation.
\newblock In {\em International Conference on Learning Representations}, 2019.

\bibitem{du2019implicit}
Yilun Du and Igor Mordatch.
\newblock Implicit generation and modeling with energy based models.
\newblock {\em Advances in Neural Information Processing Systems}, 32, 2019.

\bibitem{ellner1995chaos}
Stephen Ellner and Peter Turchin.
\newblock Chaos in a noisy world: New methods and evidence from time-series
  analysis.
\newblock {\em The American Naturalist}, 145(3):343--375, 1995.

\bibitem{farnoosh2020deep}
Amirreza Farnoosh, Bahar Azari, and Sarah Ostadabbas.
\newblock Deep switching auto-regressive factorization: Application to time
  series forecasting.
\newblock {\em arXiv preprint arXiv:2009.05135}, 2020.

\bibitem{fokianos2009poisson}
Konstantinos Fokianos, Anders Rahbek, and Dag Tj{\o}stheim.
\newblock Poisson autoregression.
\newblock {\em Journal of the American Statistical Association},
  104(488):1430--1439, 2009.

\bibitem{forestier2017generating}
Germain Forestier, Fran{\c{c}}ois Petitjean, Hoang~Anh Dau, Geoffrey~I Webb,
  and Eamonn Keogh.
\newblock Generating synthetic time series to augment sparse datasets.
\newblock In {\em IEEE International Conference on Data Mining}, pages
  865--870. IEEE, 2017.

\bibitem{fortuin2020gp}
Vincent Fortuin, Dmitry Baranchuk, Gunnar R{\"a}tsch, and Stephan Mandt.
\newblock {GP-VAE}: Deep probabilistic time series imputation.
\newblock In {\em International Conference on Artificial Intelligence and
  Statistics}, pages 1651--1661. PMLR, 2020.

\bibitem{fortuin2019som}
Vincent Fortuin, Matthias H{\"u}ser, Francesco Locatello, Heiko Strathmann, and
  Gunnar R{\"a}tsch.
\newblock {SOM-VAE}: Interpretable discrete representation learning on time
  series.
\newblock In {\em International Conference on Learning Representations}, 2019.

\bibitem{foster1992neural}
WR~Foster, F~Collopy, and LH~Ungar.
\newblock Neural network forecasting of short, noisy time series.
\newblock {\em Computers \& Chemical Engineering}, 16(4):293--297, 1992.

\bibitem{fraccaro2017disentangled}
Marco Fraccaro, Simon Kamronn, Ulrich Paquet, and Ole Winther.
\newblock A disentangled recognition and nonlinear dynamics model for
  unsupervised learning.
\newblock {\em Advances in Neural Information Processing Systems}, 30, 2017.

\bibitem{gamboa2017deep}
John Cristian~Borges Gamboa.
\newblock Deep learning for time-series analysis.
\newblock {\em arXiv preprint arXiv:1701.01887}, 2017.

\bibitem{gawlikowski2021survey}
Jakob Gawlikowski, Cedrique Rovile~Njieutcheu Tassi, Mohsin Ali, Jongseok Lee,
  Matthias Humt, Jianxiang Feng, Anna Kruspe, Rudolph Triebel, Peter Jung,
  Ribana Roscher, et~al.
\newblock A survey of uncertainty in deep neural networks.
\newblock {\em arXiv preprint arXiv:2107.03342}, 2021.

\bibitem{gers2002learning}
Felix~A Gers, Nicol~N Schraudolph, and J{\"u}rgen Schmidhuber.
\newblock Learning precise timing with {LSTM} recurrent networks.
\newblock {\em Journal of Machine Learning Research}, 3(Aug):115--143, 2002.

\bibitem{giles2001noisy}
C~Lee Giles, Steve Lawrence, and Ah~Chung Tsoi.
\newblock Noisy time series prediction using recurrent neural networks and
  grammatical inference.
\newblock {\em Machine Learning}, 44(1):161--183, 2001.

\bibitem{girin2020dynamical}
Laurent Girin, Simon Leglaive, Xiaoyu Bie, Julien Diard, Thomas Hueber, and
  Xavier Alameda-Pineda.
\newblock Dynamical variational autoencoders: A comprehensive review.
\newblock {\em arXiv preprint arXiv:2008.12595}, 2020.

\bibitem{greff2016lstm}
Klaus Greff, Rupesh~K Srivastava, Jan Koutn{\'\i}k, Bas~R Steunebrink, and
  J{\"u}rgen Schmidhuber.
\newblock {LSTM}: A search space odyssey.
\newblock {\em IEEE Transactions on Neural Networks and Learning Systems},
  28(10):2222--2232, 2016.

\bibitem{gude2020flood}
Vinayaka Gude, Steven Corns, and Suzanna Long.
\newblock Flood prediction and uncertainty estimation using deep learning.
\newblock {\em Water}, 12(3):884, 2020.

\bibitem{guo2019attention}
Shengnan Guo, Youfang Lin, Ning Feng, Chao Song, and Huaiyu Wan.
\newblock Attention based spatial-temporal graph convolutional networks for
  traffic flow forecasting.
\newblock In {\em AAAI Conference on Artificial Intelligence}, volume~33, pages
  922--929, 2019.

\bibitem{hardt2020explaining}
Michaela Hardt, Alvin Rajkomar, Gerardo Flores, Andrew Dai, Michael Howell,
  Greg Corrado, Claire Cui, and Moritz Hardt.
\newblock Explaining an increase in predicted risk for clinical alerts.
\newblock In {\em ACM Conference on Health, Inference, and Learning}, pages
  80--89, 2020.

\bibitem{higgins2016beta}
Irina Higgins, Loic Matthey, Arka Pal, Christopher Burgess, Xavier Glorot,
  Matthew Botvinick, Shakir Mohamed, and Alexander Lerchner.
\newblock {beta-VAE}: Learning basic visual concepts with a constrained
  variational framework.
\newblock 2016.

\bibitem{hillmer1982arima}
Steven~Craig Hillmer and George~C Tiao.
\newblock An {ARIMA}-model-based approach to seasonal adjustment.
\newblock {\em Journal of the American Statistical Association},
  77(377):63--70, 1982.

\bibitem{ho2020denoising}
Jonathan Ho, Ajay Jain, and Pieter Abbeel.
\newblock Denoising diffusion probabilistic models.
\newblock {\em Advances in Neural Information Processing Systems},
  33:6840--6851, 2020.

\bibitem{ho1998use}
Siu~Lau Ho and Min Xie.
\newblock The use of {ARIMA} models for reliability forecasting and analysis.
\newblock {\em Computers \& Industrial Engineering}, 35(1-2):213--216, 1998.

\bibitem{hyvarinen2009estimation}
Aapo Hyv{\"a}rinen, Jarmo Hurri, and Patrik~O Hoyer.
\newblock Estimation of non-normalized statistical models.
\newblock In {\em Natural Image Statistics}, pages 419--426. Springer, 2009.

\bibitem{ismail2020benchmarking}
Aya~Abdelsalam Ismail, Mohamed Gunady, Hector Corrada~Bravo, and Soheil Feizi.
\newblock Benchmarking deep learning interpretability in time series
  predictions.
\newblock {\em Advances in Neural Information Processing Systems},
  33:6441--6452, 2020.

\bibitem{ismail2019input}
Aya~Abdelsalam Ismail, Mohamed Gunady, Luiz Pessoa, Hector Corrada~Bravo, and
  Soheil Feizi.
\newblock Input-cell attention reduces vanishing saliency of recurrent neural
  networks.
\newblock {\em Advances in Neural Information Processing Systems}, 32, 2019.

\bibitem{ismail2019deep}
Hassan Ismail~Fawaz, Germain Forestier, Jonathan Weber, Lhassane Idoumghar, and
  Pierre-Alain Muller.
\newblock Deep learning for time series classification: a review.
\newblock {\em Data Mining and Knowledge Discovery}, 33(4):917--963, 2019.

\bibitem{karunasinghe2006chaotic}
Dulakshi~SK Karunasinghe and Shie-Yui Liong.
\newblock Chaotic time series prediction with a global model: Artificial neural
  network.
\newblock {\em Journal of Hydrology}, 323(1-4):92--105, 2006.

\bibitem{kendall2017uncertainties}
Alex Kendall and Yarin Gal.
\newblock What uncertainties do we need in bayesian deep learning for computer
  vision?
\newblock {\em Advances in Neural Information Processing Systems}, 30, 2017.

\bibitem{kim2018interpretability}
Been Kim, Martin Wattenberg, Justin Gilmer, Carrie Cai, James Wexler, Fernanda
  Viegas, et~al.
\newblock Interpretability beyond feature attribution: Quantitative testing
  with concept activation vectors ({TCAV}).
\newblock In {\em International Conference on Machine Learning}, pages
  2668--2677. PMLR, 2018.

\bibitem{kim2018disentangling}
Hyunjik Kim and Andriy Mnih.
\newblock Disentangling by factorising.
\newblock In {\em International Conference on Machine Learning}, pages
  2649--2658. PMLR, 2018.

\bibitem{kim2003financial}
Kyoung-jae Kim.
\newblock Financial time series forecasting using support vector machines.
\newblock {\em Neurocomputing}, 55(1-2):307--319, 2003.

\bibitem{kindermans2019reliability}
Pieter-Jan Kindermans, Sara Hooker, Julius Adebayo, Maximilian Alber, Kristof~T
  Sch{\"u}tt, Sven D{\"a}hne, Dumitru Erhan, and Been Kim.
\newblock The (un) reliability of saliency methods.
\newblock In {\em Explainable AI: Interpreting, Explaining and Visualizing Deep
  Learning}, pages 267--280. Springer, 2019.

\bibitem{kingma2013auto}
Diederik~P Kingma and Max Welling.
\newblock Auto-encoding variational bayes.
\newblock {\em arXiv preprint arXiv:1312.6114}, 2013.

\bibitem{kingma2016improved}
Durk~P Kingma, Tim Salimans, Rafal Jozefowicz, Xi~Chen, Ilya Sutskever, and Max
  Welling.
\newblock Improved variational inference with inverse autoregressive flow.
\newblock {\em Advances in Neural Information Processing Systems}, 29, 2016.

\bibitem{kitaev2020reformer}
Nikita Kitaev, {\L}ukasz Kaiser, and Anselm Levskaya.
\newblock Reformer: The efficient transformer.
\newblock {\em arXiv preprint arXiv:2001.04451}, 2020.

\bibitem{kumar2019maximum}
Rithesh Kumar, Sherjil Ozair, Anirudh Goyal, Aaron Courville, and Yoshua
  Bengio.
\newblock Maximum entropy generators for energy-based models.
\newblock {\em arXiv preprint arXiv:1901.08508}, 2019.

\bibitem{kunitomo2021robust}
Naoto Kunitomo and Seisho Sato.
\newblock A robust-filtering method for noisy non-stationary multivariate time
  series with econometric applications.
\newblock {\em Japanese Journal of Statistics and Data Science}, 4(1):373--410,
  2021.

\bibitem{lai2018modeling}
Guokun Lai, Wei-Cheng Chang, Yiming Yang, and Hanxiao Liu.
\newblock Modeling long-and short-term temporal patterns with deep neural
  networks.
\newblock In {\em International ACM SIGIR Conference on Research \& Development
  in Information Retrieval}, pages 95--104, 2018.

\bibitem{lake2017building}
Brenden~M Lake, Tomer~D Ullman, Joshua~B Tenenbaum, and Samuel~J Gershman.
\newblock Building machines that learn and think like people.
\newblock {\em Behavioral and Brain Sciences}, 40, 2017.

\bibitem{lea2016temporal}
Colin Lea, Rene Vidal, Austin Reiter, and Gregory~D Hager.
\newblock Temporal convolutional networks: A unified approach to action
  segmentation.
\newblock In {\em European Conference on Computer Vision}, pages 47--54.
  Springer, 2016.

\bibitem{li2019enhancing}
Shiyang Li, Xiaoyong Jin, Yao Xuan, Xiyou Zhou, Wenhu Chen, Yu-Xiang Wang, and
  Xifeng Yan.
\newblock Enhancing the locality and breaking the memory bottleneck of
  transformer on time series forecasting.
\newblock {\em Advances in Neural Information Processing Systems}, 32, 2019.

\bibitem{li2018diffusion}
Yaguang Li, Rose Yu, Cyrus Shahabi, and Yan Liu.
\newblock Diffusion convolutional recurrent neural network: Data-driven traffic
  forecasting.
\newblock In {\em International Conference on Learning Representations}, 2018.

\bibitem{li2016renyi}
Yingzhen Li and Richard~E Turner.
\newblock R{\'e}nyi divergence variational inference.
\newblock {\em Advances in Neural Information Processing Systems}, 29, 2016.

\bibitem{li2021learning}
Yuening Li, Zhengzhang Chen, Daochen Zha, Mengnan Du, Denghui Zhang, Haifeng
  Chen, and Xia Hu.
\newblock Learning disentangled representations for time series.
\newblock {\em arXiv preprint arXiv:2105.08179}, 2021.

\bibitem{li2019learning}
Zengyi Li, Yubei Chen, and Friedrich~T Sommer.
\newblock Learning energy-based models in high-dimensional spaces with
  multi-scale denoising score matching.
\newblock {\em arXiv preprint arXiv:1910.07762}, 2019.

\bibitem{lu2009financial}
Chi-Jie Lu, Tian-Shyug Lee, and Chih-Chou Chiu.
\newblock Financial time series forecasting using independent component
  analysis and support vector regression.
\newblock {\em Decision Support Systems}, 47(2):115--125, 2009.

\bibitem{maddix2018deep}
Danielle~C Maddix, Yuyang Wang, and Alex Smola.
\newblock Deep factors with {Gaussian} processes for forecasting.
\newblock {\em arXiv preprint arXiv:1812.00098}, 2018.

\bibitem{matheson1976scoring}
James~E Matheson and Robert~L Winkler.
\newblock Scoring rules for continuous probability distributions.
\newblock {\em Management Science}, 22(10):1087--1096, 1976.

\bibitem{nguyen2021temporal}
Nam Nguyen and Brian Quanz.
\newblock Temporal latent auto-encoder: A method for probabilistic multivariate
  time series forecasting.
\newblock In {\em AAAI Conference on Artificial Intelligence}, volume~35, pages
  9117--9125, 2021.

\bibitem{nichol2021improved}
Alexander~Quinn Nichol and Prafulla Dhariwal.
\newblock Improved denoising diffusion probabilistic models.
\newblock In {\em International Conference on Machine Learning}, pages
  8162--8171. PMLR, 2021.

\bibitem{petropoulos2018exploring}
Fotios Petropoulos, Rob~J Hyndman, and Christoph Bergmeir.
\newblock Exploring the sources of uncertainty: Why does bagging for time
  series forecasting work?
\newblock {\em European Journal of Operational Research}, 268(2):545--554,
  2018.

\bibitem{qin2017dual}
Yao Qin, Dongjin Song, Haifeng Chen, Wei Cheng, Guofei Jiang, and Garrison~W
  Cottrell.
\newblock A dual-stage attention-based recurrent neural network for time series
  prediction.
\newblock In {\em International Joint Conference on Artificial Intelligence},
  2017.

\bibitem{rangapuram2018deep}
Syama~Sundar Rangapuram, Matthias~W Seeger, Jan Gasthaus, Lorenzo Stella,
  Yuyang Wang, and Tim Januschowski.
\newblock Deep state space models for time series forecasting.
\newblock {\em Advances in Neural Information Processing Systems}, 31, 2018.

\bibitem{rasul2021autoregressive}
Kashif Rasul, Calvin Seward, Ingmar Schuster, and Roland Vollgraf.
\newblock Autoregressive denoising diffusion models for multivariate
  probabilistic time series forecasting.
\newblock In {\em International Conference on Machine Learning}, pages
  8857--8868. PMLR, 2021.

\bibitem{roeder2017sticking}
Geoffrey Roeder, Yuhuai Wu, and David~K Duvenaud.
\newblock Sticking the landing: Simple, lower-variance gradient estimators for
  variational inference.
\newblock {\em Advances in Neural Information Processing Systems}, 30, 2017.

\bibitem{salinas2019high}
David Salinas, Michael Bohlke-Schneider, Laurent Callot, Roberto Medico, and
  Jan Gasthaus.
\newblock High-dimensional multivariate forecasting with low-rank {Gaussian}
  copula processes.
\newblock {\em Advances in Neural Information Processing Systems}, 32, 2019.

\bibitem{salinas2020deepar}
David Salinas, Valentin Flunkert, Jan Gasthaus, and Tim Januschowski.
\newblock {DeepAR:} probabilistic forecasting with autoregressive recurrent
  networks.
\newblock {\em International Journal of Forecasting}, 36(3):1181--1191, 2020.

\bibitem{saremi2019neural}
Saeed Saremi, Aapo Hyv{\"a}rinen, et~al.
\newblock Neural empirical {Bayes}.
\newblock {\em Journal of Machine Learning Research}, 2019.

\bibitem{sen2019think}
Rajat Sen, Hsiang-Fu Yu, and Inderjit~S Dhillon.
\newblock Think globally, act locally: A deep neural network approach to
  high-dimensional time series forecasting.
\newblock {\em Advances in Neural Information Processing Systems}, 32, 2019.

\bibitem{sherstinsky2020fundamentals}
Alex Sherstinsky.
\newblock Fundamentals of recurrent neural network ({RNN}) and long short-term
  memory ({LSTM}) network.
\newblock {\em Physica D: Nonlinear Phenomena}, 404:132306, 2020.

\bibitem{arima_2020}
Qiquan Shi, Jiaming Yin, Jiajun Cai, Andrzej Cichocki, Tatsuya Yokota, Lei
  Chen, Mingxuan Yuan, and Jia Zeng.
\newblock Block hankel tensor {ARIMA} for multiple short time series
  forecasting.
\newblock In {\em AAAI Conference on Artificial Intelligence}, volume~34, pages
  5758--5766, 2020.

\bibitem{shih2019temporal}
Shun-Yao Shih, Fan-Keng Sun, and Hung-yi Lee.
\newblock Temporal pattern attention for multivariate time series forecasting.
\newblock {\em Machine Learning}, 108(8):1421--1441, 2019.

\bibitem{singh2000noisy}
Sameer Singh.
\newblock Noisy time-series prediction using pattern recognition techniques.
\newblock {\em Computational Intelligence}, 16(1):114--133, 2000.

\bibitem{smyl2016data}
Slawek Smyl and Karthik Kuber.
\newblock Data preprocessing and augmentation for multiple short time series
  forecasting with recurrent neural networks.
\newblock In {\em International Symposium on Forecasting}, 2016.

\bibitem{sohl2015deep}
Jascha Sohl-Dickstein, Eric Weiss, Niru Maheswaranathan, and Surya Ganguli.
\newblock Deep unsupervised learning using nonequilibrium thermodynamics.
\newblock In {\em International Conference on Machine Learning}, pages
  2256--2265. PMLR, 2015.

\bibitem{song2018attend}
Huan Song, Deepta Rajan, Jayaraman~J Thiagarajan, and Andreas Spanias.
\newblock Attend and diagnose: Clinical time series analysis using attention
  models.
\newblock In {\em AAAI Conference on Artificial Intelligence}, 2018.

\bibitem{song2019generative}
Yang Song and Stefano Ermon.
\newblock Generative modeling by estimating gradients of the data distribution.
\newblock {\em Advances in Neural Information Processing Systems}, 32, 2019.

\bibitem{tonekaboni2020went}
Sana Tonekaboni, Shalmali Joshi, Kieran Campbell, David~K Duvenaud, and Anna
  Goldenberg.
\newblock What went wrong and when? instance-wise feature importance for
  time-series black-box models.
\newblock {\em Advances in Neural Information Processing Systems}, 33:799--809,
  2020.

\bibitem{vahdat2020nvae}
Arash Vahdat and Jan Kautz.
\newblock {NVAE}: A deep hierarchical variational autoencoder.
\newblock {\em Advances in Neural Information Processing Systems},
  33:19667--19679, 2020.

\bibitem{vincent2011connection}
Pascal Vincent.
\newblock A connection between score matching and denoising autoencoders.
\newblock {\em Neural Computation}, 23(7):1661--1674, 2011.

\bibitem{vincent2010stacked}
Pascal Vincent, Hugo Larochelle, Isabelle Lajoie, Yoshua Bengio, Pierre-Antoine
  Manzagol, and L{\'e}on Bottou.
\newblock Stacked denoising autoencoders: Learning useful representations in a
  deep network with a local denoising criterion.
\newblock {\em Journal of Machine Learning Research}, 11(12), 2010.

\bibitem{von2021self}
Julius Von~K{\"u}gelgen, Yash Sharma, Luigi Gresele, Wieland Brendel, Bernhard
  Sch{\"o}lkopf, Michel Besserve, and Francesco Locatello.
\newblock Self-supervised learning with data augmentations provably isolates
  content from style.
\newblock {\em Advances in Neural Information Processing Systems}, 34, 2021.

\bibitem{wang2018gaussian}
Juntao Wang, Wun~Kwan Yam, Kin~Long Fong, Siew~Ann Cheong, and KY~Wong.
\newblock Gaussian process kernels for noisy time series: Application to
  housing price prediction.
\newblock In {\em International Conference on Neural Information Processing},
  pages 78--89. Springer, 2018.

\bibitem{watanabe1960information}
Satosi Watanabe.
\newblock Information theoretical analysis of multivariate correlation.
\newblock {\em Ibm Journal of Research and Development}, 4(1):66--82, 1960.

\bibitem{wen2017multi}
Ruofeng Wen, Kari Torkkola, Balakrishnan Narayanaswamy, and Dhruv Madeka.
\newblock A multi-horizon quantile recurrent forecaster.
\newblock {\em arXiv preprint arXiv:1711.11053}, 2017.

\bibitem{west2020bayesian}
Mike West.
\newblock Bayesian forecasting of multivariate time series: scalability,
  structure uncertainty and decisions.
\newblock {\em Annals of the Institute of Statistical Mathematics},
  72(1):1--31, 2020.

\bibitem{wu2022recent}
Bingzhe Wu, Jintang Li, Chengbin Hou, Guoji Fu, Yatao Bian, Liang Chen, and
  Junzhou Huang.
\newblock Recent advances in reliable deep graph learning: Adversarial attack,
  inherent noise, and distribution shift.
\newblock {\em arXiv preprint arXiv:2202.07114}, 2022.

\bibitem{wu2018beyond}
Mike Wu, Michael Hughes, Sonali Parbhoo, Maurizio Zazzi, Volker Roth, and
  Finale Doshi-Velez.
\newblock Beyond sparsity: Tree regularization of deep models for
  interpretability.
\newblock In {\em AAAI Conference on Artificial Intelligence}, volume~32, 2018.

\bibitem{wu2020deep}
Neo Wu, Bradley Green, Xue Ben, and Shawn O'Banion.
\newblock Deep transformer models for time series forecasting: The influenza
  prevalence case.
\newblock {\em arXiv preprint arXiv:2001.08317}, 2020.

\bibitem{wu2019graph}
Zonghan Wu, Shirui Pan, Guodong Long, Jing Jiang, and Chengqi Zhang.
\newblock Graph wavenet for deep spatial-temporal graph modeling.
\newblock In {\em International Joint Conference on Artificial Intelligence},
  pages 1907--1913, 2019.

\bibitem{xie2016theory}
Jianwen Xie, Yang Lu, Song-Chun Zhu, and Yingnian Wu.
\newblock A theory of generative convnet.
\newblock In {\em International Conference on Machine Learning}, pages
  2635--2644. PMLR, 2016.

\bibitem{xie2021generative}
Jianwen Xie, Yifei Xu, Zilong Zheng, Song-Chun Zhu, and Ying~Nian Wu.
\newblock Generative pointnet: Deep energy-based learning on unordered point
  sets for 3d generation, reconstruction and classification.
\newblock In {\em Proceedings of the IEEE/CVF Conference on Computer Vision and
  Pattern Recognition}, pages 14976--14985, 2021.

\bibitem{xie2019learning}
Jianwen Xie, Song-Chun Zhu, and Ying~Nian Wu.
\newblock Learning energy-based spatial-temporal generative convnets for
  dynamic patterns.
\newblock {\em IEEE transactions on pattern analysis and machine intelligence},
  43(2):516--531, 2019.

\bibitem{xu2021autoformer}
Jiehui Xu, Jianmin Wang, Mingsheng Long, et~al.
\newblock Autoformer: Decomposition transformers with auto-correlation for
  long-term series forecasting.
\newblock {\em Advances in Neural Information Processing Systems}, 34, 2021.

\bibitem{yan2021scoregrad}
Tijin Yan, Hongwei Zhang, Tong Zhou, Yufeng Zhan, and Yuanqing Xia.
\newblock {ScoreGrad}: Multivariate probabilistic time series forecasting with
  continuous energy-based generative models.
\newblock {\em arXiv preprint arXiv:2106.10121}, 2021.

\bibitem{yoon2019time}
Jinsung Yoon, Daniel Jarrett, and Mihaela Van~der Schaar.
\newblock Time-series generative adversarial networks.
\newblock {\em Advances in Neural Information Processing Systems}, 32, 2019.

\bibitem{yu2018spatio}
Bing Yu, Haoteng Yin, and Zhanxing Zhu.
\newblock Spatio-temporal graph convolutional networks: a deep learning
  framework for traffic forecasting.
\newblock In {\em International Joint Conference on Artificial Intelligence},
  pages 3634--3640, 2018.

\bibitem{yu2019review}
Yong Yu, Xiaosheng Si, Changhua Hu, and Jianxun Zhang.
\newblock A review of recurrent neural networks: {LSTM} cells and network
  architectures.
\newblock {\em Neural Computation}, 31(7):1235--1270, 2019.

\bibitem{zhou2021informer}
Haoyi Zhou, Shanghang Zhang, Jieqi Peng, Shuai Zhang, Jianxin Li, Hui Xiong,
  and Wancai Zhang.
\newblock Informer: Beyond efficient transformer for long sequence time-series
  forecasting.
\newblock In {\em AAAI Conference on Artificial Intelligence}, 2021.

\bibitem{zou2019complex}
Yong Zou, Reik~V Donner, Norbert Marwan, Jonathan~F Donges, and J{\"u}rgen
  Kurths.
\newblock Complex network approaches to nonlinear time series analysis.
\newblock {\em Physics Reports}, 787:1--97, 2019.

\end{thebibliography}

\newpage

\appendix

\section{Related Work}  \label{sec:related}

\subsection{Time Series Forecasting}

We first briefly review the related literature of time series forecasting (TSF) methods as below.
%
Complex temporal patterns can be manifested over short- and long-term as the time series evolves across time. 
To leverage the time evolution nature, 
existing statistical models, such as ARIMA~\cite{box1968some} and Gaussian process regression~\cite{brahim2004gaussian} have been well established and applied to many downstream tasks~\cite{hillmer1982arima,ho1998use,ariyo2014stock}. 
Recurrent neural network (RNN) models are also introduced to model temporal dependencies for TSF in a sequence-to-sequence paradigm~\cite{greff2016lstm,chang2017dilated,wen2017multi,lai2018modeling,maddix2018deep,rangapuram2018deep,salinas2020deepar}. 
Besides, temporal attention~\cite{qin2017dual,song2018attend,shih2019temporal} and causal convolution~\cite{bai2018empirical,borovykh2017conditional,sen2019think} are further explored to model the intrinsic temporal dependencies. 
Recent Transformer-based models have strengthened the capability of exploring hidden intricate temporal patterns for long-term TSF~\cite{xu2021autoformer,li2019enhancing,wu2020deep,zhou2021informer}.
On the other hand, the multivariate nature of TSF is another topic many works have been focusing on. 
These works treat a collection of time series as a unified entity and mine the inter-series correlations with different techniques, such as 
probabilistic models~\cite{salinas2020deepar,rangapuram2018deep}, 
matrix/tensor factorization~\cite{sen2019think,arima_2020}, 
convolution neural networks (CNNs)~\cite{bai2018empirical,lai2018modeling},  
and graph neural networks (GNNs)~\cite{guo2019attention,li2018diffusion,yu2018spatio,wu2019graph,cao2020spectral}.

%
%
To improve the reliability and performance of TSF, 
instead of modeling the raw data, there exist works inferring the underlying distribution of the time series data with generative models~\cite{yoon2019time,de2020normalizing}. 
Many studies have employed a variational auto-encoder (VAE) to model the probabilistic distribution of sequential data~\cite{fraccaro2017disentangled, girin2020dynamical, chung2015recurrent, chatzis2017recurrent, nguyen2021temporal}. 
For example, VRNN~\cite{chung2015recurrent} employs the VAE to each hidden state of RNN such that the variability of highly structured sequential data can be captured. 
To yield predictive distribution for multivariate TSF, TLAE~\cite{nguyen2021temporal} implements nonlinear transformation by replacing matrix factorization with encoder-decoder architecture and temporal deep temporal latent model. 
Another line of generative methods for TSF focus on energy-based models (EBMs), such as TimeGrad~\cite{rasul2021autoregressive} and ScoreGrad~\cite{yan2021scoregrad}. 
EBMs do not restrict the tractability of the normalizing constants~\cite{yan2021scoregrad}. 
Though flexible, the unknown normalizing constant makes the training of EBMs particularly difficult.

This paper focuses on TSF with VAE-based models. 
Besides, as many real-world time series data are relatively short and small~\cite{smyl2016data}, 
a coupled probabilistic diffusion model is proposed to augment the input series, as well as the output series, simultaneously, such that the distribution space can be enlarged without increasing the aleatoric uncertainty~\cite{kendall2017uncertainties}. 
Moreover, to guarantee the generated target series moving toward the true target, a multi-scaled score-matching denoising network is plugged in for accurate future series prediction. 
To our knowledge, this is the first work focusing on generative TSF with the diffusion model and denoising techniques.

\subsection{Time Series Augmentation}

Both the traditional methods and deep learning methods can deteriorate when limited time series data are encountered. 
Generating synthetic time series is commonly adopted for augmenting short time series~\cite{dau2019ucr,forestier2017generating, yoon2019time}.
Transforming the original time series by cropping, flipping, and warping~\cite{ismail2019deep, debnath2021exploring} is another approach dedicated to TSF when the training data is limited. 
Whereas the synthetic time series may not respect the original feature relationship across time,  and the transformation methods do not change the distribution space. Thus, the overfitting issues cannot be avoided. 
Incorporating the probabilistic diffusion model for TSF differentiates our work from existing time series augmentation methods.

\subsection{Uncertainty Estimation and Denoising for Time Series Forecasting}

There exist works aiming to estimate the uncertainty~\cite{kendall2017uncertainties} for  time series forecasting~\cite{petropoulos2018exploring,west2020bayesian,gude2020flood} by epistemic uncertainty. 
Nevertheless, the inevitable aleatoric uncertainty of time series is often ignored, which may stem from error-prone data measurement, collection, and so forth~\cite{wu2022recent}. 
Another line of studies focuses on detecting noise in time series data~\cite{lu2009financial} or devising suitable models for noise alleviation~\cite{giles2001noisy}.
However, none of the existing works attempts to quantify the aleatoric uncertainty, which further differentiates our work from priors.

It is necessary to relieve the effect of noise in real-world time series data~\cite{ellner1995chaos}. 
\cite{barahona1996detection,kunitomo2021robust} propose to preprocess the time series with smoothing and filtering techniques. 
However, such preprocessing methods can only be applied to the noise raised by the irregular data of time series. 
Neural networks are also introduced to denoise the time series~\cite{foster1992neural,singh2000noisy,giles2001noisy,karunasinghe2006chaotic}, while these deep networks can only deal with specific types of time series as well.

\subsection{Interpretability of Time Series Forecasting}

A number of works put effort into explaining the deep neural networks~\cite{wu2018beyond,kim2018interpretability,ancona2018towards} to make the prediction more interpretable, 
but these methods often lack reliability when the explanation is sensitive to factors that do not contribute to the prediction~\cite{kindermans2019reliability}. 
Several works have been proposed to increase the reliability of TSF tasks~\cite{ismail2020benchmarking,ismail2019input}. 
For multivariate time series, the interpretability of the representations can be improved by mapping the time series into latent space~\cite{fortuin2019som}.
Besides, recent works have been proposed to disentangle the latent variables to identify the independent factors of the data, which can further lead to improved interpretability of the representation and  higher performance~\cite{higgins2016beta,lake2017building,kim2018disentangling}. 
The disentangled VAE has been applied to time series to benefit the generated results~\cite{li2021learning}. 
However, the choice of the latent variables is crucial for the disentanglement of time series data.  
We devise a bidirectional VAE (BVAE)  
and take the dimensions of each latent variable as the factors to be disentangled.

\section{Proofs of Lemma 1 and Lemma 2}  \label{appendix:lemma:derivation}

With the coupled diffusion process and \cref{eq:target,eq:true}, 
as well as  \cref{prop1}, 
introduced in the main text, 
the diffused target series and generated target series can be decomposed as 
$\widetilde{Y}^{(t)} = \langle \widetilde{Y}^{(t)}, \delta_{\widetilde{Y}}^{(t)} \rangle$ and 
$\widehat{Y}^{(t)} = \langle \widehat{Y}^{(t)}, \delta_{\widehat{Y}}^{(t)} \rangle$. 
Then, we can draw the following two conclusions: 

\begin{replemma}{theo1}
$ \forall \varepsilon > 0 $, 
there exists a probabilistic model 
$ \, f_{\phi, \theta} \vcentcolon = (p_{\phi}, p_{\theta})$ 
to guarantee that 
$ \mathcal{D}_{\mathrm{KL}} (q(\widetilde{Y}_r^{(t)}) || p_{\theta}(\widehat{Y}_r^{(t)})) < \varepsilon $, 
where 
$ \widehat{Y}_r^{(t)} = f_{\phi,\theta} (X^{(t)})  $. 
\end{replemma}

\vspace{-3ex}

\begin{proof}
According to Proposition 1, 
$ \widehat{Y}_r $ can be fully captured by the  model.
That is,  $ \| Y_r - \widehat{Y}_r \| \longrightarrow 0 $ where $Y_r$ is the ideal part of ground truth target series $Y$. 
And, with \cref{eq:true} 
(in the main text), $\widetilde{Y}_r^{(t)} = \sqrt{\bar{\alpha}^{\prime}_t}Y_r$. 
Therefore, 
$ \| \widetilde{Y}_r^{(t)} - \widehat{Y}_r^{(t)} \| \longrightarrow 0 $ when $t \rightarrow \infty$. 
\end{proof}
\vspace{1ex}

\begin{replemma}{theo2}
With the coupled diffusion process, 
the difference between  diffusion noise and  generation noise will be reduced, 
i.e., 
$ \lim_{t \rightarrow \infty} \mathcal{D}_{\mathrm{KL}} (q ({\delta_{\widetilde{Y}}^{(t)}}) || p_{\theta} (\delta_{\widehat{Y}}^{(t)} | Z^{(t)}) ) < 
\mathcal{D}_{\mathrm{KL}} (q(\epsilon_Y) || p_{\theta}(\epsilon_{\widehat{Y}}))$ . 
\end{replemma}
\vspace{-3ex}

\begin{proof}
According to  \cref{prop1}, 
the noise of $Y$ consists of the estimation noise $\epsilon_{\widehat{Y}}$ and residual noise $\delta_Y$, 
i.e., 
$ \epsilon_Y = \langle \epsilon_{\widehat{Y}}, \delta_Y \rangle $ where $\epsilon_{\widehat{Y}}$ and $\delta_Y$ are independent of each other, 
then $ q(\epsilon_Y) = q(\epsilon_{\widehat{Y}}) q(\delta_Y) $. 
Let $ 
\Delta = \mathcal{D}_{\mathrm{KL}}(q(\epsilon_Y)||p_\theta(\epsilon_{\widehat{Y}}))  - \mathcal{D}_{\mathrm{KL}}(q(\epsilon_{\widehat{Y}}) || p_\theta(\epsilon_{\widehat{Y}})) 
$, 
we have 
\begin{equation*} 
\begin{aligned}
    \Delta
    &=  
    \mathcal{D}_{\mathrm{KL}}(q(\epsilon_{\widehat{Y}}) q(\delta_Y) || p_\theta(\epsilon_{\widehat{Y}})) - 
    \mathcal{D}_{\mathrm{KL}}(q(\epsilon_{\widehat{Y}}) || p_\theta(\epsilon_{\widehat{Y}})) 
    \\
    &= 
    \mathcal{D}_{\mathrm{KL}}(q(\epsilon_{\widehat{Y}}) || p_\theta(\epsilon_{\widehat{Y}})) + \mathcal{D}_{\mathrm{KL}}(q(\delta_Y) || p_\theta(\epsilon_{\widehat{Y}}))  - \mathcal{D}_{\mathrm{KL}}(q(\epsilon_{\widehat{Y}}) || p_\theta(\epsilon_{\widehat{Y}})) 
    \\
    &= 
    \mathcal{D}_{\mathrm{KL}}(q(\delta_Y) || p_\theta(\epsilon_{\widehat{Y}})) 
    > 0 \, ,
\end{aligned}
\end{equation*}
which leads to 
$ \mathcal{D}_{\mathrm{KL}} (q(\epsilon_Y) || p_{\theta}(\epsilon_{\widehat{Y}})) > \mathcal{D}_{\mathrm{KL}} (q(\epsilon_{\widehat{Y}}) || p_{\theta} (\epsilon_{\widehat{Y}}) > 0 $. 
Moreover, 
both $ {\delta}_{\widetilde{Y}}^{(t)} $ and $ {\delta}_{\widehat{Y}}^{(t)} $ are Gaussian noises, 
when $t \rightarrow \infty$, 
$\exists$ $\varepsilon^{\prime} > 0$, 
we have 
$ 
\mathcal{D}_{\mathrm{KL}} (q({\delta}_{\widetilde{Y}}^{(t)})||p_\theta({\delta}_{\widehat{Y}}^{(t)}|Z^{(t)}))  
\leq \varepsilon^{\prime} < \mathcal{D}_{\mathrm{KL}} (q(\epsilon_Y)||p_\theta(\epsilon_{\widehat{Y}}))
$. 
%
\end{proof}

\begin{table}[t]
    \centering
    \setlength\tabcolsep{2pt}
    \renewcommand{\arraystretch}{1.2}
    \caption{
    Statistical descriptions of the real-world datasets.
    }   \label{tab:dataaset}
    \begin{tabular}{c|c|c|c|c|c|c|c}
    \toprule
    \multirow{2}{*}{Datasets} & \multirow{2}{*}{\# Dims.} & \multicolumn{2}{c|}{Full Data} & \multicolumn{2}{c|}{Sliced Data} &  \multirow{2}{*}{Target Variable} & \multirow{2}{*}{Time Interval}
    \\
    \cline{3-6} 
    ~&~& Time Span & \# Points & Pct. of Full Data & \# Points &~ &~
    \\
    \midrule 
    Traffic & 862 & 2015-2016 & 17544 & 5\% & 877 & Sensor 862 & 1 hour \\
    Electricity & 321 & 2011-2014 &18381 & 3\%& 551& MT\_321 & 10 mins \\
    Weather & 21 & 2020-2021 &36761 & 2\% & 735 & CO2~(ppm) & 10 mins \\
    ETTm1 & 7 & 2016-2018 &69680 & 1\% & 697 & OT & 15 mins \\
    ETTh1 & 7 & 2016-2018 & 17420 & 5\%& 871 & OT & 1 hour \\
    Wind & 7 & 2020-2021 & 45550 & 2\%& 911 & wind\_power & 15 mins \\
    \bottomrule
    \end{tabular}
\end{table}

\section{Extra Implementation Details}  \label{exp}

\subsection{Experimental Settings}  \label{exp_repro}

\textbf{Datasets Description.~~}
The main descriptive statistics of the real-world datasets adopted in the experiments of this work are demonstrated in~\cref{tab:dataaset}.

\textbf{Input Representation.~~} 
We adopt the embedding method introduced in~\cite{zhou2021informer} and feed it to an RNN to extract the temporal dependency.
Then we concatenate them as follows:
\begin{equation*}
    X_{\text{input}} = \mathrm{CONCAT} (\mathrm{RNN} (\mathcal{E}(X)), \mathcal{E}(X)) \, ,
\end{equation*}
where $X$ is the raw time series data and $\mathcal{E}(\cdot)$ denotes the embedding operation.
Here, we use a two-layer gated recurrent unit (GRU), and the dimensionality of the hidden state and embeddings are $128$ and $64$, respectively.

{\bf Diffusion Process Configuration.~~}
Besides, the diffusion process is configured to be $\beta_t \in [0, 0.1]$ and $T=100$ for the \textbf{Weather} dataset, 
$\beta_t \in [0, 0.1]$ and $T=1000$ for the \textbf{ETTh1} dataset, 
$\beta_t \in [0, 0.08]$ and $T=1000$ for the \textbf{Wind} dataset,
and 
$\beta_t \in [0, 0.01]$ and $T=1000$ for the other datasets.

\subsection{Implementation Details of Baselines}

We select previous state-of-the-art generative models as our baselines in the experiments on synthetic and real-world datasets. 
Specifically, 
{\bf 1) GP-copula} \cite{salinas2019high} is a method based on the Gaussian process, which is devoted to high-dimensional multivariate time series, 
{\bf 2) DeepAR} \cite{salinas2020deepar} combines traditional auto-regressive models with RNNs by modeling a probabilistic distribution in an auto-encoder fashion, 
{\bf 3) TimeGrad} \cite{rasul2021autoregressive} is an auto-regressive model for multivariate probabilistic time series forecasting with the help of an energy-based model, 
{\bf 4) Vanilla VAE} (VAE for short) \cite{kingma2013auto} is a classical statistical variational inference method on top of auto-encoder, 
{\bf 5) NVAE} \cite{vahdat2020nvae} is a deep hierarchical VAE built for image generation using depth-wise separable convolutions and batch normalization, 
{\bf 6) factor-VAE} (f-VAE for short) \cite{kim2018disentangling} disentangles the latent variables by encouraging the distribution of representations to be factorial and independent across dimensions, 
and {\bf 7) $\beta$-TCVAE} \cite{chen2018isolating} learns the disentangled representations with total correlation variational auto-encoder algorithm.

\begin{figure}[t]
    \centering
    \includegraphics[width=0.9\textwidth]{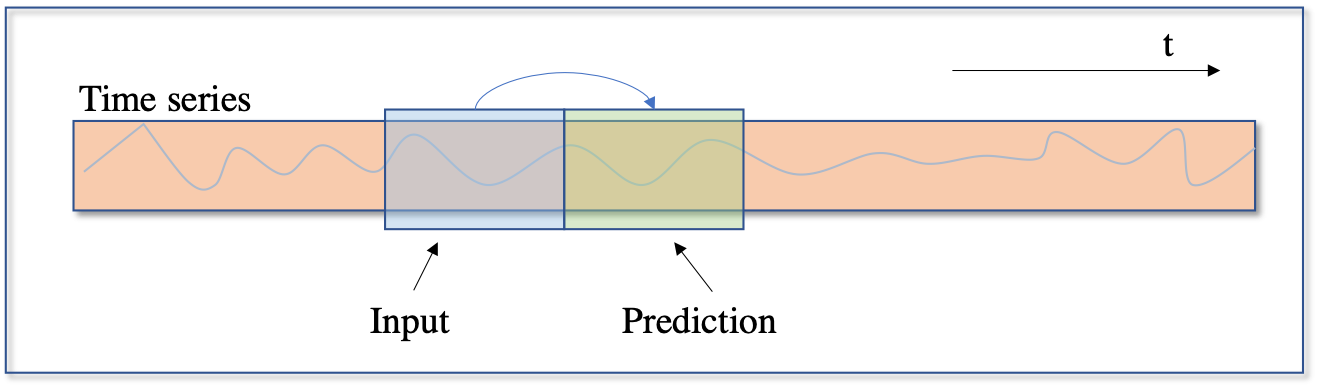}
    \caption{
    Forecasting process of DeepAR, TimeGrad, and GP-copula. The sliding step is set to 1.
    }
    \label{fig:set}
    \vspace{-2ex}
\end{figure}

To train DeepAR, TimeGrad, and GP-copula in accordance with their original settings, the batch is constructed without shuffling the samples. 
The instances (sampled with the input-$l_x$-predict-$l_y$ rolling window and $l_x = l_y$, as illustrated in~\cref{fig:set}) are fed to the training procedure of these three baselines in chronological order. 
Besides, these three baselines employ the cumulative distribution function (CDF) for training, so the CDF needs to be reverted to the real distribution for testing.

For f-VAE, $\beta$-TCVAE, and VAE, since the dimensionality of different time series varies, we design a preprocess block to map the original time series into a tensor with the fix-sized dimensionality, which can further suit the VAEs well. 
The preprocess block consists of three nonlinear layers with the sizes of the hidden states: $\{128, 64, 32\}$. 
For NVAE, we keep the original settings suggested in~\cite{vahdat2020nvae} and use Gaussian distribution as the prior.
All the baselines are trained using early stopping, and the patience is set to 5.

\begin{table}[htbp]
  \caption{
  Performance comparisons of short-term and long-term TSF in real-world datasets in terms of MSE and CRPS.
  For MSE and CRPS, the lower, the better. The best results are in boldface.
  }     \label{longer-real}
    \centering
    \small
    \setlength\tabcolsep{2.0pt}
    \renewcommand{\arraystretch}{1.2}
    \begin{threeparttable}
    \begin{tabular}{c|c|ccccccccc}
    \toprule  
    \multicolumn{2}{c}{Model}&\ourmodel&NVAE&$\beta$-TCVAE&f-VAE&DeepAR&TimeGrad&GP-copula&VAE\\
    \midrule
    \multirow{8}{*}{\rotatebox{90}{Traffic}} & \multirow{2}{*}{8}&$\textbf{0.081}_{\pm.003}$&$1.300_{\pm.024}$&$1.003_{\pm.006}$&$0.982_{\pm.059}$&$3.895_{\pm.306}$&$3.695_{\pm.246}$&$4.299_{\pm.372}$&$0.794_{\pm.130}$\\
    ~&~&$\textbf{0.207}_{\pm.003}$&$0.593_{\pm.004}$&$0.894_{\pm.003}$&$0.666_{\pm.032}$&$1.391_{\pm.071}$&$1.410_{\pm.027}$&$1.408_{\pm.046}$&$0.759_{\pm.07}$\\
    \cline{2-10}
    ~& \multirow{2}{*}{16}&$\textbf{0.081}_{\pm.009}$&$1.271_{\pm.019}$&$0.997_{\pm.004}$&$0.998_{\pm.042}$&$4.140_{\pm.320}$&$3.495_{\pm.362}$&$4.575_{\pm.141}$&$0.632_{\pm.057}$\\
    ~&~&$\textbf{0.200}_{\pm.014}$&$0.589_{\pm.001}$&$0.893_{\pm.002}$&$0.692_{\pm.026}$&$1.338_{\pm.043}$&$1.329_{\pm.057}$&$1.506_{\pm.025}$&$0.671_{\pm.038}$\\
    \cline{2-10}
    ~& \multirow{2}{*}{32}&$\textbf{0.091}_{\pm.007}$&$0.126_{\pm.013}$&$1.254_{\pm0.019}$&$0.977_{\pm.002}$&$4.234_{\pm.139}$&$5.195_{\pm2.26}$&$3.717_{\pm.361}$&$0.735_{\pm.084}$\\
    ~&~&$\textbf{0.216}_{\pm.012}$&$0.422_{\pm.012}$&$0.937_{0.007}$&$0.882_{\pm.001}$&$ 1.367_{\pm.015}$&$1.565_{\pm.329}$&$1.342_{\pm.048}$&$0.735_{\pm.048}$\\
    \cline{2-10}
    ~& \multirow{2}{*}{64}&$\textbf{0.125}_{\pm.005}$&$1.263_{\pm0.014}$&$0.903_{\pm.111}$&$0.936_{\pm.190}$&$3.381_{\pm.130}$&$3.692_{\pm1.54}$&$3.492_{\pm.092}$&$0.692_{\pm.059}$\\
    ~&~&$\textbf{0.244}_{\pm.006}$&$0.940_{\pm0.005}$&$0.839_{\pm.062}$&$0.829_{\pm.078}$&$1.233_{\pm.027}$&$1.412_{\pm0.257}$&$1.367_{\pm.012}$&$0.710_{\pm.035}$\\
    \midrule
     
    \multirow{6}{*}{\rotatebox{90}{Electricity}} &  \multirow{2}{*}{8}&$\textbf{0.251}_{\pm.015}$&$1.134_{\pm.029}$&$0.901_{\pm.052}$&$0.893_{\pm.069}$&$2.934_{\pm.173}$&$2.703_{\pm.087}$&$2.924_{\pm.218}$&$0.853_{\pm.040}$\\
    ~&~&$\textbf{0.398}_{\pm.011}$&$0.542_{\pm.003}$&$0.831_{\pm.004}$&$0.809_{\pm.024}$&$1.244_{\pm.037}$&$1.208_{\pm.024}$&$1.249_{\pm.048}$&$0.795_{\pm.016}$\\
    \cline{2-10}
    ~& \multirow{2}{*}{16}&$\textbf{0.308}_{\pm.030}$&$1.150_{\pm.032}$&$0.850_{\pm.003}$&$0.807_{\pm.034}$&$2.803_{\pm.199}$&$2.770_{\pm.237}$&$3.065_{\pm.186}$&$0.846_{\pm.062}$\\
    ~&~&$\textbf{0.437}_{\pm.020}$&$0.531_{\pm.003}$&$0.814_{\pm.002}$&$0.782_{\pm.024}$&$1.220_{\pm.048}$&$1.240_{\pm.048}$&$1.307_{\pm.042}$&$0.793_{\pm.029}$\\
    \cline{2-10}
    ~& \multirow{2}{*}{32}&$\textbf{0.410}_{\pm.075}$&$1.302_{\pm0.011}$ &$0.844_{\pm.025}$&$0.861_{\pm.105}$&$2.402_{\pm.156}$&$2.640_{\pm.138}$&$2.880_{\pm.221}$&$0.841_{\pm.071}$\\
    ~&~&$\textbf{0.534}_{\pm.058}$&$0.944_{\pm0.005}$&$0.808_{\pm.005}$&$0.797_{\pm.037}$&$1.130_{\pm.055}$&$1.234_{\pm.027}$&$1.281_{\pm.054}$&$0.790_{\pm.026}$\\
    \midrule
     
    \multirow{8}{*}{\rotatebox{90}{Weather}} & \multirow{2}{*}{8} & $\textbf{0.169}_{\pm.022}$&$0.801_{\pm.024}$&$0.234_{\pm.042}$&$0.591_{\pm.198}$&$2.317_{\pm.357}$&$2.715_{\pm.189}$&$2.412_{\pm.761}$&$0.560_{\pm.192}$\\
    ~&~&$\textbf{0.357}_{\pm.024}$&$0.757_{\pm.013}$&$0.404_{\pm.040}$&$0.565_{\pm.080}$&$0.858_{\pm.078}$&$0.920_{\pm.013}$&$0.897_{\pm.115}$&$0.572_{\pm.077}$\\
     \cline{2-10}
    ~& \multirow{2}{*}{16}&$\textbf{0.187}_{\pm.047}$&$0.811_{\pm.016}$&$0.212_{\pm.012}$&$0.530_{\pm.167}$&$1.269_{\pm.187}$&$1.110_{\pm.083}$&$1.357_{\pm.145}$&$0.424_{\pm.141}$\\
    ~&~&$\textbf{0.361}_{\pm.046}$&$0.759_{\pm.009}$&$0.388_{\pm.014}$&$0.547_{\pm.067}$&$0.783_{\pm.059}$&$0.733_{\pm.016}$&$0.811_{\pm.032}$&$0.503_{\pm.068}$\\
    \cline{2-10}
    ~&\multirow{2}{*}{32}&$\textbf{0.203}_{\pm.008}$&$0.836_{\pm0.014}$&$0.439_{\pm.394}$&$0.337_{\pm.086}$&$2.518_{\pm.546}$&$1.178_{\pm.069}$&$1.065_{\pm.145}$&$0.329_{\pm.083}$\\
    ~&~&$\textbf{0.383}_{\pm.007}$&$0.777_{\pm0.007}$&$0.508_{\pm.176}$&$0.461_{\pm.031}$&$0.847_{\pm.036}$&$0.724_{\pm.021}$&$0.747_{\pm.035}$&$0.459_{\pm.045}$\\
    \cline{2-10}
    ~&\multirow{2}{*}{64}&$\textbf{0.191}_{\pm.022}$&$0.932_{0.020}$&$0.276_{\pm.026}$&$0.676_{\pm.484}$&$3.595_{\pm.956}$&$1.063_{\pm.061}$&$0.992_{\pm.114}$&$0.721_{\pm.496}$\\
    ~&~&$\textbf{0.358}_{\pm.044}$&$0.836_{0.009}$&$0.463_{\pm.026}$&$0.612_{\pm.176}$&$0.994_{\pm.100}$&$0.696_{\pm.011}$&$0.699_{\pm.016}$&$0.635_{\pm.204}$\\
    \midrule
     
    \multirow{6}{*}{\rotatebox{90}{ETTm1}} &  \multirow{2}{*}{8} & $\textbf{0.527}_{\pm.073}$&$0.921_{\pm.026}$&$1.538_{\pm.254}$&$2.326_{\pm.445}$&$2.204_{\pm.420}$&$1.877_{\pm.245}$&$2.024_{\pm.143}$&$2.375_{\pm.405}$\\
    ~&~&$\textbf{0.557}_{0.048}$&$0.760_{\pm.026}$&$1.015_{\pm.112}$&$1.260_{\pm.167}$&$0.984_{\pm.074}$&$0.908_{\pm.038}$&$0.961_{\pm.027}$&$1.258_{\pm.104}$\\
    \cline{2-10}
    ~& \multirow{2}{*}{16}&$\textbf{0.968}_{\pm.104}$&$1.100_{\pm.032}$&$1.744_{\pm.100}$&$2.339_{\pm.270}$&$2.350_{\pm.170}$&$2.032_{\pm.234}$&$2.486_{\pm.207}$&$2.321_{\pm.469}$\\
    ~&~&$\textbf{0.821}_{\pm.072}$&$0.822_{\pm.026}$&$1.104_{\pm.041}$&$1.249_{\pm.088}$&$0.974_{\pm.016}$&$0.919_{\pm.031}$&$0.984_{\pm.016}$&$1.259_{\pm.132}$\\
    \cline{2-10}
    ~& \multirow{2}{*}{32}&$\textbf{0.707}_{\pm.061}$&$1.298_{\pm.028} $&$1.438_{\pm.429}$&$2.563_{\pm.358}$&$4.855_{\pm.179}$&$1.251_{\pm.133}$&$1.402_{\pm.187}$&$2.660_{\pm.349}$\\
    ~&~&$\textbf{0.697}_{\pm.040}$&$0.893_{\pm.010}$&$0.953_{\pm.173}$&$1.330_{\pm.104}$&$1.787_{\pm.029}$&$0.822_{\pm.032}$&$0.844_{\pm.043}$&$1.367_{\pm.083}$\\
    \midrule
     
    \multirow{8}{*}{\rotatebox{90}{ETTh1}} & \multirow{2}{*}{8} &$\textbf{0.292}_{\pm.036}$&$0.483_{\pm.017}$&$0.703_{\pm.054}$&$0.870_{\pm.134}$&$3.451_{\pm.335}$&$4.259_{\pm1.13}$&$4.278_{\pm1.12}$&$1.006_{\pm.281}$\\
    ~&~&$\textbf{0.424}_{\pm.033}$&$0.461_{\pm.011}$&$0.644_{\pm.038}$&$0.730_{\pm.060}$&$1.194_{\pm.034}$&$1.092_{\pm.028}$&$1.169_{\pm.055}$&$0.762_{\pm.115}$\\
    \cline{2-10}
    ~& \multirow{2}{*}{16}&$\textbf{0.374}_{\pm.061}$&$0.488_{\pm.010}$&$0.681_{\pm.018}$&$0.983_{\pm.139}$&$1.929_{\pm.105}$&$1.332_{\pm.125}$&$1.701_{\pm.088}$&$0.681_{\pm.104}$\\
    ~&~&$0.488_{\pm.039}$&$\textbf{0.463}_{\pm.018}$&$0.640_{\pm.008}$&$0.760_{\pm.062}$&$1.029_{\pm.030}$&$0.879_{\pm.037}$&$0.999_{\pm.023}$&$0.641_{\pm.055}$\\
    \cline{2-10}
    ~& \multirow{2}{*}{32}&$\textbf{0.334}_{\pm.008}$&$0.464_{\pm0.007}$&$0.477_{\pm.035}$&$0.669_{\pm.092}$&$6.153_{\pm.715}$&$1.514_{\pm.042}$&$1.922_{\pm.032}$&$0.578_{\pm.062}$\\
    ~&~&$\textbf{0.461}_{\pm.004}$&$0.543_{\pm0.004}$&$0.537_{\pm.019}$&$0.646_{\pm.048}$&$1.689_{\pm.112}$&$0.925_{\pm.016}$&$1.068_{\pm.011}$&$0.597_{\pm.035}$\\
    \cline{2-10}
    ~&\multirow{2}{*}{64}&$\textbf{0.349}_{\pm.039}$& $0.425_{\pm.006}$&$0.418_{\pm.021}$&$0.484_{\pm.051}$&$2.419_{\pm.520}$&$1.150_{0.118}$&$1.654_{\pm.117}$&$0.463_{\pm.081}$\\
    ~&~&$\textbf{0.473}_{\pm.024}$&$0.523_{0.004}$&$0.517_{\pm.013}$&$0.551_{\pm.027}$&$1.223_{\pm.127}$&$0.835_{\pm.045}$&$0.987_{\pm.036}$&$0.546_{\pm.042}$\\
    \midrule
    
    \multirow{8}{*}{\rotatebox{90}{Wind}} & \multirow{2}{*}{8} & $\textbf{0.681}_{\pm.075}$&$1.854_{\pm.032}$&$1.321_{\pm.379}$&$1.942_{\pm.101}$&$12.53_{\pm2.25}$&$12.67_{\pm1.75}$&$11.35_{\pm6.61}$&$2.006_{\pm.145}$\\
    ~&~&$\textbf{0.596}_{\pm.052}$&$1.223_{\pm.014}$&$0.863_{\pm.143}$&$1.067_{\pm.086}$&$1.370_{\pm.107}$&$1.440_{\pm.059}$&$1.305_{\pm.369}$&$1.103_{\pm.100}$\\
    \cline{2-10}
    ~& \multirow{2}{*}{16}&$1.033_{\pm.062}$&$1.955_{\pm.015}$&$\textbf{0.894}_{\pm.038}$&$1.262_{\pm.178}$&$13.96_{\pm.1.53}$&$12.86_{\pm2.60}$&$13.79_{\pm5.37}$&$1.138_{\pm.205}$\\
    ~&~&$\textbf{0.757}_{\pm.053}$&$1.247_{\pm.011}$&$0.785_{\pm.037}$&$0.843_{\pm.066}$&$1.347_{\pm.060}$&$1.240_{\pm.070}$&$1.261_{\pm.171}$&$0.862_{\pm.092}$\\
    \cline{2-10}
    ~& \multirow{2}{*}{32}&$\textbf{1.224}_{\pm.060}$&$1.784_{\pm.011}$&$1.266_{\pm.006}$&$1.434_{\pm.126}$&$5.398_{\pm.179}$&$13.10_{\pm.955}$&$15.33_{1.904}$&$1.480_{\pm.072}$\\
    ~&~&$\textbf{0.869}_{\pm.074}$&$1.200_{\pm.007}$&$0.872_{\pm.010}$&$0.920_{\pm.077}$&$1.434_{\pm.013}$&$1.518_{\pm.020}$&$ 1.614_{\pm.118}$&$0.987_{\pm.010}$\\
    \cline{2-10}
    ~&\multirow{2}{*}{64}&$0.902_{\pm.024}$& $1.652_{\pm.010}$&$\textbf{0.786}_{\pm.022}$&$0.898_{\pm.095}$&$4.403_{\pm.301}$&$3.857_{\pm.597}$&$3.564_{\pm.293}$&$1.374_{\pm1.02}$\\
    ~&~&$0.761_{\pm.021}$&$1.167_{\pm.005}$&$\textbf{0.742}_{\pm.017}$&$0.789_{\pm.048}$&$1.361_{\pm.021}$&$1.110_{\pm.143}$&$1.152_{\pm.081}$&$0.842_{\pm.215}$\\
    \bottomrule
    \end{tabular}
    \end{threeparttable}
\end{table}

\begin{table}[ht]
    \caption{
    Performance comparisons of TSF in 100\%-Electricity and 100\%-ETTm1 datasets in terms of MSE and CRPS. 
    The best results are highlighted in boldface. 
    }   \label{full_dataset}
    \centering
    \small
    \setlength\tabcolsep{2.0pt}
    \renewcommand{\arraystretch}{1.2}
    \begin{tabular}{c|c|ccccccccc}
    \toprule  
    \multicolumn{2}{c}{Model}&\ourmodel&NVAE&$\beta$-TCVAE&f-VAE&DeepAR&TimeGrad&GP-copula&VAE\\
   \midrule
    \multirow{4}{*}{\rotatebox{90}{Electricity}} & \multirow{2}{*}{16}& $\textbf{0.330}_{\pm.033}$ & $1.408_{\pm.015} $ & $0.801_{\pm.001}$& $0.765_{\pm.026}$& $33.93_{\pm1.85}$ & $ 46.69_{\pm3.13}$& $50.25_{\pm4.39}$&$0.680_{\pm.022}$\\
    ~&~&$\textbf{0.445}_{\pm.020}$&$0.999_{\pm.006}$ & $0.723_{\pm.001}$& $0.710_{\pm.013}$& $2.650_{\pm.030}$ &$2.702_{\pm.079}$ &
    $2.796_{\pm.072}$&$0.675_{\pm.008}$\\
    \cline{2-10}
   
    ~&\multirow{2}{*}{32}& $\textbf{0.336}_{\pm.017}$&$1.403_{\pm.014}$ & $0.802_{\pm.001}$ &$0.748_{\pm.033}$& $46.10_{\pm2.00}$&$30.94_{\pm1.70}$ & $ 32.13_{\pm1.96}$& $0.727_{\pm.033}$ \\
    ~&~& $\textbf{0.444}_{\pm.015}$& $0.997_{\pm.007}$& $0.724_{\pm.001}$ & $0.703_{\pm.016}$& $2.741_{\pm.011}$&$ 2.476_{\pm.042}$ &$2.591_{\pm.064}$& $0.692_{\pm.014}$\\
     \midrule
     
    \multirow{4}{*}{\rotatebox{90}{ETTm1}} & \multirow{2}{*}{16} &$\textbf{0.018}_{\pm.002}$ & $2.577_{\pm.047}$& $0.918_{\pm.015}$ & $1.285_{\pm.236}$ & $73.82_{\pm3.25}$& $68.26_{\pm2.04}$ & $66.97_{\pm2.02}$ & $1.335_{\pm.156}$\\
    ~&~&$\textbf{0.102}_{\pm.003}$ & $1.509_{0.016}$& $ 0.766_{\pm.005}$ & $0.911_{\pm.090}$ & $1.136_{\pm.013}$ & $1.153_{\pm.019}$ & $1.111_{\pm.016}$ & $0.923_{\pm.056}$\\
    
    \cline{2-10}
    ~& \multirow{2}{*}{32}& $\textbf{0.034}_{\pm.001}$& $2.622_{\pm.057} $& $0.929_{\pm.010}$ & $1.420_{\pm.073}$ &$68.11_{\pm2.60}$ &$53.47_{\pm26.1}$ & $ 63.67_{\pm1.14}$ & $1.223_{\pm.213}$\\
    ~&~&$\textbf{0.144}_{\pm.006}$&$1.524_{\pm.018}$ & $0.770_{\pm.004}$ & $0.960_{\pm.021}$ &$1.121_{\pm.024}$ & $1.083_{\pm.109}$ & $1.097_{\pm.008}$ & $0.888_{\pm.082}$\\
    \bottomrule
    \end{tabular}
\end{table}

\section{Supplementary Experimental Results} \label{extra}

\subsection{Comparisons of Predictive Performance for TSF Under Different Settings}   \label{appendix:supp-main-results}

\textbf{Longer-Term Time Series Forecasting.~~}
To further inspect the performance of our method, we additionally conduct more experiments for longer-term time series forecasting. 
In particular, by configuring the output length with 32 and 64\footnote{The length of the input time series is the same as the output time series.}, 
we compare \ourmodel~to other baselines in terms of MSE and CRPS, and the results (including short-term and long-term) are reported in~\cref{longer-real}.
We can conclude that \ourmodel~also outperforms the competitive baselines consistently under the longer-term forecasting settings.

\textbf{Time Series Forecasting in Full Datasets.~~}
Moreover, we evaluate the predictive performance for time series forecasting in  two ``full-version'' datasets, 
i.e. 100\%-Electricity and 100\%-ETTm1.
The split of train/validation/test is 7:1:2 which is the same as the main experiments. 
The comparisons in terms of MSE and CRPS can be found in~\cref{full_dataset}. 
With sufficient data, compared to previous state-of-the-art generative models, the MSE and CRPS reductions of our method are also satisfactory under different settings (including input-16-predict-16 and input-32-predict-32). 
For example, in the Electricity dataset, compared to the second best results, \ourmodel~achieves 52\% (0.680 $ \rightarrow $ 0.330) and 54\% (0.727 $ \rightarrow $ 0.336) MSE reductions, and 34\% (0.675 $\rightarrow$ 0.445) and 36\% (0.692 $\rightarrow$ 0.444) CRPS reductions, under input-16-predict-16 and input-32-predict-32 settings, respectively.

\begin{figure}[htbp]
    \centering
    \begin{minipage}[c]{\textwidth}
    \begin{subfigure}[t]{0.24\textwidth}
      \includegraphics[width=\textwidth]{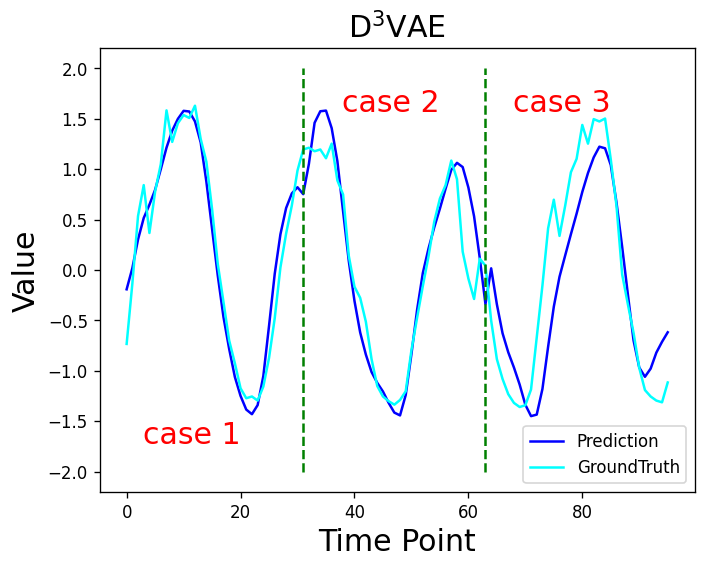}
    \end{subfigure}
    \begin{subfigure}[t]{0.24\textwidth}
      \includegraphics[width=\textwidth]{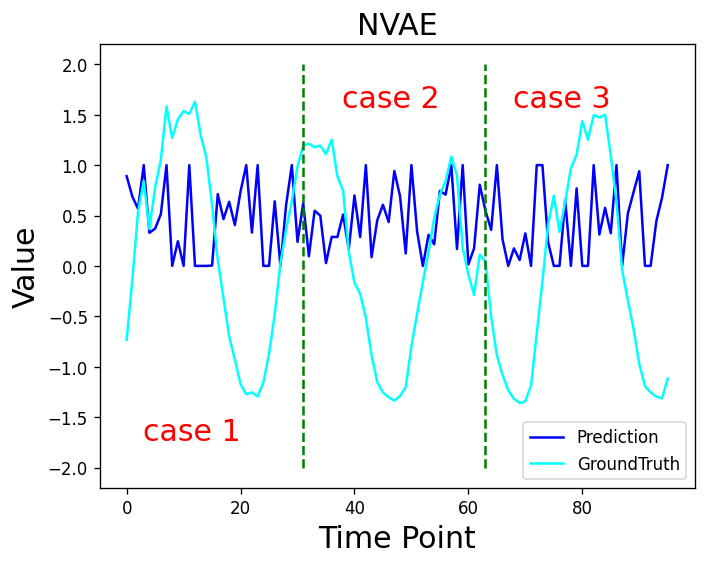}
    \end{subfigure}
    \begin{subfigure}[t]{0.24\textwidth}
      \includegraphics[width=\textwidth]{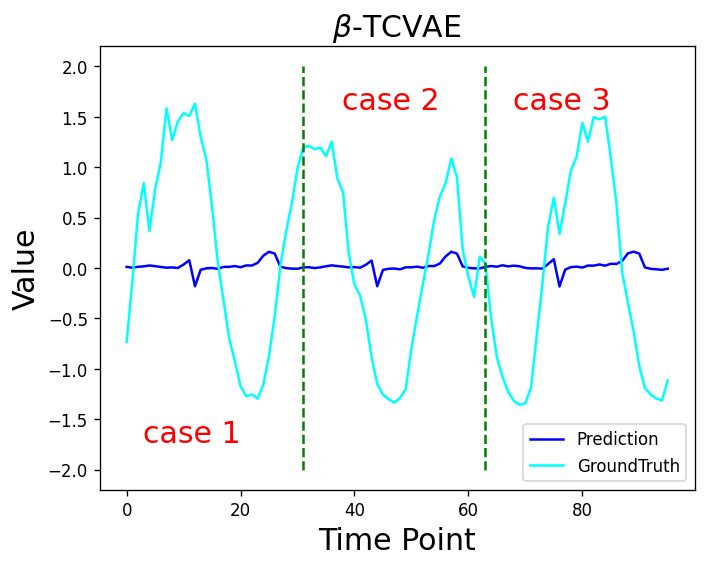}
    \end{subfigure}
    \begin{subfigure}[t]{0.24\textwidth}
      \includegraphics[width=\textwidth]{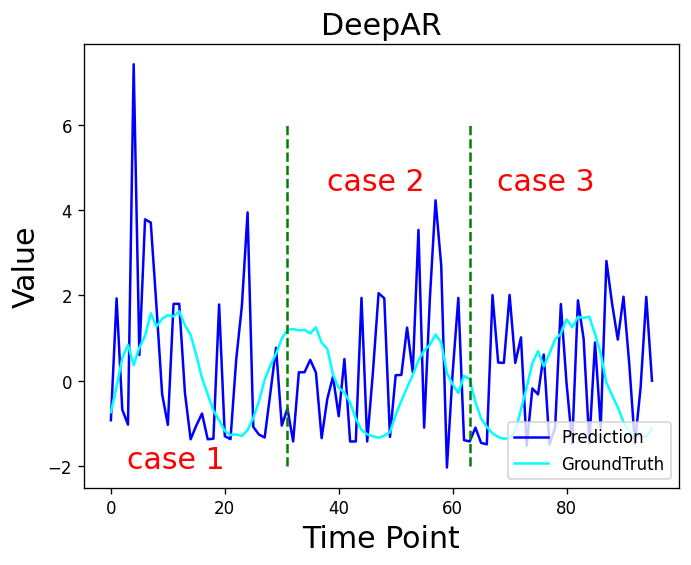}
    \end{subfigure}
    \end{minipage}
    
    \begin{minipage}[c]{\textwidth}
    \begin{subfigure}[t]{0.24\textwidth}
      \includegraphics[width=\textwidth]{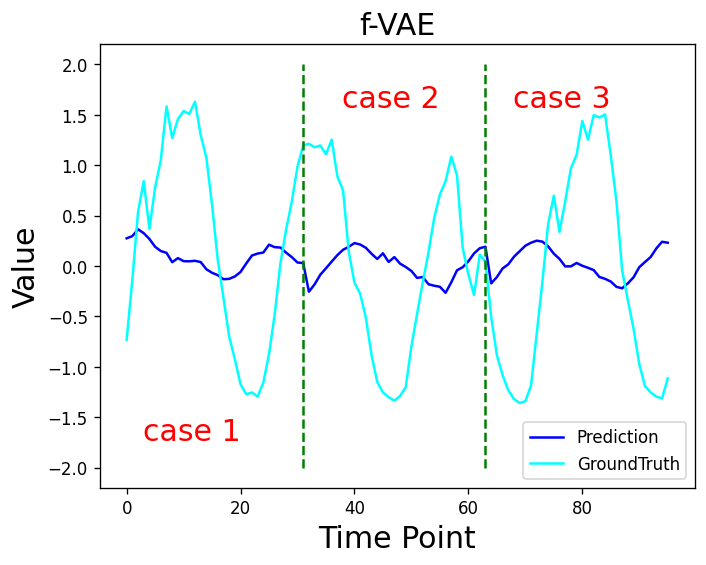}
    \end{subfigure}
    \begin{subfigure}[t]{0.24\textwidth}
      \includegraphics[width=\textwidth]{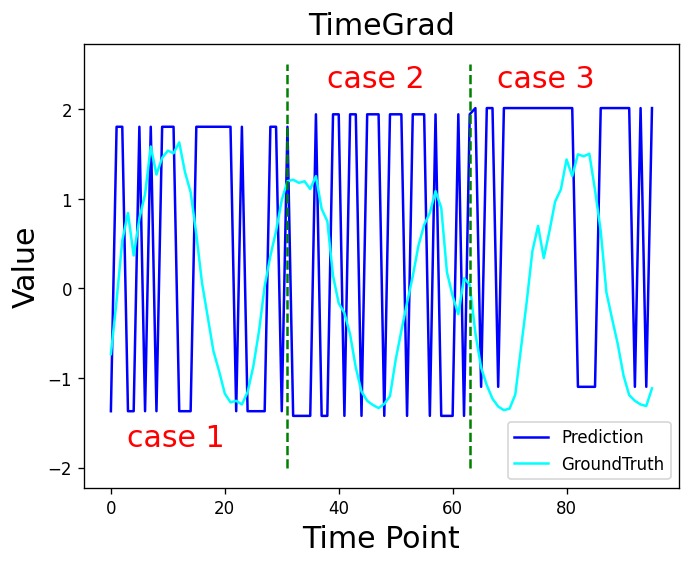}
    \end{subfigure}
    \begin{subfigure}[t]{0.24\textwidth}
      \includegraphics[width=\textwidth]{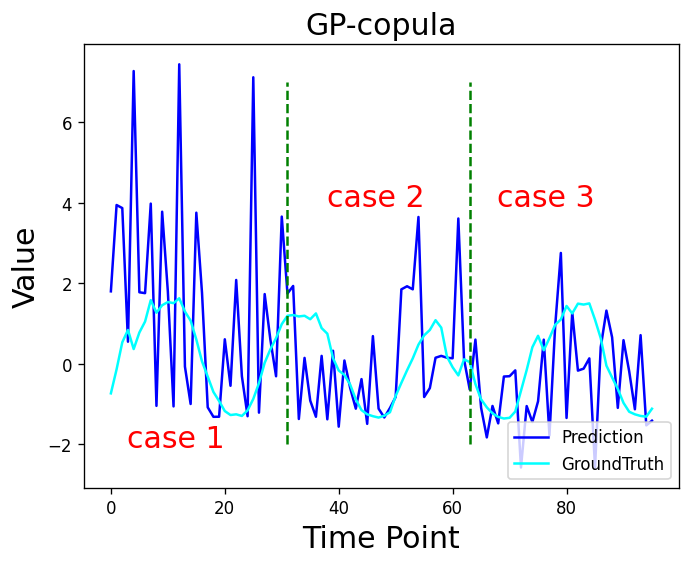}
    \end{subfigure}
    \begin{subfigure}[t]{0.24\textwidth}
      \includegraphics[width=\textwidth]{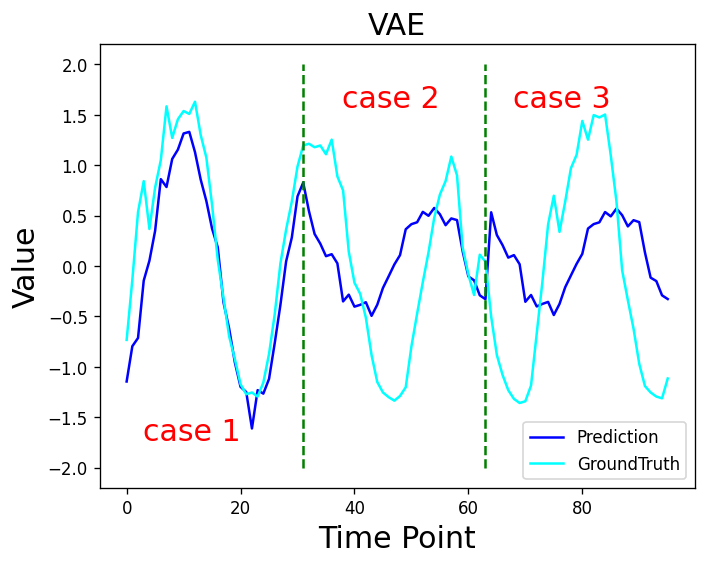}
    \end{subfigure}
    \end{minipage}
    \caption{
    The case study of forecasting results on the Traffic dataset under input-32-predict-32 settings. 
    Only the last dimension is plotted. 
    To demonstrate the forecasting results in a long range, we show the predictions of three cases ordered chronologically without overlapping. 
    } \label{fig:show_traff}
\end{figure}

\begin{figure}[htbp]
    \centering
    \begin{minipage}[c]{\textwidth}
    \begin{subfigure}[t]{0.24\textwidth}
      \includegraphics[width=\textwidth]{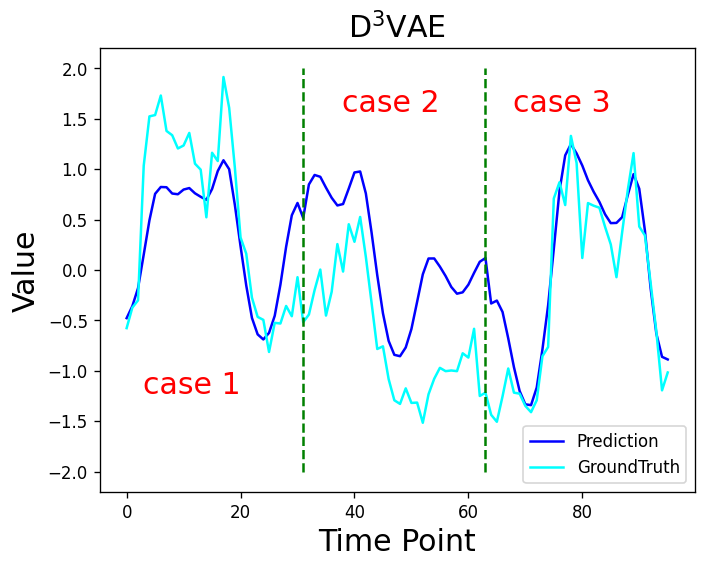}
    \end{subfigure}
    \begin{subfigure}[t]{0.24\textwidth}
      \includegraphics[width=\textwidth]{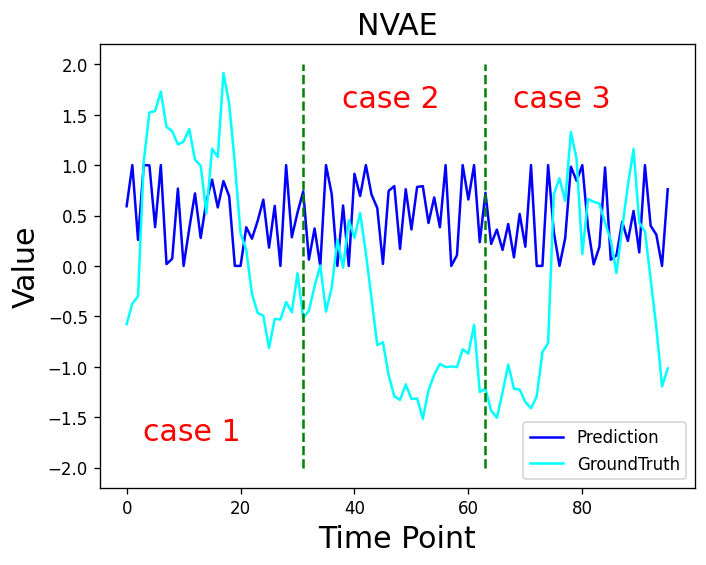}
    \end{subfigure}
    \begin{subfigure}[t]{0.24\textwidth}
      \includegraphics[width=\textwidth]{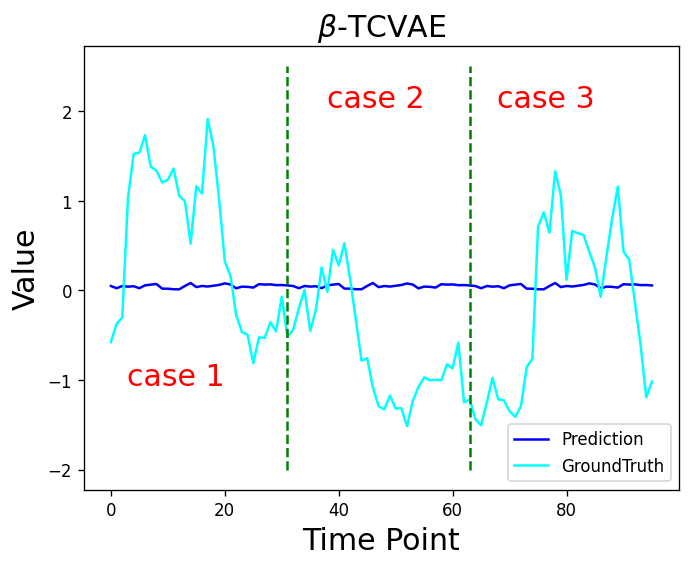}
    \end{subfigure}
    \begin{subfigure}[t]{0.24\textwidth}
      \includegraphics[width=\textwidth]{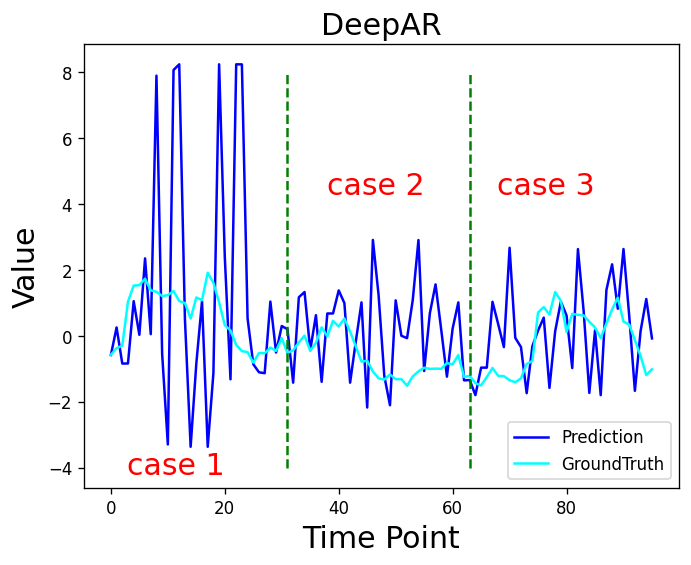}
    \end{subfigure}
    \end{minipage}
    
    \begin{minipage}[c]{\textwidth}
    \begin{subfigure}[t]{0.24\textwidth}
      \includegraphics[width=\textwidth]{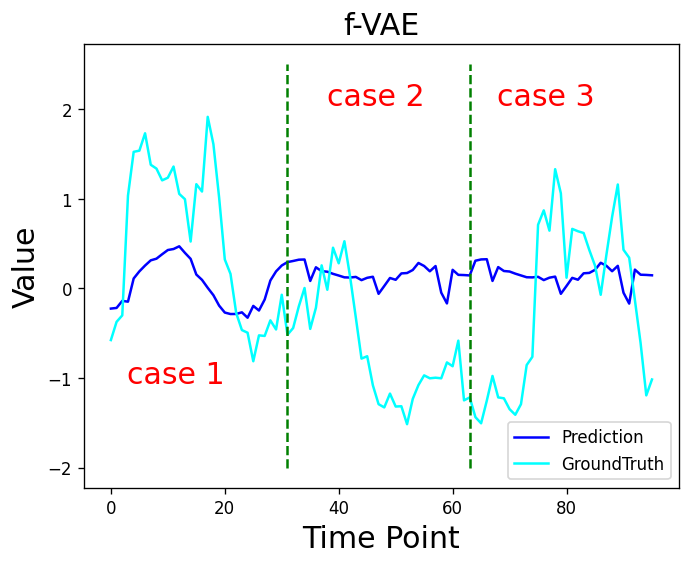}
    \end{subfigure}
    \begin{subfigure}[t]{0.24\textwidth}
      \includegraphics[width=\textwidth]{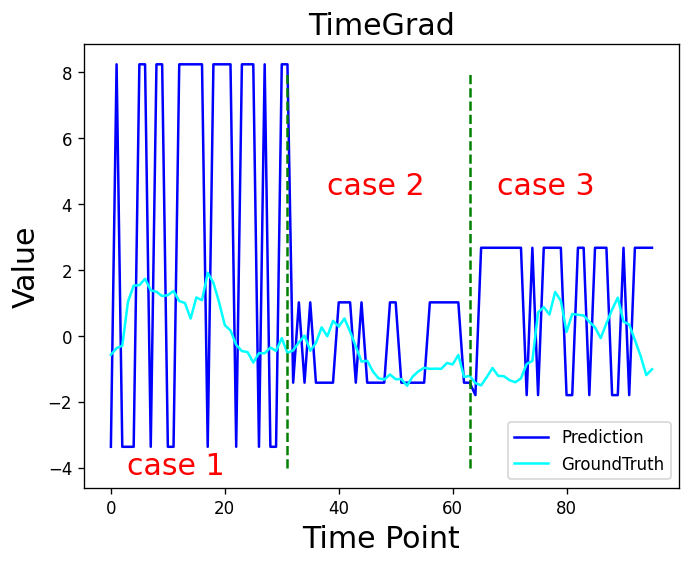}
    \end{subfigure}
    \begin{subfigure}[t]{0.24\textwidth}
      \includegraphics[width=\textwidth]{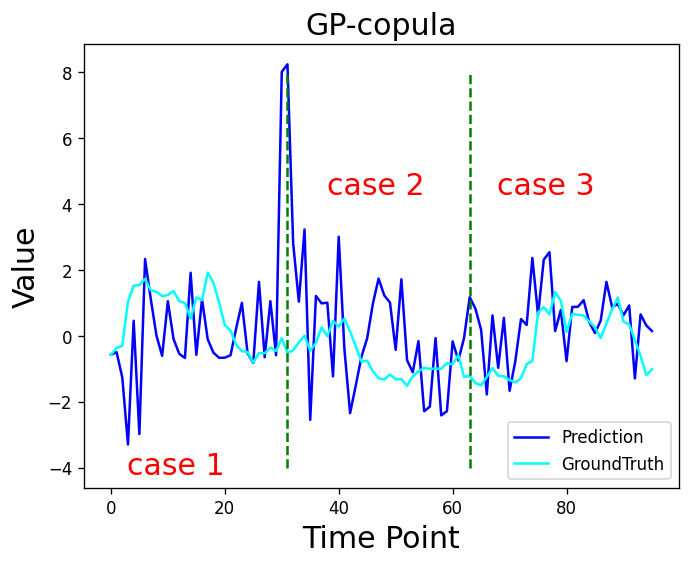}
    \end{subfigure}
    \begin{subfigure}[t]{0.24\textwidth}
      \includegraphics[width=\textwidth]{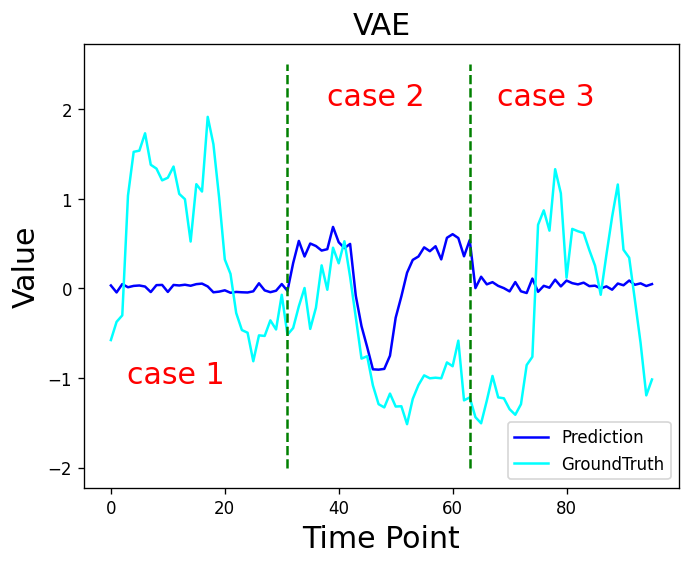}
    \end{subfigure}
    \end{minipage}
    \caption{
    Case study of the forecasting results from the Electricity dataset (same settings as \cref{fig:show_traff}).
    }  \label{fig:show_elec}
    \vspace{-1ex}
\end{figure}

\begin{figure}[t]
    \centering
    \begin{minipage}[c]{\textwidth}
     \begin{subfigure}[t]{0.24\textwidth}
      \includegraphics[width=\textwidth]{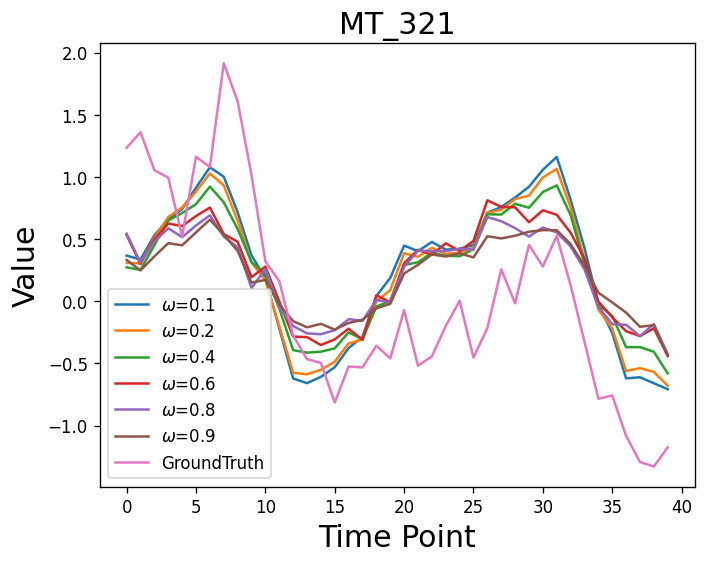}
    \end{subfigure}
    \begin{subfigure}[t]{0.24\textwidth}
      \includegraphics[width=\textwidth]{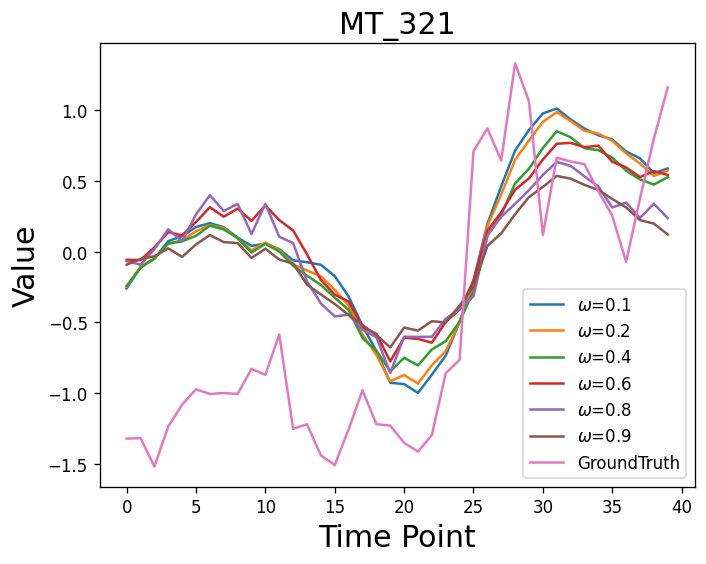}
    \end{subfigure}
    \begin{subfigure}[t]{0.24\textwidth}
      \includegraphics[width=\textwidth]{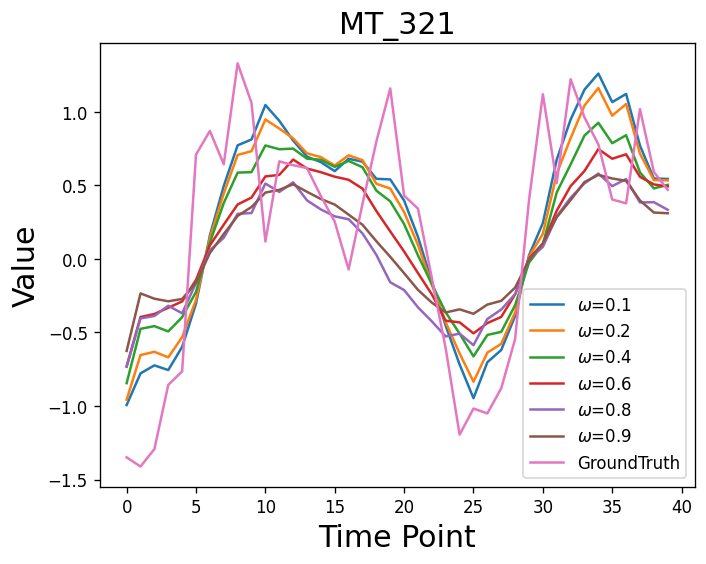}
    \end{subfigure}
    \begin{subfigure}[t]{0.24\textwidth}
      \includegraphics[width=\textwidth]{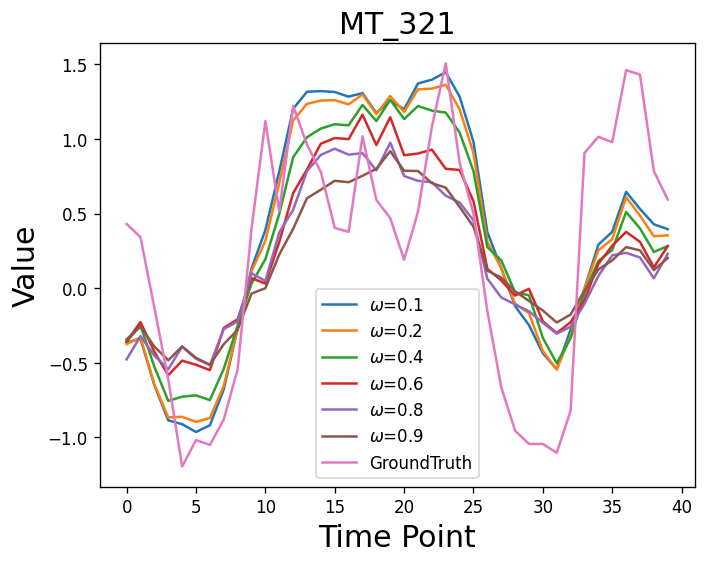}
    \end{subfigure}
    \caption{
    Forecasting results (under the input-40-predict-40 setting) of a case from the Electricity dataset with $\omega$ increasing from $0.1$ to $0.9$.
    }   \label{fig:omega_e}
    \vspace{2ex}
    \end{minipage}
    \begin{minipage}[c]{\textwidth}
    \begin{subfigure}[t]{0.24\textwidth}
      \includegraphics[width=\textwidth]{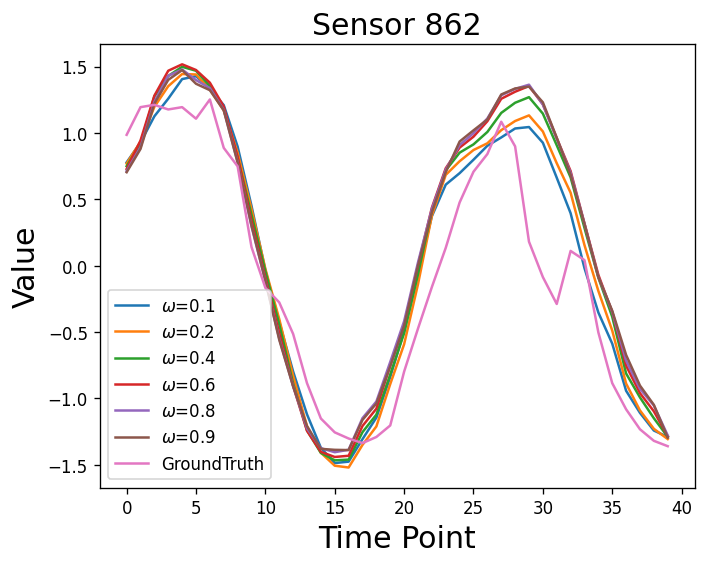}
    \end{subfigure}
    \begin{subfigure}[t]{0.24\textwidth}
      \includegraphics[width=\textwidth]{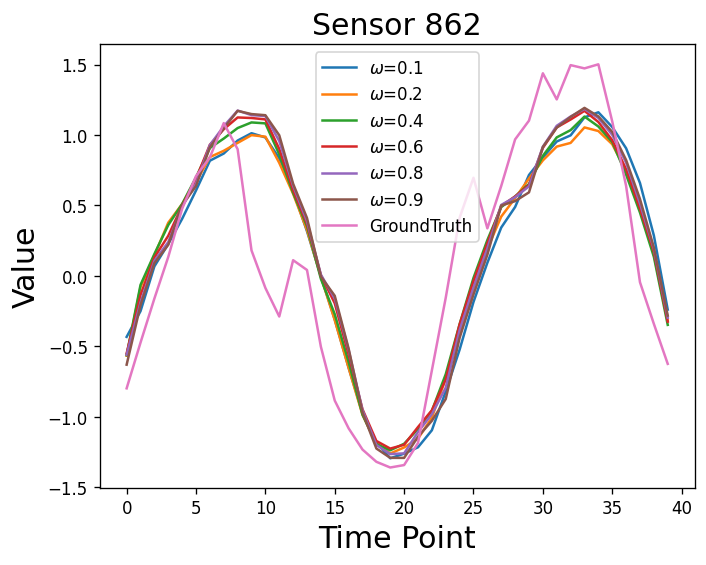}
    \end{subfigure}
    \begin{subfigure}[t]{0.24\textwidth}
      \includegraphics[width=\textwidth]{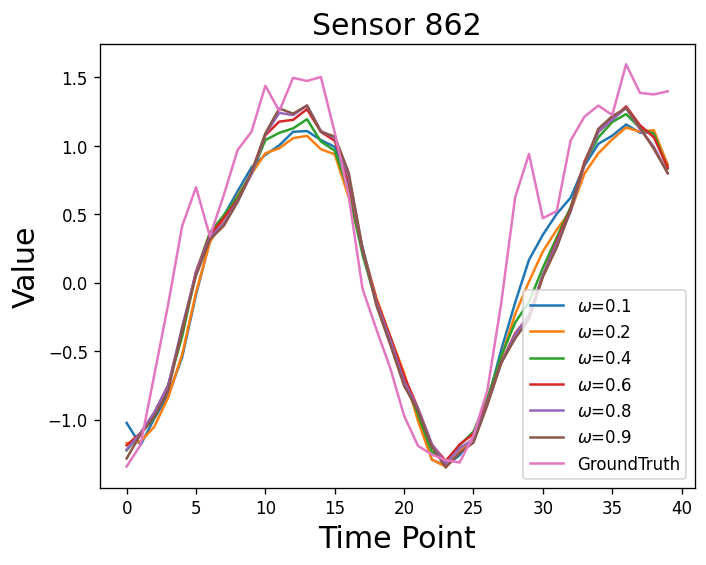}
    \end{subfigure}
    \begin{subfigure}[t]{0.24\textwidth}
      \includegraphics[width=\textwidth]{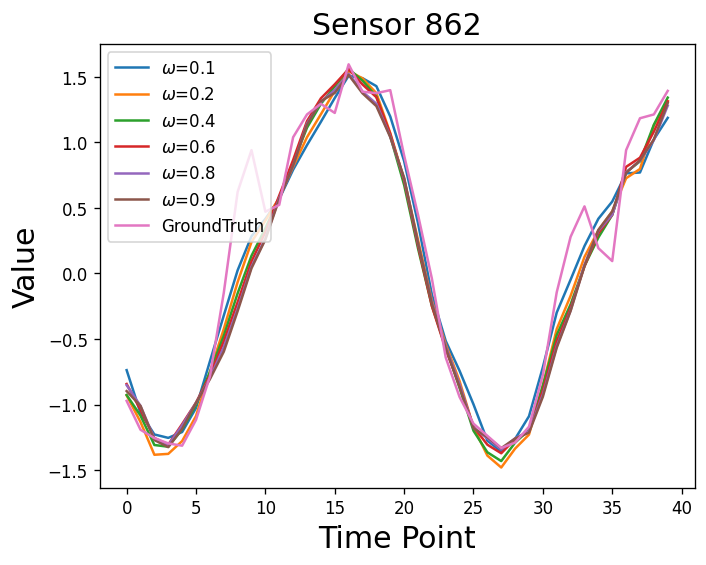}
    \end{subfigure}
    \caption{
    Forecasting results of a case from the Traffic dataset under the input-40-predict-40 setting. 
    }
    \label{fig:omega_t}
    \vspace{2ex}
    \end{minipage}
    \begin{minipage}[c]{\textwidth}
    \begin{subfigure}[t]{0.31\textwidth}
      \includegraphics[width=\textwidth]{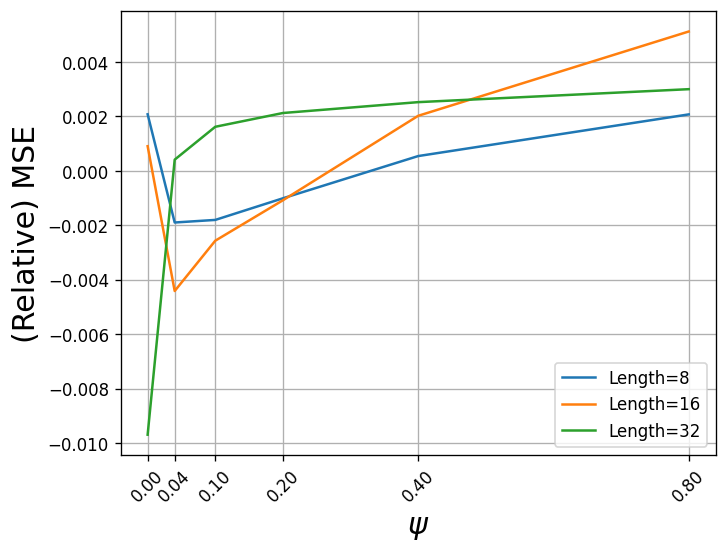}
    \end{subfigure}
    \begin{subfigure}[t]{0.325\textwidth}
      \includegraphics[width=\textwidth]{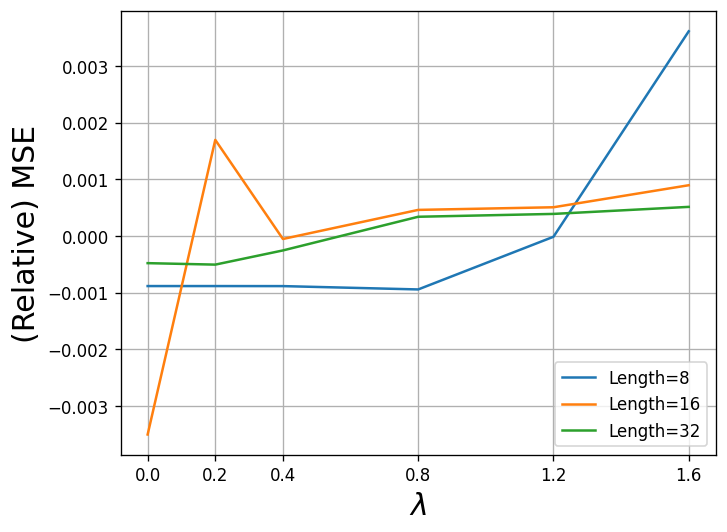}
    \end{subfigure}
    \begin{subfigure}[t]{0.31\textwidth}
      \includegraphics[width=\textwidth]{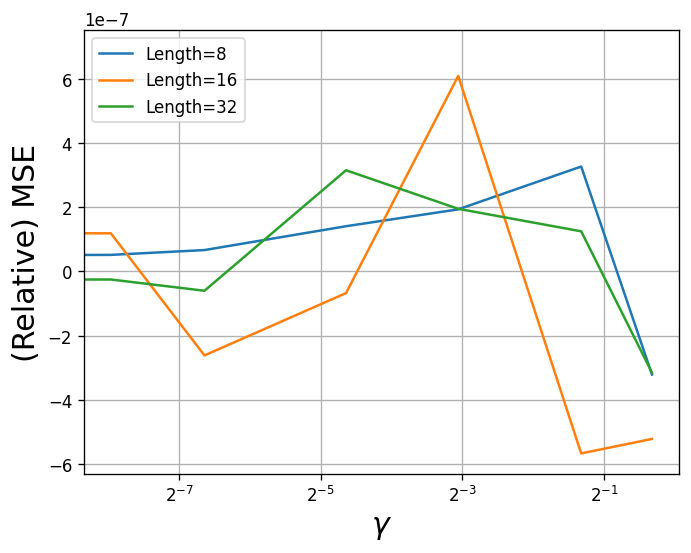}
    \end{subfigure}
    \caption{
    Sensitivity analysis of the trade-off hyperparameters in reconstruction loss $\mathcal{L}$.
    To highlight the changes in prediction performance against hyperparameters, the relative value of MSE is used. 
    }
    \label{fig:para}
    \vspace{-2ex}
    \end{minipage}
\end{figure}

\subsection{Case Study}

We showcase the prediction results of our model and seven baseline models on the Traffic and Electricity datasets in~\cref{fig:show_traff,fig:show_elec}.
Our model can provide the most accurate forecasting results regarding trends and variations.

\subsection{The Effect of Scale Parameter $\omega$} \label{App_omega}

We demonstrate the forecasting results with different values of $\omega$ on  Electricity and Traffic datasets, and the results are plotted in~\cref{fig:omega_e,fig:omega_t}. 
It can be depicted that larger or smaller $\omega$ would lead to deviated prediction, which is far from the ground truth. 
Therefore, the value of $\omega$ does affect the prediction performance, which should be tuned properly.

\begin{table}[tbp]
    \begin{minipage}[t]{\textwidth}
    \centering
    \small
    \setlength\tabcolsep{3.0pt}
    \renewcommand{\arraystretch}{1.2}
    \caption{
    Performance comparisons of \ourmodel~w.r.t. varying the length of (input and output) time series and the data size. 
    The results are reported on the Electricity dataset.
    }
    \begin{tabular}{c|c|ccccccc}
    \toprule
    \multirow{2}{*}{Length} & 
    \multirow{2}{*}{Metric}     & 
    \multicolumn{7}{c}{Percentage of Full Electricity Data}
    \\
    \cline{3-9}
       &  & 100\% & 80\% & 60\% & 40\% & 20\% & 10\% & 5\%\\
       \midrule
        \multirow{2}{*}{8} 
        &MSE& $0.258_{\pm.019}$& $0.227_{\pm.016}$ & $0.368_{\pm.019}$ & $0.389_{\pm.034}$&$3.861_{\pm.480}$&$0.693_{\pm.223}$ &$0.206_{\pm.018}$\\
        ~& CRPS & $0.383_{\pm.015}$ & $0.355_{\pm.015}$ & $0.453_{\pm.009}$ &  $0.504_{\pm.034}$&$1.728_{\pm.110}$&$0.673_{\pm.132}$&$0.352_{\pm.016}$\\
        \midrule
        \multirow{2}{*}{16}& MSE & $0.330_{\pm.033}$& $0.253_{\pm.018}$&$0.343_{\pm.024}$&$0.463_{\pm.089}$&$4.428_{\pm.694}$&$0.401_{\pm.068}$&$0.247_{\pm.056}$\\
        ~ & CRPS & $0.445_{\pm.020}$ & $0.373_{\pm.014}$&$0.433_{\pm.015}$&$0.562_{\pm.049}$&$1.858_{\pm.147}$&$0.496_{\pm.047}$&$0.378_{\pm.036}$\\
        \midrule
        \multirow{2}{*}{32}& MSE & $0.336_{\pm.017}$& $0.300_{\pm.039}$&$0.484_{\pm.048}$&$0.739_{\pm.209}$&$5.029_{\pm.811}$&$0.884_{\pm.237}$&$0.304_{\pm.094}$\\
        ~ & CRPS & $0.444_{\pm.015}$& $0.413_{\pm.034}$&$0.537_{\pm.025}$&$0.693_{\pm.099}$&$1.989_{\pm.172}$&$0.723_{\pm.112}$&$0.418_{\pm.065}$\\
    \bottomrule
    \end{tabular}
    \label{tab:per_length}
    \vspace{1ex}
    \end{minipage}
    \begin{minipage}[t]{\textwidth}
    \centering
    \small
    \setlength\tabcolsep{3.0pt}
    \renewcommand{\arraystretch}{1.2}
    \captionof{table}{
    Performance comparisons of \ourmodel~w.r.t. varying the length of (input and output) time series and the data size. 
    The results are reported on the Traffic dataset.
    }
    \begin{tabular}{c|c|ccccccc}
    \toprule
    \multirow{2}{*}{Length} & 
    \multirow{2}{*}{Metric}     & 
    \multicolumn{7}{c}{Percentage of Full Traffic Data}
    \\
    \cline{3-9}
       &  & 100\% & 80\% & 60\% & 40\% & 20\% & 10\% & 5\%\\
       \midrule
        \multirow{2}{*}{8} 
        &MSE& $0.370_{\pm.021}$& $0.215_{\pm.016}$ & $0.063_{\pm.002}$ & $0.062_{\pm.002}$&$0.054_{\pm.004}$&$0.210_{\pm.012}$ &$0.081_{\pm.003}$\\
        ~& CRPS & $0.415_{\pm.013}$ & $0.347_{\pm.015}$ & $0.184_{\pm.003}$ &  $0.179_{\pm.005}$&$0.172_{\pm.008}$&$0.251_{\pm.005}$&$0.207_{\pm.003}$\\
        \midrule
        \multirow{2}{*}{16}& MSE & $0.272_{\pm.007}$& $0.189_{\pm.006}$&$0.063_{\pm.001}$&$0.058_{\pm.003}$&$0.056_{\pm.003}$&$0.178_{\pm.006}$&$0.081_{\pm.009}$\\
        ~ & CRPS & $0.334_{\pm.009}$ & $0.321_{\pm.008}$&$0.180_{\pm.002}$&$0.168_{\pm.006}$&$0.169_{\pm.005}$&$0.239_{\pm.007}$&$0.200_{\pm.003}$\\
        \midrule
        \multirow{2}{*}{32}& MSE & $0.307_{\pm.015}$& $0.197_{\pm.005}$&$0.064_{\pm.002}$&$0.063_{\pm.002}$&$0.056_{\pm.004}$&$0.191_{\pm.011}$&$0.091_{\pm.007}$\\
        ~ & CRPS & $0.363_{\pm.008}$& $0.335_{\pm.004}$&$0.179_{\pm.002}$&$0.179_{\pm.003}$&$0.170_{\pm.005}$&$0.235_{\pm.008}$&$0.216_{\pm.012}$\\
    \bottomrule
    \end{tabular}
    \label{tab:per_length:traffic}
    \end{minipage}
\end{table}

\subsection{Sensitivity Analysis of Trade-off Parameters in Reconstruction Loss $\mathcal{L}$} \label{para_effective}

To examine the effect of the trade-off hyperparameters in loss $\mathcal{L}$, we plot the mean square error (MSE) against different values of trade-off parameters, i.e., $ \psi $, $\lambda$ and $\gamma$, in the Traffic dataset. 
Note that the relative value of MSE is plotted to ensure the difference is distinguishable. 
This experiment is conducted under different settings: input-8-predict-8, input-16-predict-16, and input-32-predict-32. 
For $\psi$, the value ranges from $0$ to $0.8$, $\lambda$ ranges from $0$ to $1.6$, and $\gamma$ ranges from $0$ to $0.5$. 
The results are shown in~\cref{fig:para}.
We can see that the model's performance varies slightly as the trade-off parameters take different values, which shows that our model is robust enough against different trade-off parameters.

\subsection{Scalability Analysis of Varying Time Series Length and Dataset Size}

We additionally investigate the scalability of \ourmodel ~against different lengths of the time series and varying amounts of available data. 
The experiments are conducted on the Electricity and Traffic datasets, and the results are reported  in~\cref{tab:per_length,tab:per_length:traffic}, respectively.~%
We can observe that the predictive performance of \ourmodel~is relatively stable under different settings.
In particular, the longer the target series to predict, the worse performance might be obtained.~%
Besides, when the amount of available data is shrunk, \ourmodel~ performs more stable than expected.~%
Note that on the 20\%-Electricity dataset, the performance of \ourmodel ~is much worse than other subsets of the Electricity dataset, mainly because the sliced  20\%-Electricity dataset involves more irregular values.

\begin{figure}[htbp]
    \centering
    \begin{minipage}[b]{0.05\textwidth}
     \centering
      \raisebox{0.3\height}{\rotatebox{90}{\small{\# factors = 2}}}
    \end{minipage}
    \begin{minipage}[t]{0.9\textwidth}
    \begin{subfigure}[t]{0.3\textwidth}
      \includegraphics[width=\textwidth]{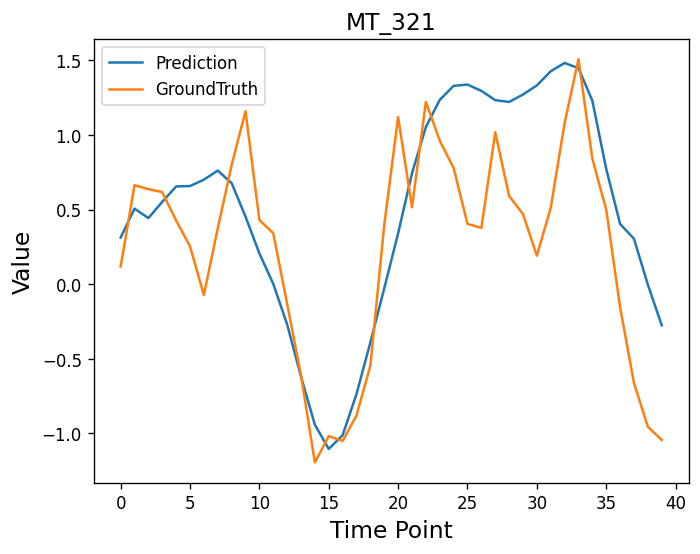}
    \end{subfigure}
    \begin{subfigure}[t]{0.34\textwidth}
      \includegraphics[width=\textwidth]{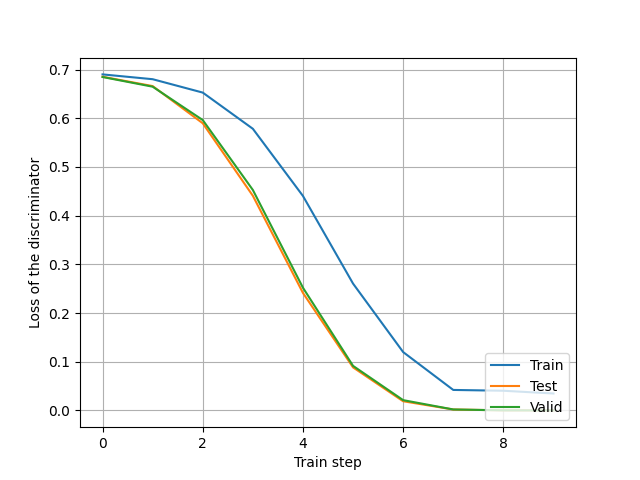}
    \end{subfigure}
    \begin{subfigure}[t]{0.34\textwidth}
      \includegraphics[width=\textwidth]{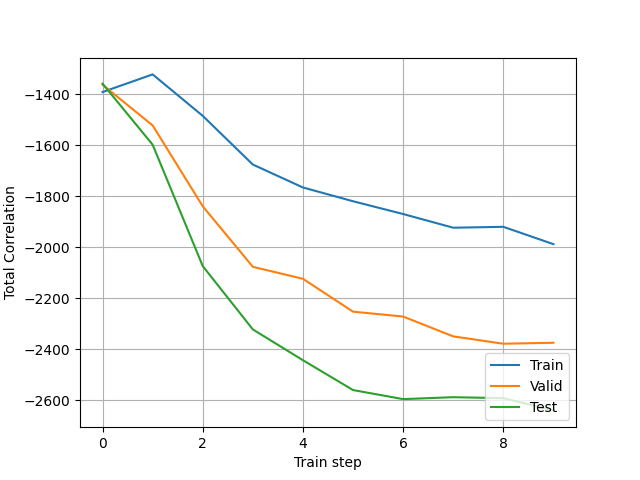}
    \end{subfigure}
    \end{minipage}
    
    \begin{minipage}[b]{0.05\textwidth}
     \centering
       \raisebox{0.3\height}{\rotatebox{90}{\small{\# factors = 4}}}
    \end{minipage}
    \begin{minipage}[t]{0.9\textwidth}
    \begin{subfigure}[t]{0.3\textwidth}
      \includegraphics[width=\textwidth]{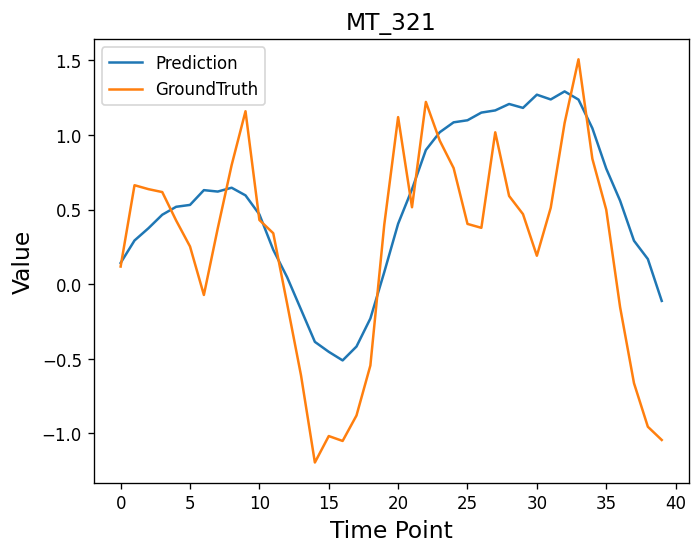}
    \end{subfigure}
    \begin{subfigure}[t]{0.34\textwidth}
      \includegraphics[width=\textwidth]{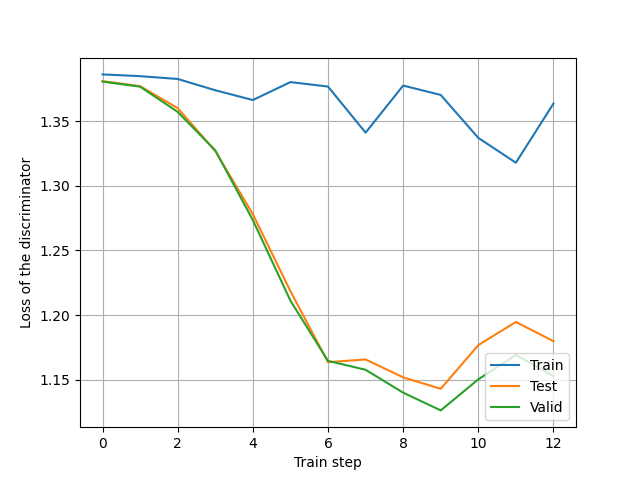}
    \end{subfigure}
    \begin{subfigure}[t]{0.34\textwidth}
      \includegraphics[width=\textwidth]{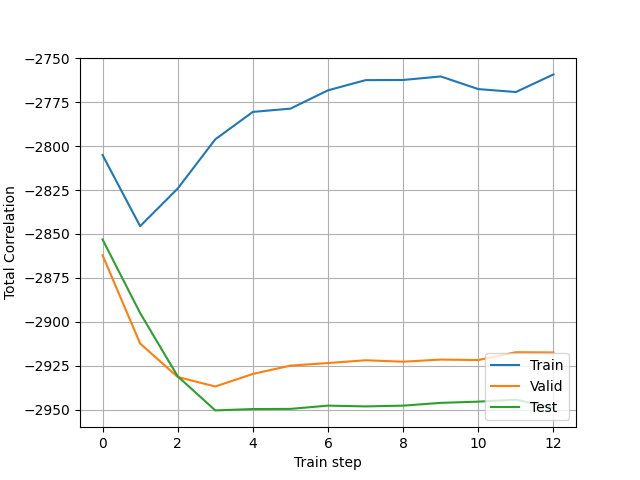}
    \end{subfigure}
    \end{minipage}
    
    \begin{minipage}[b]{0.05\textwidth}
     \centering
       \raisebox{0.3\height}{\rotatebox{90}{\small{\# factors = 6}}}
    \end{minipage}
    \begin{minipage}[t]{0.9\textwidth}
    \begin{subfigure}[t]{0.3\textwidth}
      \includegraphics[width=\textwidth]{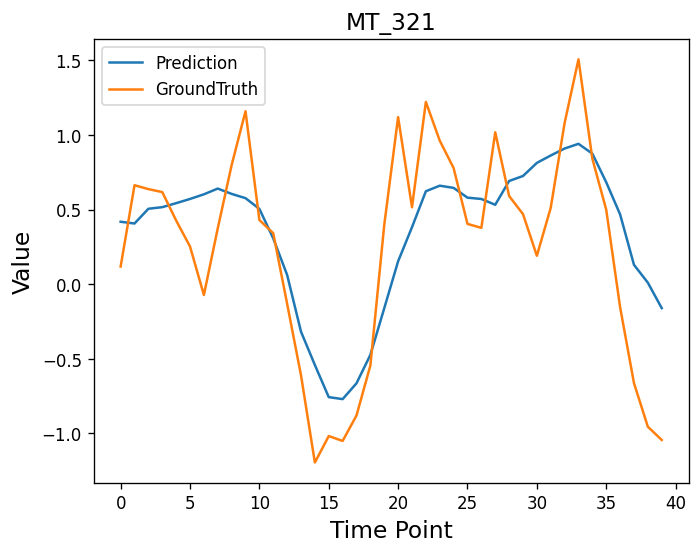}
    \end{subfigure}
    \begin{subfigure}[t]{0.34\textwidth}
      \includegraphics[width=\textwidth]{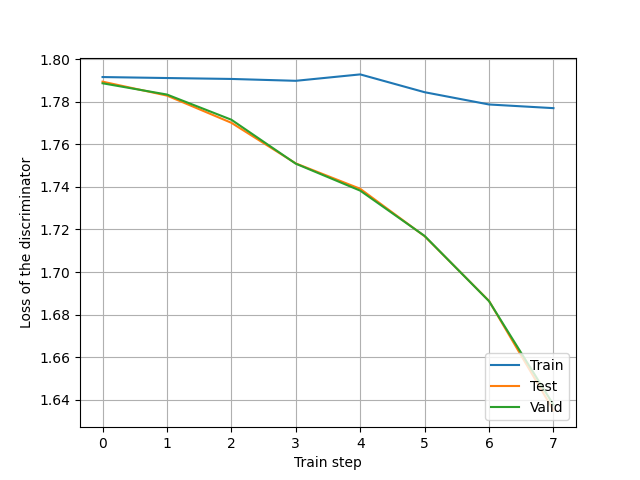}
    \end{subfigure}
    \begin{subfigure}[t]{0.34\textwidth}
      \includegraphics[width=\textwidth]{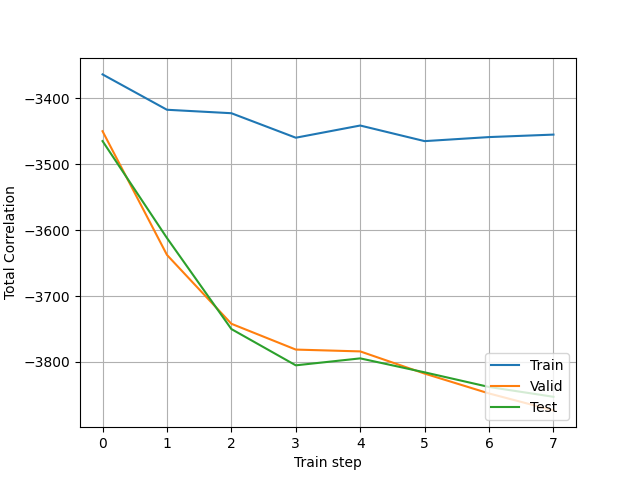}
    \end{subfigure}
    \end{minipage}
    
    \begin{minipage}[b]{0.05\textwidth}
     \centering
       \raisebox{0.3\height}{\rotatebox{90}{\small{\# factors = 8}}}
    \end{minipage}
    \begin{minipage}[t]{0.9\textwidth}
    \begin{subfigure}[t]{0.3\textwidth}
      \includegraphics[width=\textwidth]{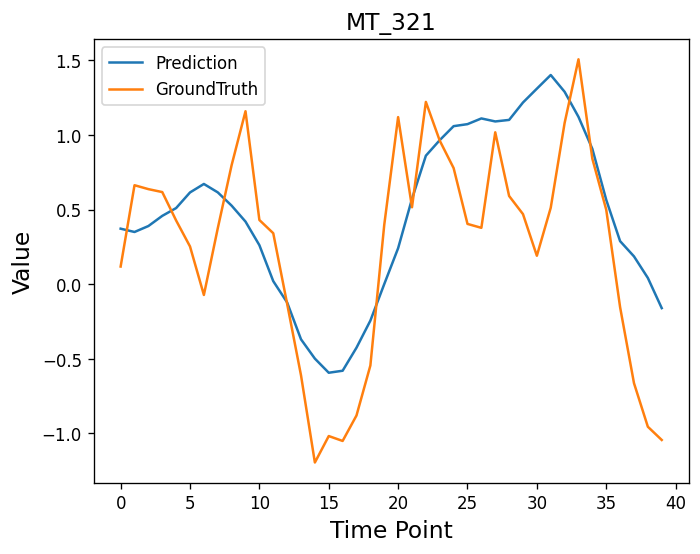}
    \end{subfigure}
    \begin{subfigure}[t]{0.34\textwidth}
      \includegraphics[width=\textwidth]{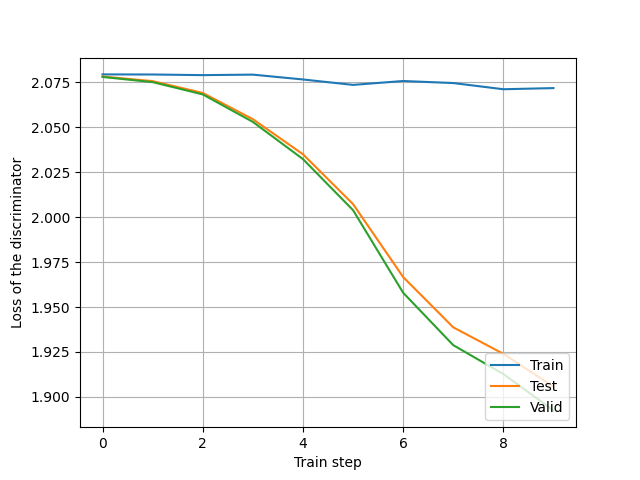}
    \end{subfigure}
    \begin{subfigure}[t]{0.34\textwidth}
      \includegraphics[width=\textwidth]{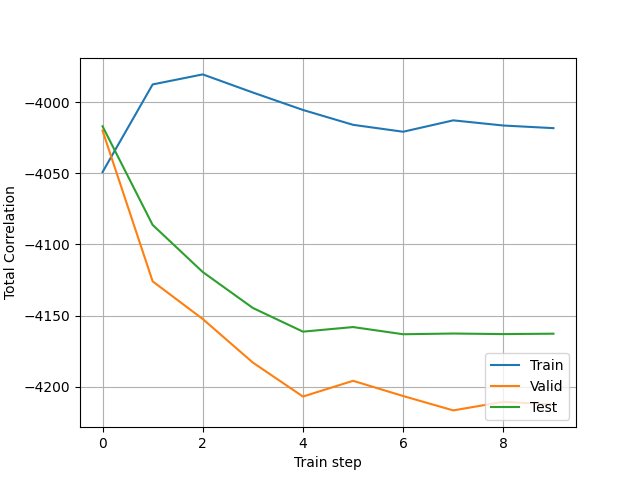}
    \end{subfigure}
    \end{minipage}
    
    \begin{minipage}[b]{0.05\textwidth}
     \centering
       \raisebox{0.3\height}{\rotatebox{90}{\small{\# factors = 10}}}
    \end{minipage}
    \begin{minipage}[t]{0.9\textwidth}
    \begin{subfigure}[t]{0.3\textwidth}
      \includegraphics[width=\textwidth]{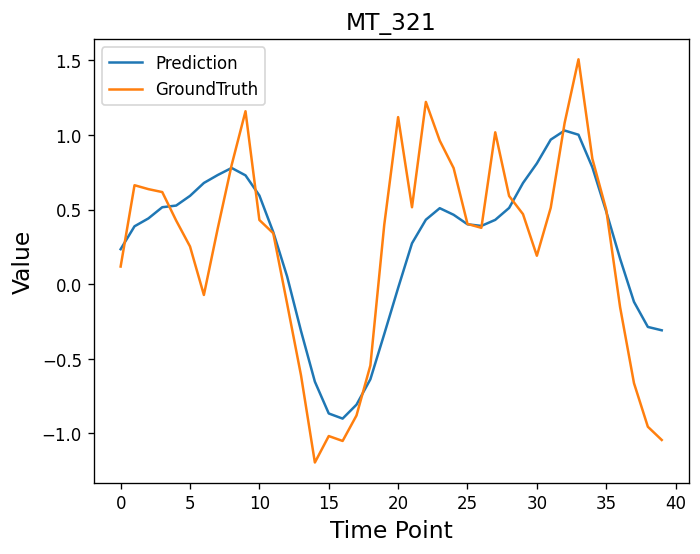}
    \end{subfigure}
    \begin{subfigure}[t]{0.34\textwidth}
      \includegraphics[width=\textwidth]{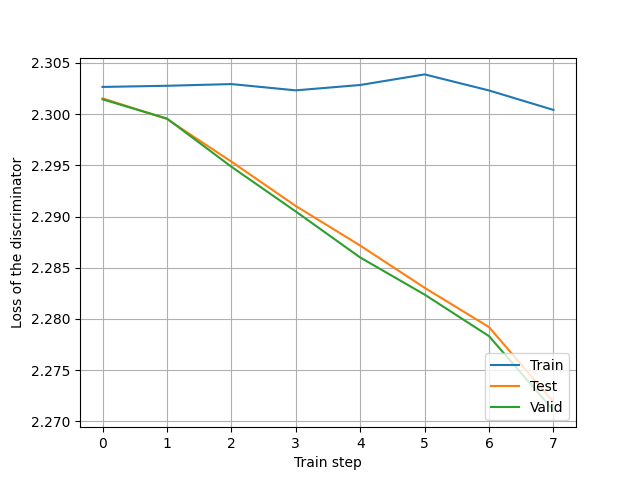}
    \end{subfigure}
    \begin{subfigure}[t]{0.34\textwidth}
      \includegraphics[width=\textwidth]{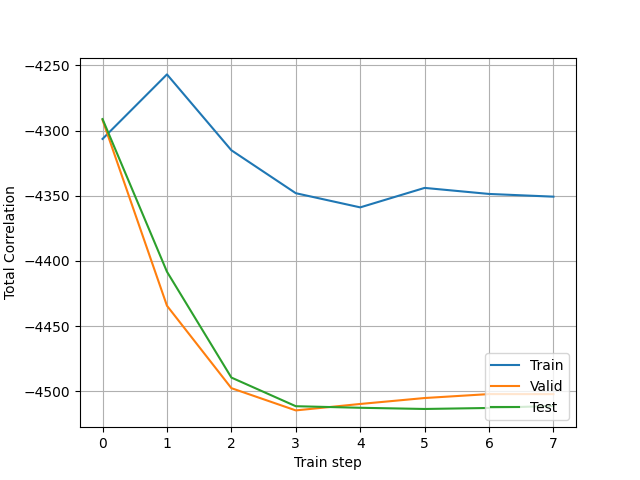}
    \end{subfigure}
    \end{minipage}
    
    \begin{minipage}[b]{0.05\textwidth}
     \centering
       \raisebox{0.3\height}{\rotatebox{90}{\small{\# factors = 12}}}
    \end{minipage}
    \begin{minipage}[t]{0.9\textwidth}
    \begin{subfigure}[t]{0.3\textwidth}
      \includegraphics[width=\textwidth]{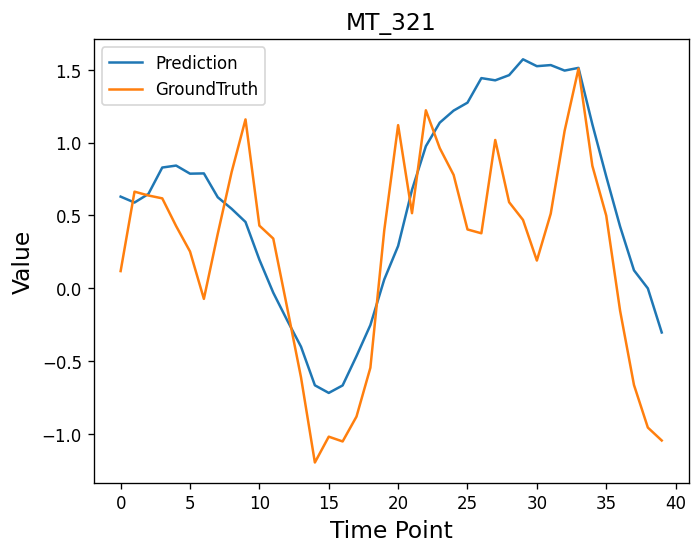}
    \end{subfigure}
    \begin{subfigure}[t]{0.34\textwidth}
      \includegraphics[width=\textwidth]{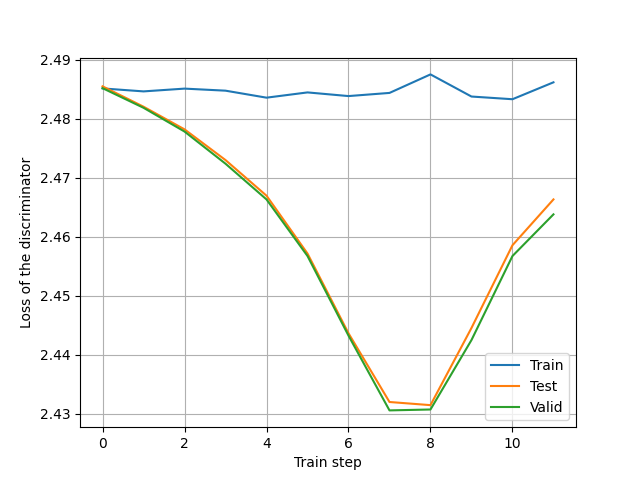}
    \end{subfigure}
    \begin{subfigure}[t]{0.34\textwidth}
      \includegraphics[width=\textwidth]{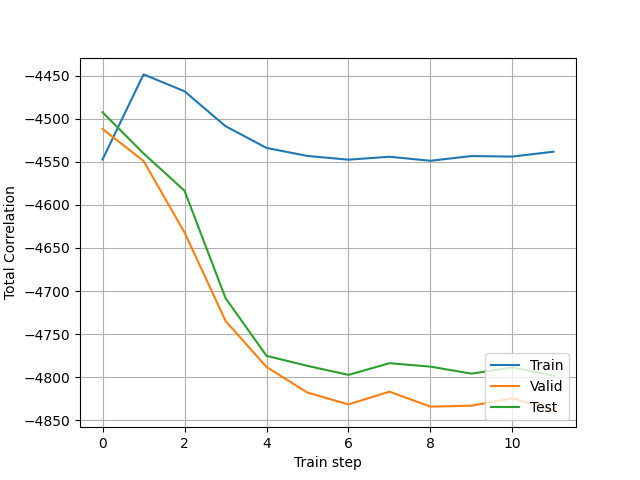}
    \end{subfigure}
    \end{minipage}
    \caption{
    We showcase an instance from the Electricity dataset and demonstrate the results when different numbers of factors in disentanglement are adopted.
    For each row, from left to right, the prediction result of TSF, the learning curve of the discriminator, and the total correlation are plotted, respectively. 
    }   \label{fig:tc}
\end{figure}

\section{Disentanglement for Time Series Forecasting} \label{disentangle}

\begin{figure}[t]
    \centering
    \includegraphics[width=\textwidth]{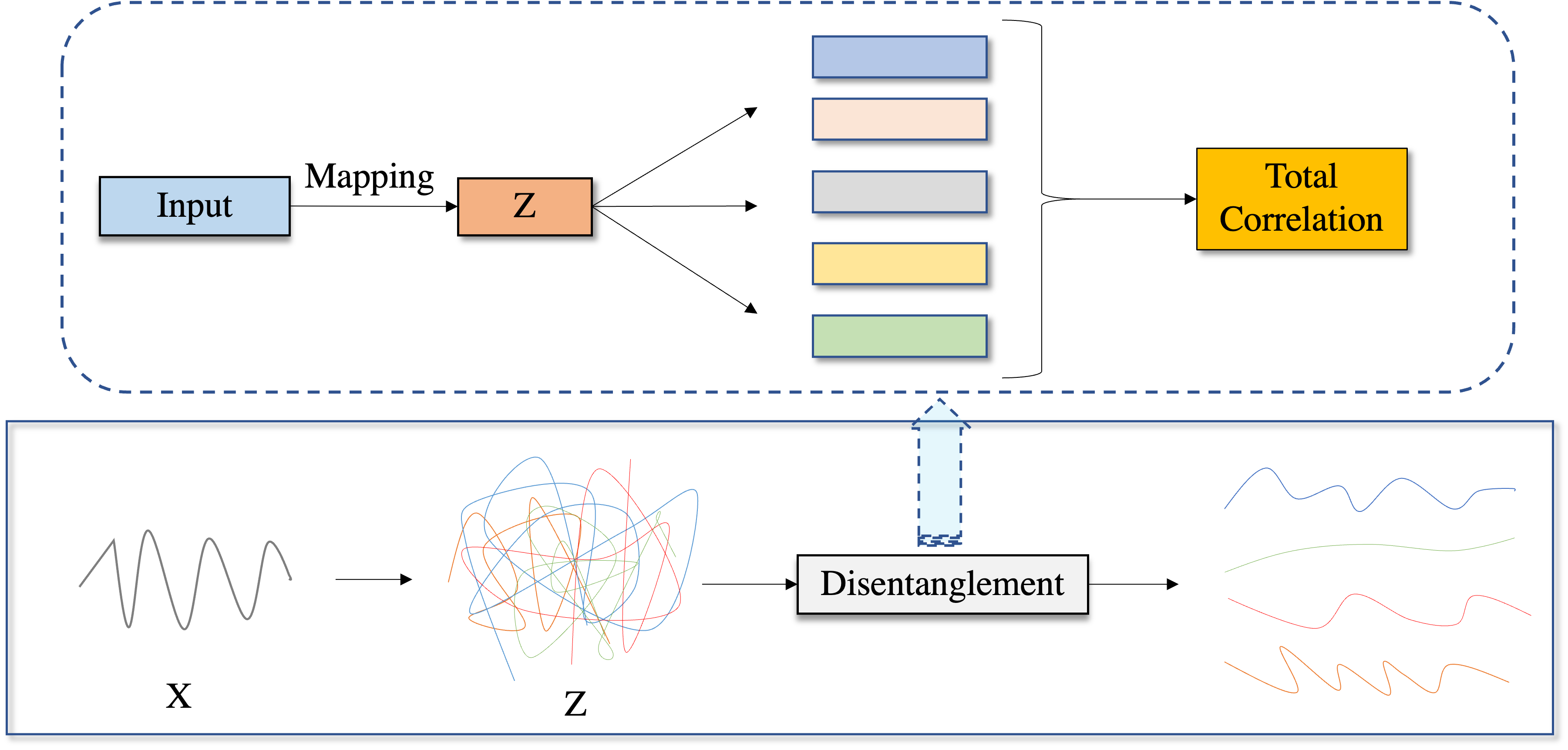}
    \caption{
    Disentangling latent variable $Z$ of time series.
    Specifically, the input $X$ is first mapped into $Z$.
    Then $\forall z_i \in Z$ is decomposed as $z_i = [z_{i,1}, \cdots, z_{i, m}]$ and the metric of total correlation is utilized to minimize the inter-dependencies among ``hand-crafted'' factors. 
    In this way, the disentangled factors tend to be not only discriminative but also informative. 
    }
    \label{fig:disen}
\end{figure}

\cref{fig:disen} illustrates the disentanglement of latent variable $Z$ for time series forecasting. 
It is difficult to choose suitable disentanglement factors under the unsupervised learning of disentanglement. 
Therefore, we attempt to inspect the TSF performance against different numbers of factors to be disentangled. 
We implement a simple classifier as a discriminator to further evaluate the disentanglement quality in \cref{fig:tc} (and \cref{alg3} demonstrates the training procedure of the discriminator). 
To be specific, 
we take different dimensions of $Z$ as the factors to be disentangled: $ z_i = [ z_{i, 1}, \cdots, z_{i, m}] $ ($z_i \in Z $), 
then an instance consisting of factor and label $(z_{i,j}, j)$ is constructed. 
We shuffle these $m$ examples for each $z_i$ and attempt to classify them with a discriminator, then the disentanglement can be evaluated by measuring the loss of the discriminator.  
The learning curve of the discriminator can be leveraged to assess the disentanglement, and the discriminator is implemented by an MLP with six nonlinear layers and 100 hidden states. 
The results of prediction, discriminator loss, and the total correlation w.r.t. different numbers of factors are plotted in \cref{fig:tc}, respectively. 
As shown in~\cref{fig:tc}, the number of factors does affect the prediction performance, as well as the disentanglement quality. 
On the other hand, 
the learning curves can be converged when different factors are adopted, which validates that the disentanglement of the latent factors is of high quality. 

%
\begin{algorithm}[t]
\caption{Train a discriminator for time series disentanglement.}  \label{alg3}
\begin{algorithmic}[1]
\small{
\REPEAT
  \STATE 
  Initialize the loss of a discriminator $ \mathcal{D}_{\varphi} $: ${L}(\mathcal{D}_{\varphi}) = 0$ 
  \STATE 
  Decompose the latent variable generated in \cref{alg_train} 
  as $z_i = [z_{i,1}, z_{i,2}, ..., z_{i,m}]$ ($i = 1, \cdots, n$) 
  \FOR{$z_i$ in Z}
    \STATE ${L} = {L} + \sum_{j=1}^{m} \| \mathcal{D}_{\varphi}(z_{i,j})- j \|^2$
  \ENDFOR
  \STATE 
  Optimize the discriminator: $\varphi \leftarrow \mathrm{argmin}({L})$
\UNTIL Convergence
}
\end{algorithmic}
\end{algorithm}
%

\begin{figure}[t]
    \centering
    \centering
    \begin{subfigure}[t]{0.24\textwidth}
      \includegraphics[width=\textwidth]{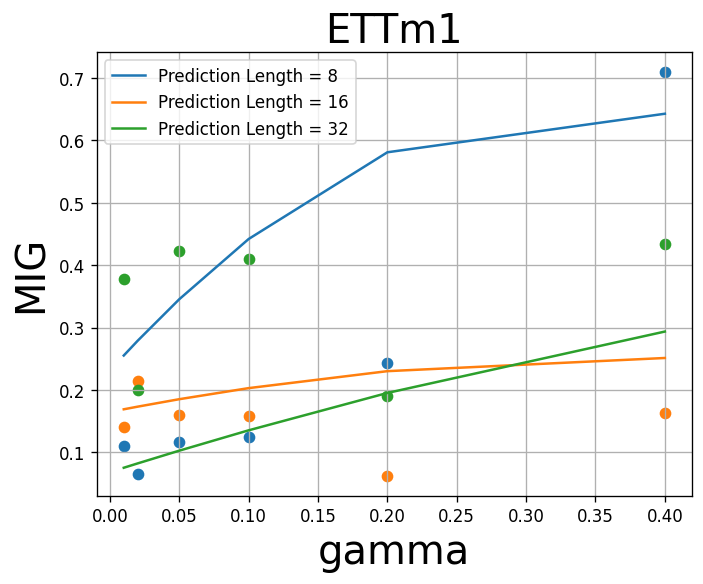}
    \caption{~} \label{fig:mig:a}
    \end{subfigure}
    \begin{subfigure}[t]{0.24\textwidth}
      \includegraphics[width=\textwidth]{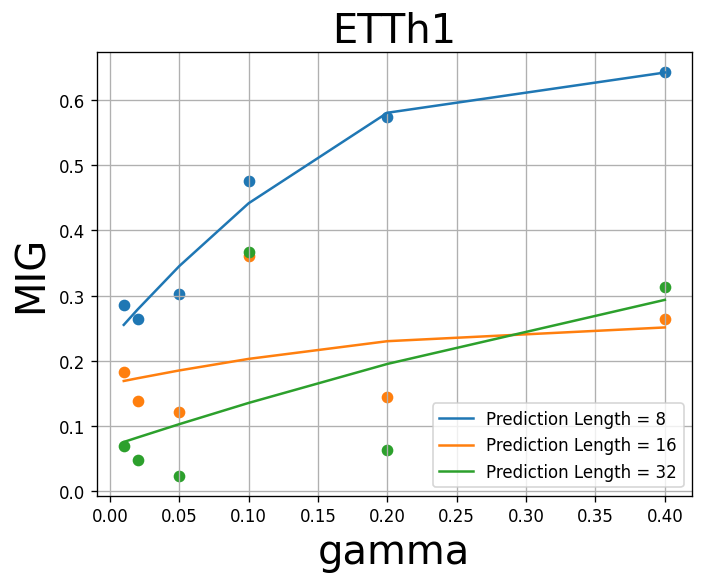}
    \caption{~} \label{fig:mig:b}
    \end{subfigure}
    \centering
    \begin{subfigure}[t]{0.24\textwidth}
      \includegraphics[width=\textwidth]{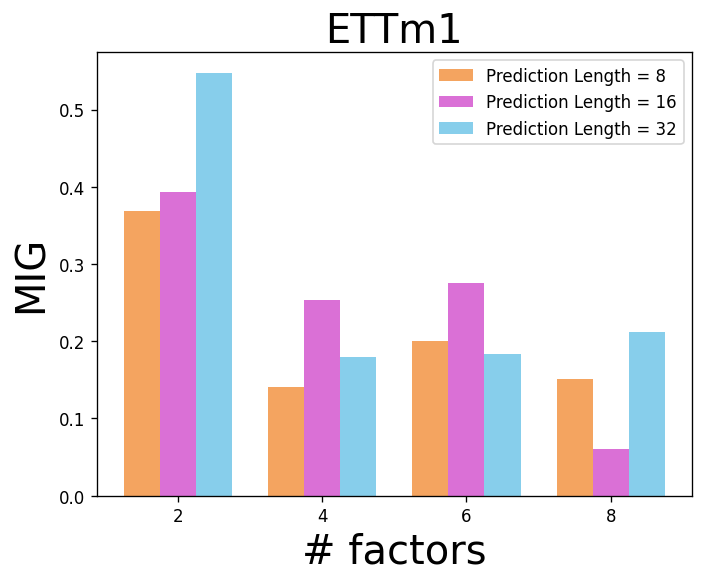}
    \caption{~} \label{fig:mig:c}
    \end{subfigure}
    \begin{subfigure}[t]{0.24\textwidth}
      \includegraphics[width=\textwidth]{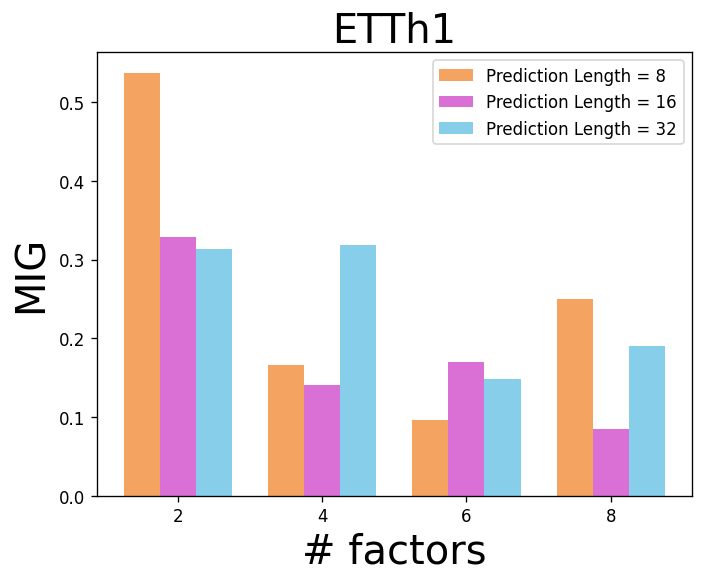}
    \caption{~} \label{fig:mig:d}
    \end{subfigure}
    \caption{
    We evaluate the quality of disentanglement on ETTm1 and ETTh1 datasets regarding the mutual information gap (MIG). 
    {{\bf (a-b)}} 
    The scatter plots of the MIG against varying weights $\gamma$ in loss function (refer to Eq. (14) 
    in the main text).
    {{\bf (c-d)}}
    MIG v.s. different numbers of factors.
    }
    \label{fig:mig}
\end{figure}

In addition to the above method evaluating the disentanglement indirectly, we adopt another metric named Mutual Information Gap (MIG)~\cite{chen2018isolating} to evaluate the quality of disentanglement in a more straightforward way. 
Specifically, 
for a latent variable $z_i \in Z$, the mutual information between $z_{i, j}$, and a factor $v_k \in [1, m]$ can be calculated by
\begin{equation}
    I_d(z_{i, j}, v_k) = 
    \mathbb{E}_{q(z_{i,j}, v_k)} 
    [log \sum_{d \in \mathcal{S}_{v_k}} q(z_{i,j}|d) p(d|v_k)] 
    + 
    H(z_{i, j}) \, ,
\end{equation}
%
where $d$ denotes the sample of $z_{i,j}$ and 
$ \mathcal{S}_{v_k} $ is the support set of $v_k$. 
Then, for $z_{i,j}$ 
\begin{equation}
    MIG(z_{i, j}) = \frac{1}{m}\sum_{1}^{m}
    \frac{1}{H(v_k)}(max(I_d(z_{i, j}, v_k)) - submax(I_d(z_{i, j}, v_k))) \, ,
\end{equation}
%
where $submax$ means the second max value of $I_d(z_{i, j}, v_k)$, then the MIG of $Z$ can be obtained as 
\begin{equation}
  MIG(Z) = \sum_{i=1}^{n}MIG(z_i), \quad MIG(z_i) = \frac{1}{m}\sum_{j=1}^{m}MIG(z_{i,j}) \, .
\end{equation}
We evaluate the quality of disentanglement in terms of MIG on ETTm1 and ETTh1 datasets, respectively, which can be seen in~\cref{fig:mig}.
From \cref{fig:mig:a,fig:mig:b}, when the weight of disentanglement (i.e., $\gamma$ in \cref{loss} 
of the main text) grows, the disentangled factors are of higher quality.
In other words, the latent variables can be disentangled with the help of the disentanglement module in \ourmodel. 
In addition, we examine the changes in MIG against different numbers of factors. 
We can observe that the difficulty of disentanglement climbs up as the number of factors increases.

\begin{figure}[htbp]
    \centering
    \begin{minipage}[c]{0.02\textwidth}
    \raisebox{0.1\height}{\rotatebox{90}{T = 10}}
    \end{minipage}
    \begin{minipage}[c]{0.96\textwidth}
    \begin{subfigure}[t]{0.24\textwidth}
      \caption{$\beta_t \in [0, 0.1]$}
      \includegraphics[width=\textwidth]{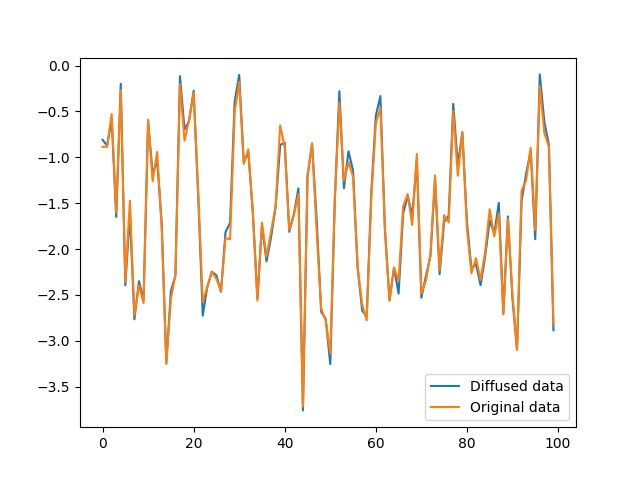}
    \end{subfigure}
     \begin{subfigure}[t]{0.24\textwidth}
     \caption{$\beta_t \in [0, 0.2]$}
      \includegraphics[width=\textwidth]{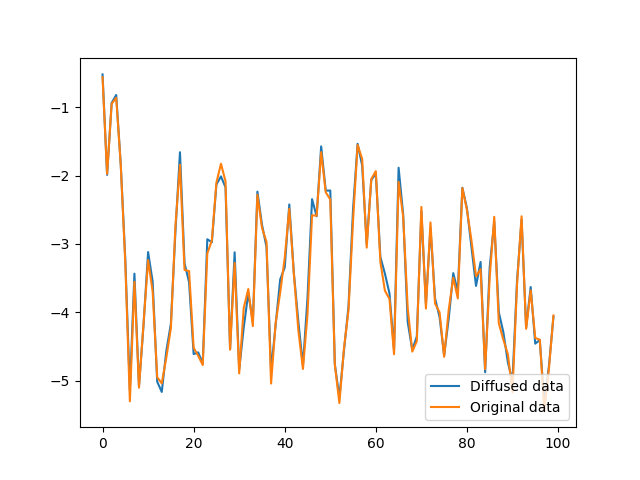}
    \end{subfigure}
    \begin{subfigure}[t]{0.24\textwidth}
    \caption{$\beta_t \in [0, 0.3]$}
      \includegraphics[width=\textwidth]{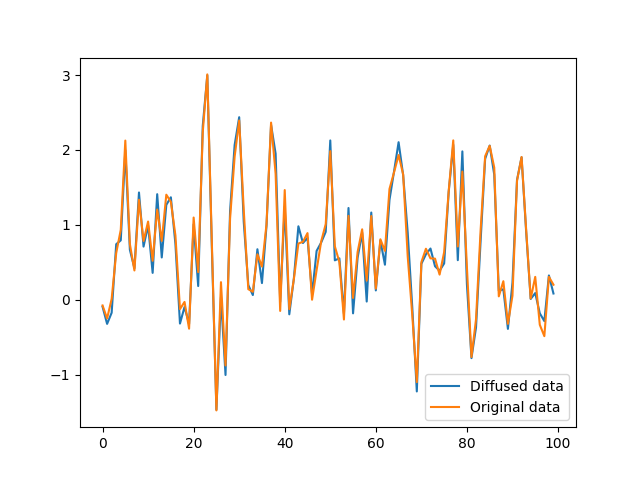}
    \end{subfigure}
    \begin{subfigure}[t]{0.24\textwidth}
    \caption{$\beta_t \in [0, 0.4]$}
      \includegraphics[width=\textwidth]{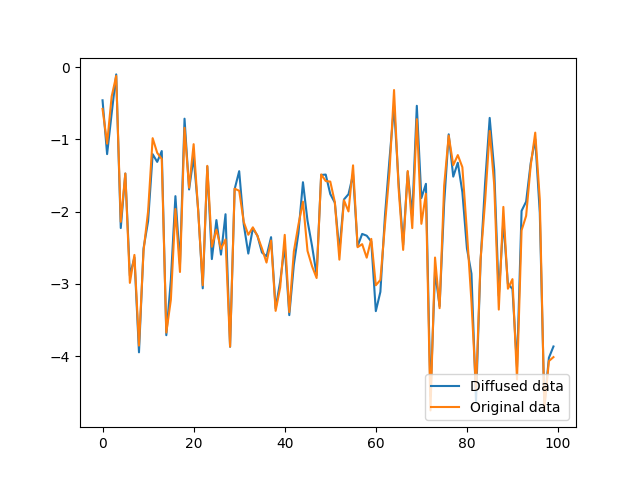}
    \end{subfigure}
    \end{minipage}
    
    \begin{minipage}[c]{0.02\textwidth}
    \raisebox{0.1\height}{\rotatebox{90}{T = 50}}
    \end{minipage}
    \begin{minipage}[c]{0.96\textwidth}
    \begin{subfigure}[t]{0.24\textwidth}
      \includegraphics[width=\textwidth]{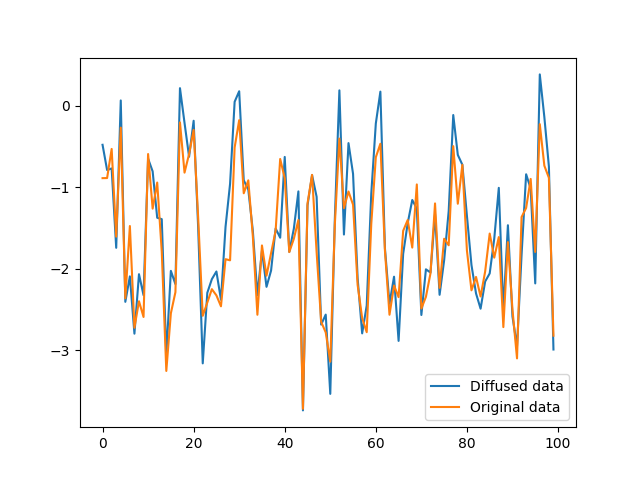}
    \end{subfigure}
     \begin{subfigure}[t]{0.24\textwidth}
      \includegraphics[width=\textwidth]{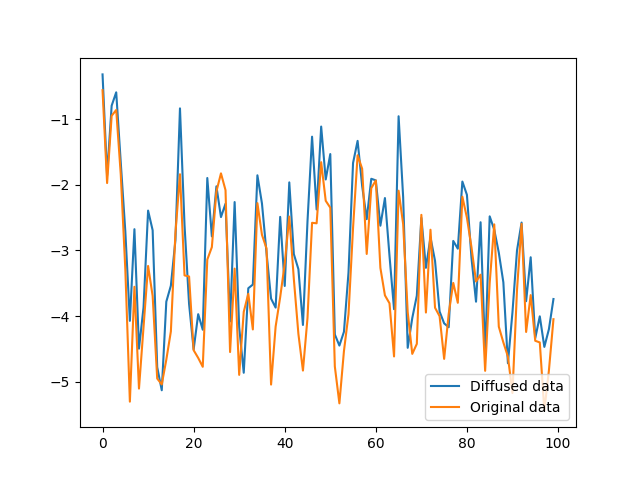}
    \end{subfigure}
    \begin{subfigure}[t]{0.24\textwidth}
      \includegraphics[width=\textwidth]{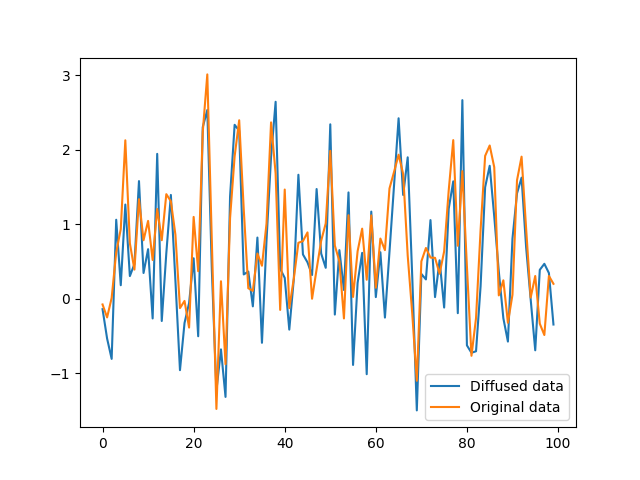}
    \end{subfigure}
    \begin{subfigure}[t]{0.24\textwidth}
      \includegraphics[width=\textwidth]{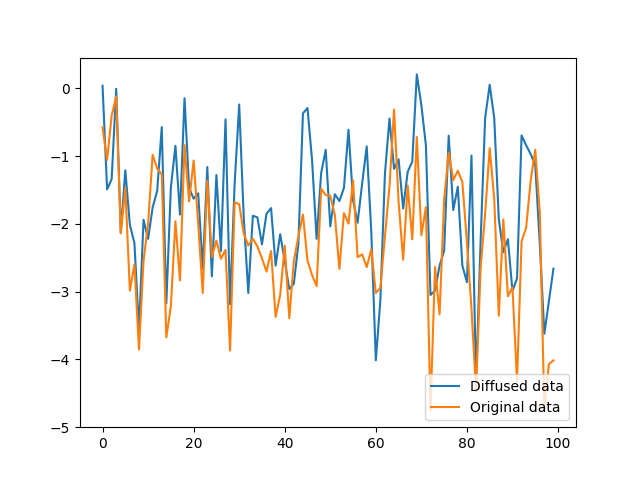}
    \end{subfigure}
    \end{minipage}
    
     \begin{minipage}[c]{0.02\textwidth}
      \raisebox{0.1\height}{\rotatebox{90}{T = 100}}
    \end{minipage}
    \begin{minipage}[c]{0.96\textwidth}
    \begin{subfigure}[t]{0.24\textwidth}
      \includegraphics[width=\textwidth]{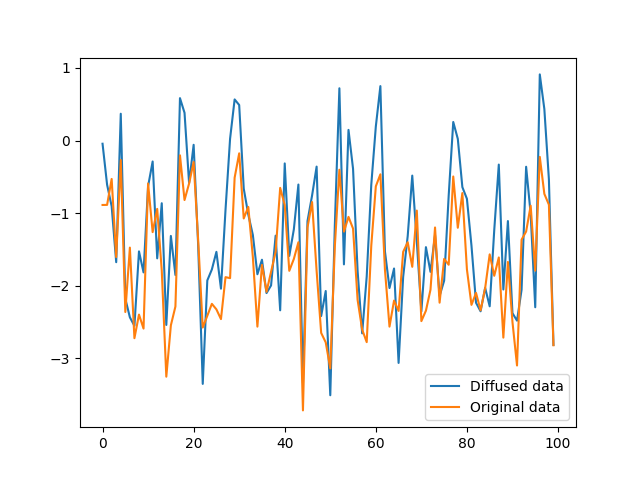}
    \end{subfigure}
     \begin{subfigure}[t]{0.24\textwidth}
      \includegraphics[width=\textwidth]{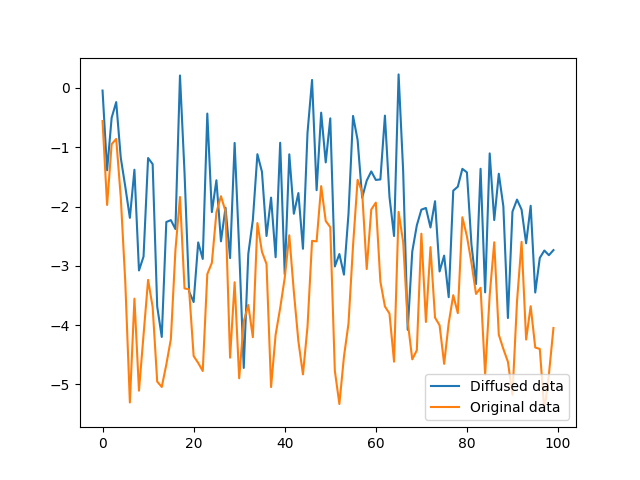}
    \end{subfigure}
    \begin{subfigure}[t]{0.24\textwidth}
      \includegraphics[width=\textwidth]{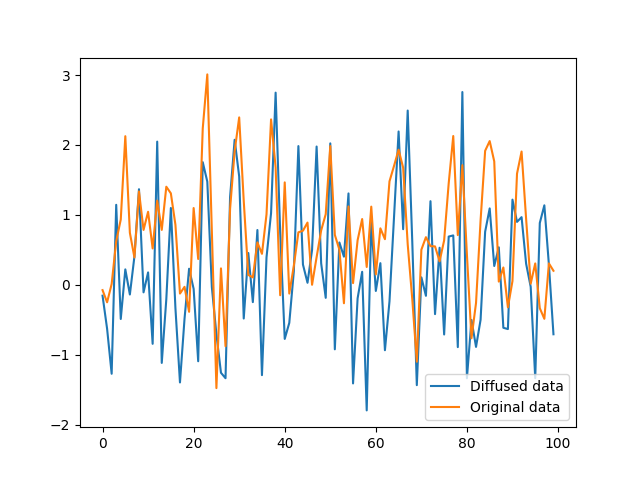}
    \end{subfigure}
    \begin{subfigure}[t]{0.24\textwidth}
      \includegraphics[width=\textwidth]{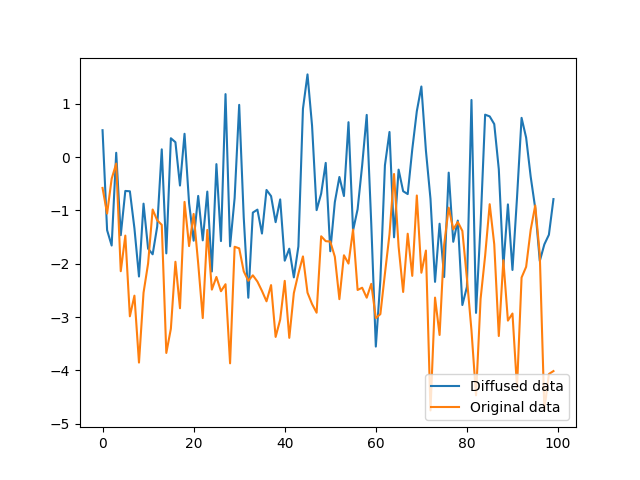}
    \end{subfigure}
    \end{minipage}
    
     \begin{minipage}[c]{0.02\textwidth}
    \raisebox{0.1\height}{\rotatebox{90}{T = 200}}
    \end{minipage}
    \begin{minipage}[c]{0.96\textwidth}
    \begin{subfigure}[t]{0.24\textwidth}
      \includegraphics[width=\textwidth]{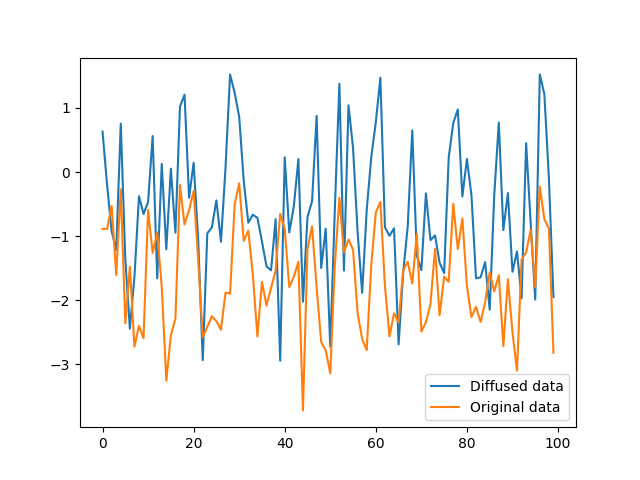}
    \end{subfigure}
     \begin{subfigure}[t]{0.24\textwidth}
      \includegraphics[width=\textwidth]{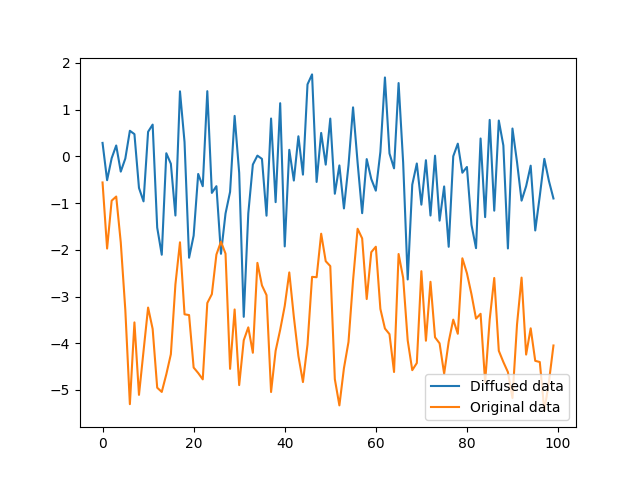}
    \end{subfigure}
    \begin{subfigure}[t]{0.24\textwidth}
      \includegraphics[width=\textwidth]{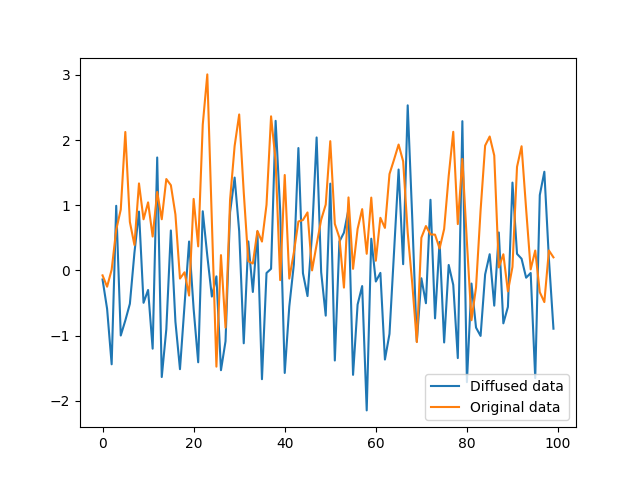}
    \end{subfigure}
    \begin{subfigure}[t]{0.24\textwidth}
      \includegraphics[width=\textwidth]{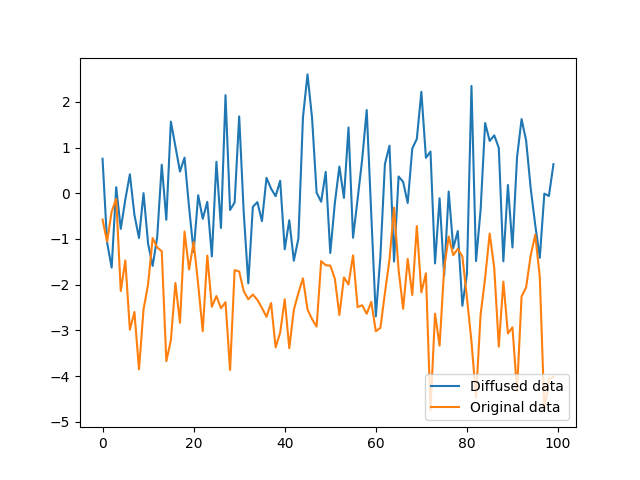}
    \end{subfigure}
    \end{minipage} 

\begin{minipage}[c]{0.02\textwidth}
    \raisebox{0.1\height}{\rotatebox{90}{T = 400}}
\end{minipage}
    \begin{minipage}[c]{0.96\textwidth}
    \begin{subfigure}[t]{0.24\textwidth}
      \includegraphics[width=\textwidth]{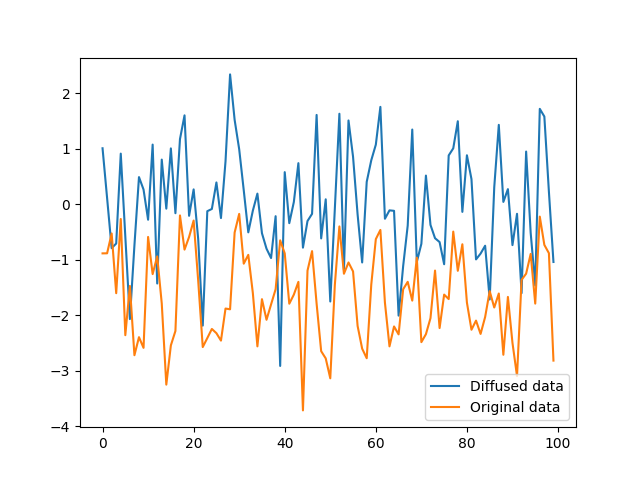}
    \end{subfigure}
     \begin{subfigure}[t]{0.24\textwidth}
      \includegraphics[width=\textwidth]{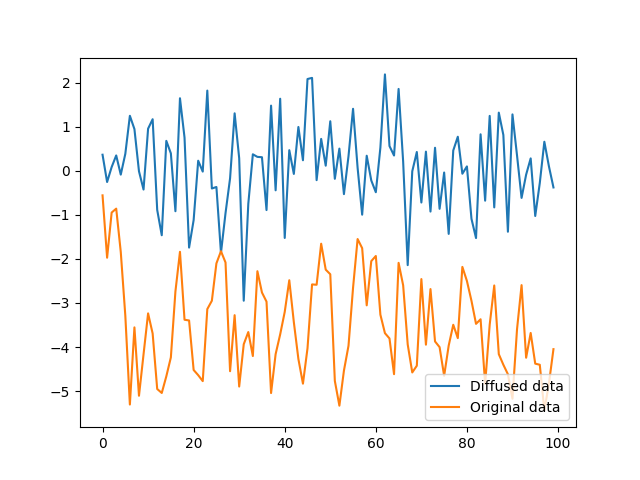}
    \end{subfigure}
    \begin{subfigure}[t]{0.24\textwidth}
      \includegraphics[width=\textwidth]{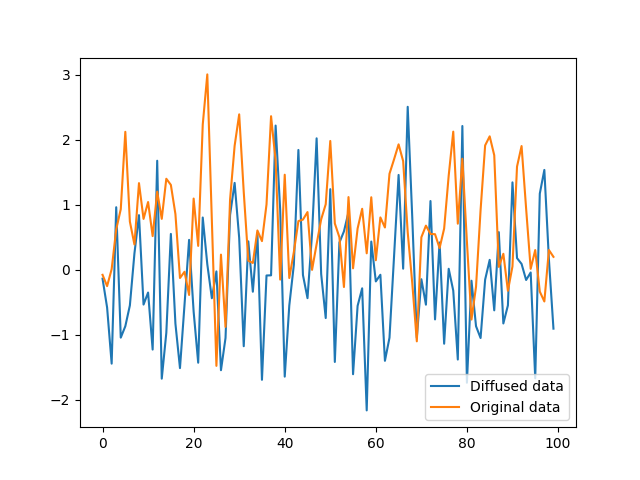}
    \end{subfigure}
    \begin{subfigure}[t]{0.24\textwidth}
      \includegraphics[width=\textwidth]{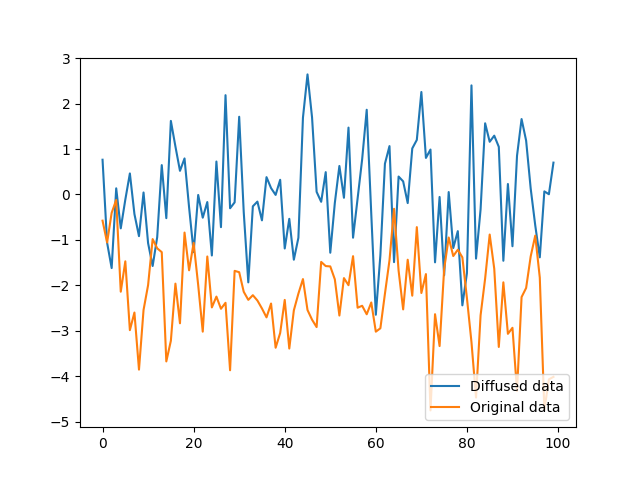}
    \end{subfigure}
    \end{minipage}
    
    \begin{minipage}[c]{0.02\textwidth}
    \raisebox{0.1\height}{\rotatebox{90}{T = 600}}
\end{minipage}
    \begin{minipage}[c]{0.96\textwidth}
    \begin{subfigure}[t]{0.24\textwidth}
      \includegraphics[width=\textwidth]{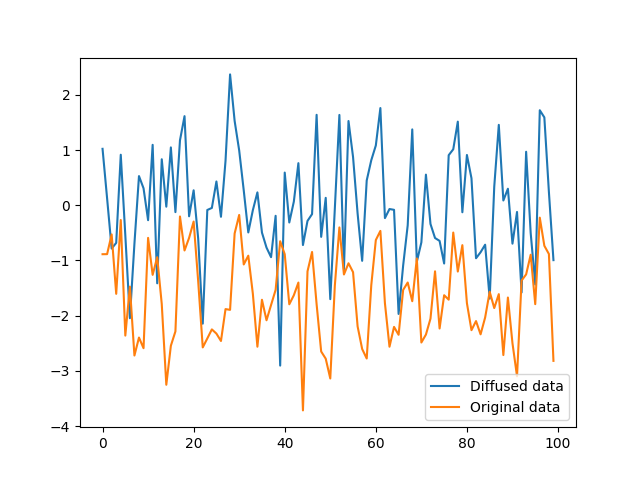}
    \end{subfigure}
     \begin{subfigure}[t]{0.24\textwidth}
      \includegraphics[width=\textwidth]{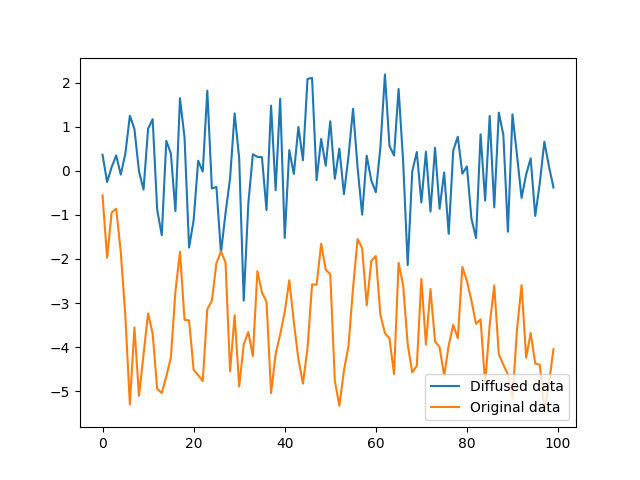}
    \end{subfigure}
    \begin{subfigure}[t]{0.24\textwidth}
      \includegraphics[width=\textwidth]{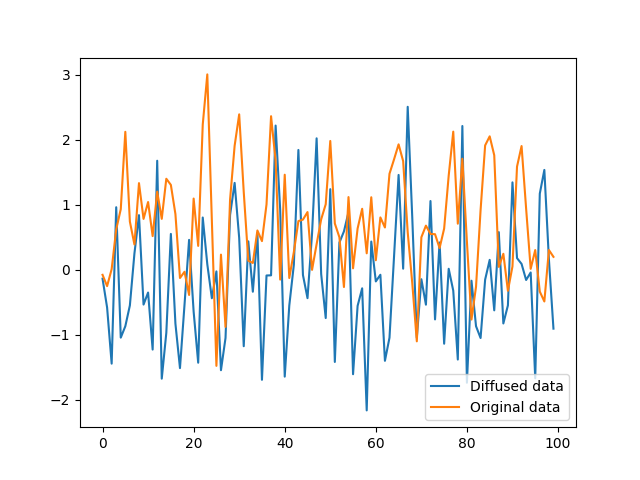}
    \end{subfigure}
    \begin{subfigure}[t]{0.24\textwidth}
      \includegraphics[width=\textwidth]{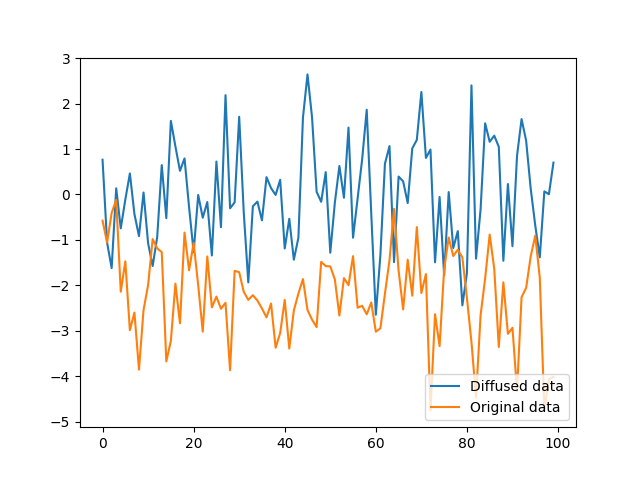}
    \end{subfigure}
    \end{minipage}
    
    \begin{minipage}[c]{0.02\textwidth}
    \raisebox{0.1\height}{\rotatebox{90}{T = 800}}
\end{minipage}
    \begin{minipage}[c]{0.96\textwidth}
    \begin{subfigure}[t]{0.24\textwidth}
      \includegraphics[width=\textwidth]{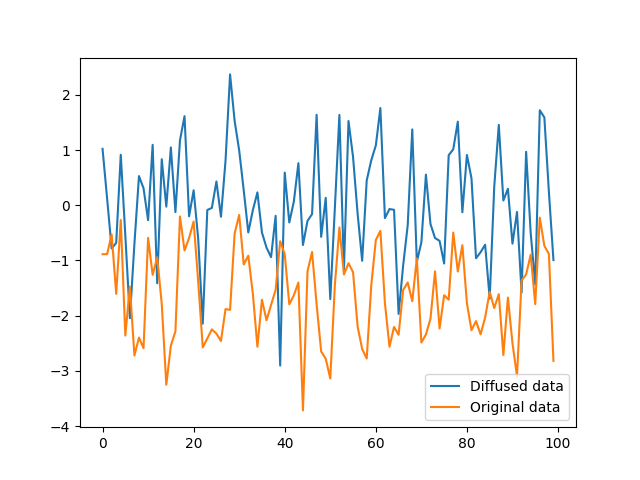}
    \end{subfigure}
     \begin{subfigure}[t]{0.24\textwidth}
      \includegraphics[width=\textwidth]{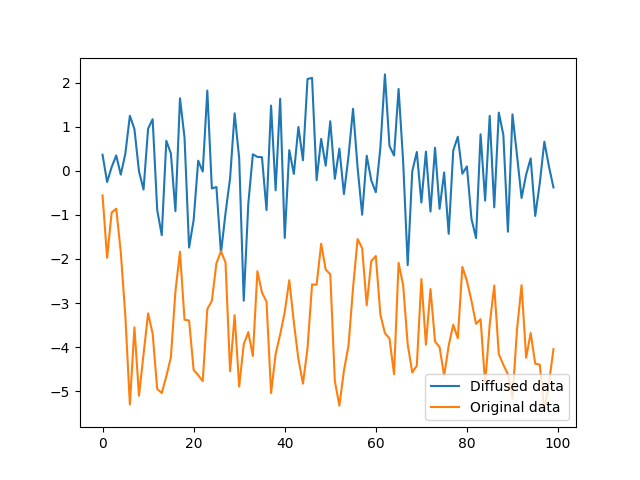}
    \end{subfigure}
    \begin{subfigure}[t]{0.24\textwidth}
      \includegraphics[width=\textwidth]{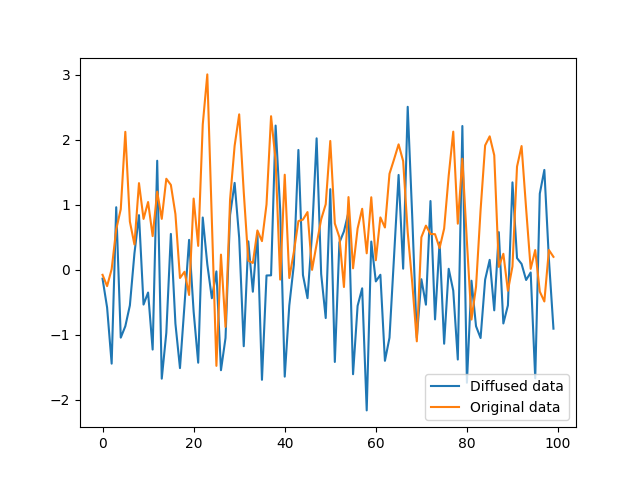}
    \end{subfigure}
    \begin{subfigure}[t]{0.24\textwidth}
      \includegraphics[width=\textwidth]{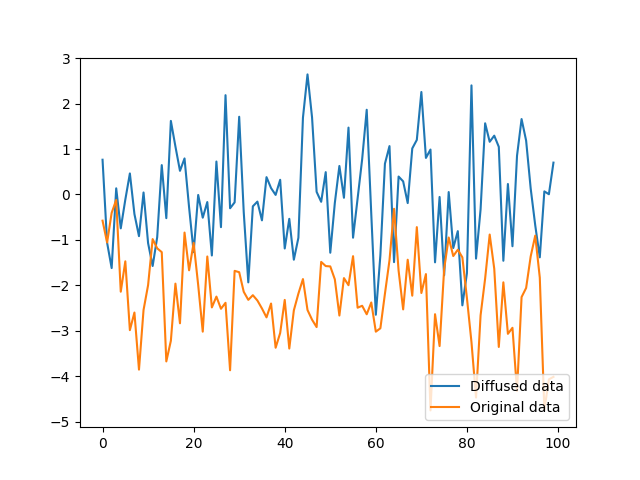}
    \end{subfigure}
    \end{minipage}
    \begin{minipage}[c]{0.02\textwidth}
    \raisebox{0.1\height}{\rotatebox{90}{T = 900}}
    \end{minipage}
    \begin{minipage}[c]{0.96\textwidth}
    \begin{subfigure}[t]{0.24\textwidth}
      \includegraphics[width=\textwidth]{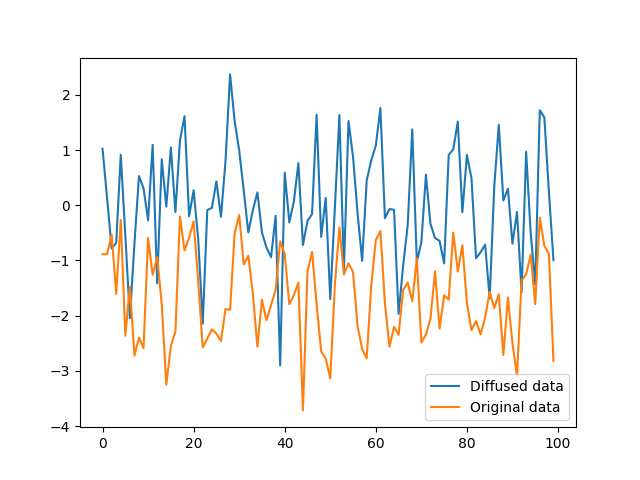}
    \end{subfigure}
     \begin{subfigure}[t]{0.24\textwidth}
      \includegraphics[width=\textwidth]{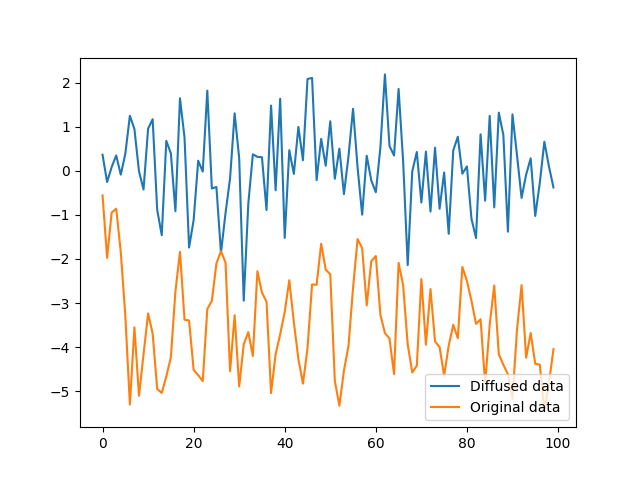}
    \end{subfigure}
    \begin{subfigure}[t]{0.24\textwidth}
      \includegraphics[width=\textwidth]{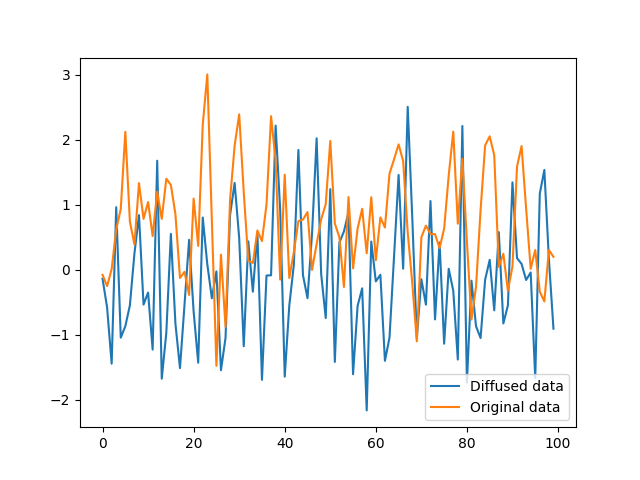}
    \end{subfigure}
    \begin{subfigure}[t]{0.24\textwidth}
      \includegraphics[width=\textwidth]{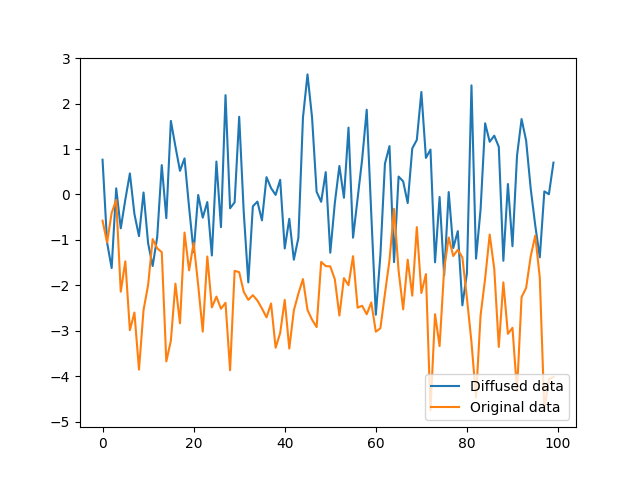}
    \end{subfigure}
    \end{minipage}
    \caption{
    Diffused time series with different variance schedules and diffusion steps.
    We randomly choose a sample series from the synthetic dataset D2 and plot the original time series data, as well as the diffused series.
    }
    \label{fig:diff_e}
\end{figure}

\section{Model Inspection: Coupled Diffusion Process} \label{B1}

To gain more insights into the coupled diffusion process, we demonstrate how a time series can be diffused under different settings in terms of variance schedule $\bm{\beta}$ and the max number of diffusion steps $T$. 
The examples are illustrated in~\cref{fig:diff_e}. 
It can be seen that when larger diffusion steps or a wider variance schedule is employed, the diffused series deviates far from the original data gradually, which may result in the loss of useful signals, like, temporal dependencies.
Therefore, it is important to choose a suitable variance schedule and diffusion steps to ensure that the distribution space is deviated enough without losing useful signals.

\section{Necessity of Data Augmentation for Time Series Forecasting} \label{A1}

Limited data would result in overfitting and poor performance. 
To demonstrate the necessity of enlarging the size of data for time series forecasting when deep models are employed, we implement a two-layer RNN and evaluate how many time points are required to ensure the generalization ability. 
A synthetic dataset is adopted for this demonstration. 

\begin{figure}[t]
    \centering
    \begin{subfigure}[t]{0.32\textwidth}
        \includegraphics[width=\textwidth]{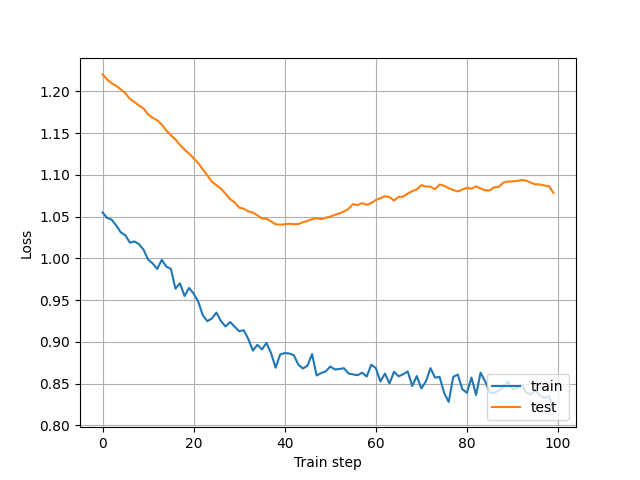}
        \caption{400 time points.}
    \end{subfigure}
   \begin{subfigure}[t]{0.32\textwidth}
    \includegraphics[width=\textwidth]{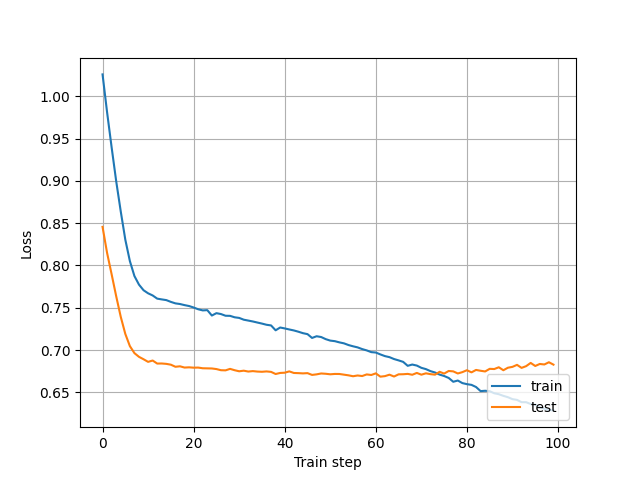}
    \caption{800 time points.}
    \end{subfigure}
   \begin{subfigure}[t]{0.32\textwidth}
    \includegraphics[width=\textwidth]{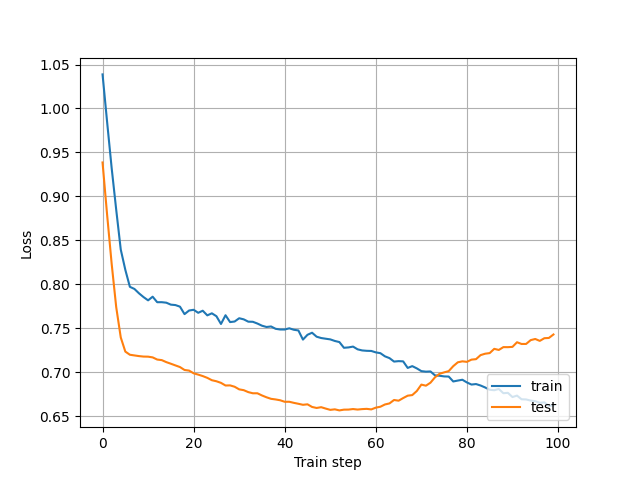}
    \caption{1000 time points.}
    \end{subfigure}

    \begin{subfigure}[t]{0.32\textwidth}
     \includegraphics[width=\textwidth]{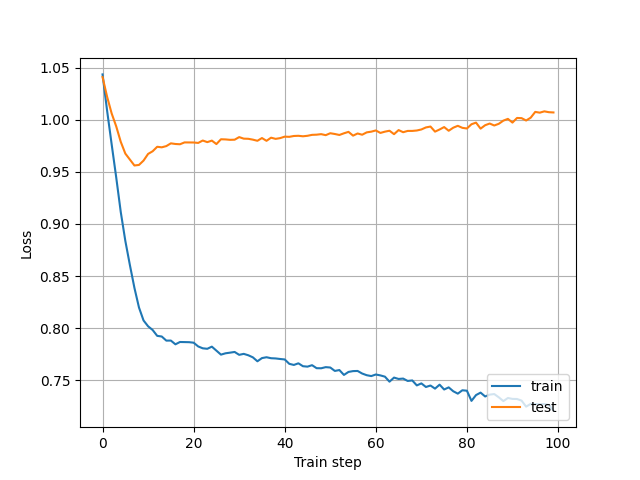}
     \caption{1200 time points.}
    \end{subfigure}
    \begin{subfigure}[t]{0.32\textwidth}
     \includegraphics[width=\textwidth]{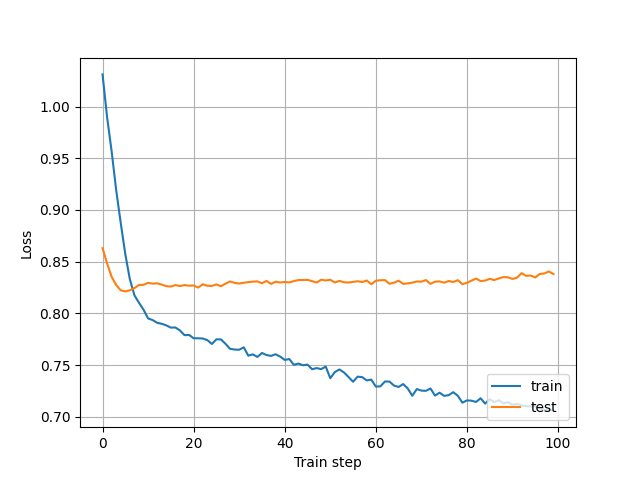}
     \caption{1400 time points.}
    \end{subfigure}
    \begin{subfigure}[t]{0.32\textwidth}
      \includegraphics[width=\textwidth]{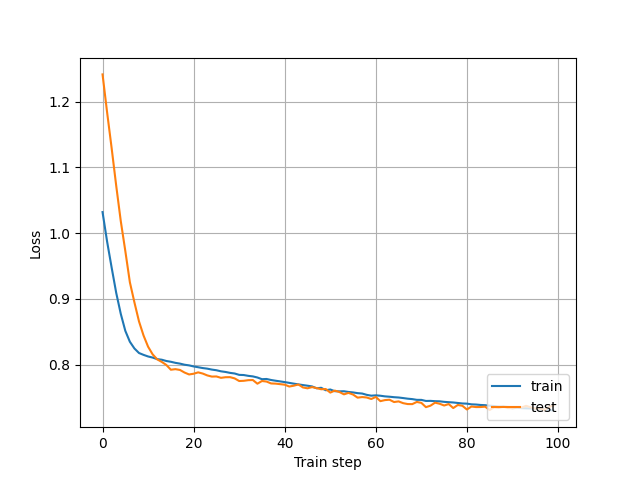}
      \caption{1600 time points.}
    \end{subfigure}
`   %
    \caption{
    The curves of training and testing losses when the available time series data are of  different sizes.
    }
    \label{fig:toy_example}
    \vspace{85ex}
\end{figure}

According to \cite{farnoosh2020deep}, we generate a toy time series dataset with $n$ time points in which each point is a $d$-dimension variable:
\begin{equation*}
    w_t = 0.5w_{t-1} + \text{tanh}(0.5w_{t-2}) + \text{sin}(w_{t-3}) + \epsilon \, ,
    \quad
    X = [w_1, w_2, ..., w_n]*F + v
\end{equation*}
where $w_t \in \mathcal{R}^2$, $F \in \mathcal{R}^{2 \times d} \sim \mathcal{U}[-1, 1]$, $\epsilon \sim \mathcal{N}(0,I)$, $v \sim \mathcal{N}(0, 0.5I)$, and $d=5$. 
An input-8-predict-8 window is utilized to roll this synthetic dataset.
We split this synthetic dataset into training and test sets with a ratio of 7:3. 
We train the RNN in 100 epochs at most, and the MSE loss of training and testing are plotted in~\cref{fig:toy_example}.
It can be seen that the inflection points of the loss curves move back gradually and disappear as increasing the size of the dataset. 
Besides, with fewer time points, like, 400, the model can be overfitted more easily.

\end{document}